%% file: paper.tex
\documentclass[11pt]{article}

\usepackage{amsmath}

\usepackage{tikz}
\usepackage[utf8]{inputenc} 
\usepackage[T1]{fontenc}    
\usepackage{hyperref}       
\usepackage{url}            
\usepackage{booktabs}       
\usepackage{amsfonts}       
\usepackage{nicefrac}       
\usepackage{microtype}      
\usepackage[capitalize,noabbrev]{cleveref}      

\usepackage{lipsum}         
\usepackage{graphicx}
\usepackage{natbib}
\usepackage{doi}

\usepackage{mathtools}
\usepackage{amsthm}
\usepackage{subfigure}
\usepackage{gensymb}
\usepackage{enumitem}
\usepackage[margin=1in]{geometry}
\usepackage{authblk}
\usepackage{algorithm}
\usepackage{algorithmic}
\newcommand{\ignore}[1]{}

\input{macro.tex}
\title{
A Categorical Framework of General Intelligence
}

\author[1,2,3]{Yang Yuan}

\affil[1]{\footnotesize IIIS, Tsinghua University}
\affil[2]{\footnotesize Shanghai Artificial Intelligence Laboratory}
\affil[3]{\footnotesize Shanghai Qi Zhi Institute}

\let\svthefootnote\thefootnote
\newcommand\freefootnote[1]{%
	\let\thefootnote\relax%
	\footnotetext{#1}%
	\let\thefootnote\svthefootnote%
}
\begin{document}
\maketitle

\begin{abstract}
Can machines think? Since Alan Turing asked this question in 1950, nobody is able to give a direct answer, due to the lack of solid mathematical foundations for general intelligence. In this paper, we introduce a categorical framework towards this goal, with two main results. 
First, we investigate object representation through presheaves, introducing the notion of self-state awareness as a categorical analogue to self-consciousness, along with corresponding algorithms for its enforcement and evaluation.
Secondly, we extend object representation to scenario representation using diagrams and limits, which then become building blocks for mathematical modeling, interpretability and AI safety.
As an ancillary result, our framework introduces various categorical invariance properties that can serve as the alignment signals for model training.
\end{abstract}

\section{Introduction}
In recent years, remarkable progress has been made in training foundation models with  enormous computational power, vast amounts of data, and gigantic neural networks~\citep{radford2021learning, chen2020simple, radford2019language, brown2020language, ramesh2021zero, ramesh2022hierarchical, sohl2015deep, rombach2022high, he2022masked}.
Surprisingly, despite the impressive achievements, the internal working mechanisms of these models remain mysterious.
People seem to reach the consensus that the foundation models are inherently black-box and uninterpretable, therefore empirical experimentation is the only way of pushing AI forward.

While this is indeed what happened in the past decade and 
is analogous to how intelligence is acquired through evolution,
relying solely on empirical experimentation without theoretical understanding can be both inefficient and dangerous. The inefficiency arises from the fact that the progress is made through trial and error, often guided by intuition, and the milestones are defined indirectly based on performance on specific tasks rather than a comprehensive understanding of intelligence itself. 
The potential danger stems from the fact that nobody knows what we will get at the final destination, and perhaps more importantly, how close we are right now. 
We do not even know whether we have already created the general intelligence --- maybe not yet, but how to make such evaluations?

In this paper, we present a categorical framework of general intelligence, which contributes to answering the following questions:

\begin{enumerate}
	\item[Q1:] Can the model be aware of its self-state? (Section~\ref{sec:world})
	\item[Q2:] How shall the model represent complex scenarios?  (Section~\ref{sec:communication})
	\item[Q3:] How shall we train the model towards general intelligence? (Section~\ref{sec:train})
\end{enumerate}

It will be extremely challenging, if not impossible, to prove that our framework is for \emph{the} general intelligence, given the absence of consensus on the formal definition of general intelligence among human beings. 
Instead, we take the categorical approach: we formally define all the basic elements, state their theoretical implications, specify the algorithmic requirements, and finally integrate all the elements into a comprehensive framework. Therefore, even if one disagrees with our definition of general intelligence or believes that certain crucial pieces are missing, our framework remains relevant and applicable.
\begin{figure}[th]
	\centering
	\input{1.tex}
	\caption{Our categorical framework}
	\label{fig:framework}
\end{figure}
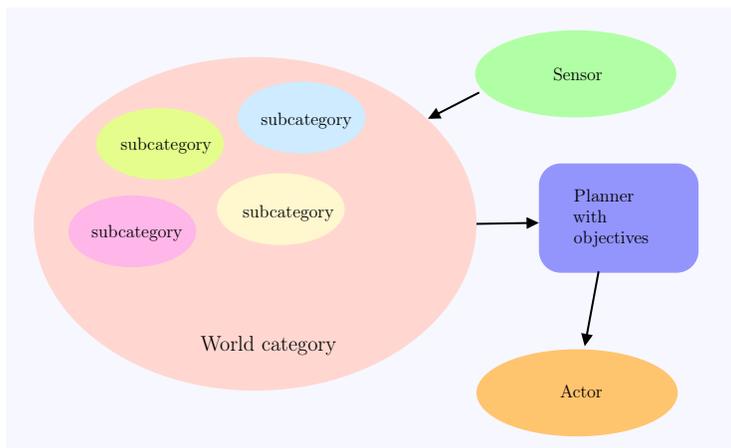

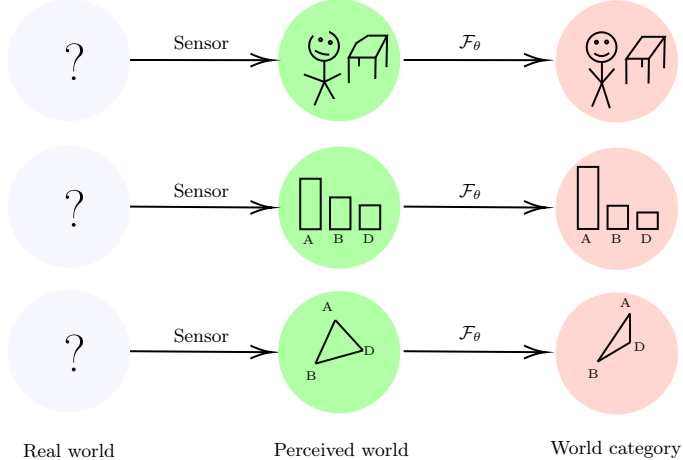
\begin{figure}[th]
	\centering
	\input{5.tex}	
	\caption{
		Case studies on three different worlds. 
		In the first row, the model sees a human being with a desk, but the human is partly occluded and the desk has one corner missing. In the world category, when the model memorizes the scene, it may complete the occluded parts with minor adjustments on the human body, and slightly modify the shape of the desk. 
		In the second row, the model sees a chart representing the wealth of Alice, Bob and David. However, the model may not accurately memorize the information in its world category. 
		In the third row, the model perceives the relationships among Alice, Bob and David, e.g., whether two of them are close friends. In its world category, such relationships might be preserved with distortion.
	}
	\label{fig:three_worlds}
\end{figure}

Our framework is surprisingly simple, consisting of four main components\footnote{Temporarily ignoring the categorical part, which is our main contribution, we shall remark that the components and connections in our framework are similar but different to the ones proposed in \citet{lecun2022path}, as discussed in Section~\ref{sec:related}.}: the sensor, world category, planner with objectives, and actor, with single direction information-flow  (see Figure~\ref{fig:framework}).
The sensor receives multi-modal signals from the external environment, including but not limited to text-input, video/image-input, audio-input, sense of touch, etc. The world category perceives and comprehends the incoming signals, and updates its internal state accordingly. 
The planner continuously monitors the status of the world category and generates plans based on its objectives. 
Finally, the actor executes these plans, influencing the external environment by generating outputs such as text-output, video/image-output, audio-output, robot-manipulation signals, etc. We elaborate the details below, starting from three different world views.

\textbf{Three different worlds} exist in our framework (see Figure~\ref{fig:three_worlds}): the real world, perceived world, and world category. The model cannot directly access the real world\footnote{It is entirely possible that this ``real world'' is simulated, and we do not make any distinctions here.}, and can only access the perceived world using its sensor. 
For any given time $t\in \mathbb{N}$, the real world can be much larger than the perceived world of the model as the model may only perceive a small part of the world. Even under perceived scope of the sensor, the real world and perceived world are not necessarily the same, because the sensor may have limited sensory ability, be biased or contain noise. 
Moreover, the world category encodes what the model understands about the world, so it can be much larger than the perceived world. Intuitively, the perceived world is what the model sees and feels momentarily, and the world category is what it understands, memorizes and predicts about the world. 

In order to demonstrate intelligent behaviors, the model should interact with the external environment through its sensor and actor, so the notion of time $t$ is important. 
In case the model does not have the access of the time stamps, 
$t$ may instead denote the index of distinct events that are detected by the sensor, e.g., chatGPT answering various queries from the user.
For the sake of simplicity, we do not make such distinctions and simply treat $t$ as the time stamp.

The world category can be seen as an imaginary reconstruction of the external world by the model through its sensor. It comprises all the people, animals, objects, knowledge from the external environment that are perceptible by the model through its sensor, as well as abstract representations on top of them. More precisely, it is a function in $\mathbb{N}\rightarrow \mathbf{Cat}$, representing a dynamic category\footnote{Readers not familiar with category theory may check Section~\ref{sec:prelim} for a basic introduction.} containing various objects and changing over time $t\in \mathbb{N}$. 
The sensor decides the types of elementary objects in the world category. For example, if the model is incapable of detecting visual signals, the world category will not contain image objects. Moreover, if the sensor receives signals from a simulated environment, the world category will only contain simulated knowledge, which can differ significantly from the real world.

\textbf{Object representation.} 
For any given object $X$, the world category never directly stores $X$, but uses neural networks to store all the relationships between other objects and $X$ instead, which contains sufficient information about $X$. As a notable example, when the model is able to perceive the relationships between 
itself and other objects, its world category can  represent an object called ``self-state'' for storing such relationships. Is maintaining the self-state equivalent to having self-consciousness?
This is a controversial question that we choose not to answer.
Instead, we formally define the notion of self-state based on category theory without bothering its relationship to self-consciousness (Section~\ref{sec:self}).

Based on our definition of self-state,
we introduce two algorithms for enforcing and evaluating self-state awareness of the model. Unlike the identification of self-consciousness, which is a binary variable denoting whether a subject possesses self-consciousness or not, our evaluation generates a continuous value within the interval $[0,1]$ to indicate the degree of self-state awareness. 
This degree corresponds to the proportion of all relevant relationships between the subject and other objects or tasks that the subject is aware of.
According to our definition, it appears that even many human beings, especially the kids, may not possess perfect self-state awareness.

\textbf{Scenario representation.} 
How shall the model represent scenarios with multiple objects and morphisms in between? 
Using the language of category theory, we can define the scenario content as a  diagram, and define the scenario itself as a projective limit over the diagram. 

As we will see, using diagrams and limits for scenario representation has various interesting consequences.
For example, it makes mathematical modeling much easier, because we may take proper abstraction of diagram, to directly convert it into a mathematical problem. Besides, 
by treating the scenario content as a diagram of the world category, the model can generate interpretations based on its internal knowledge, not only limited to assigning weights to the input variables like the attribution methods~\citep{sundararajan2017axiomatic,lundberg2017unified}. 
Moreover, diagram representation of the scenario content allows a functional approach for AI safety, by injecting self-state into the diagram and enforcing the self-state to be human-friendly.

\textbf{Invariance property as training signals}. Category theory employs commutative diagrams to characterize the equivalence of distinct computational paths, which naturally leads to various invariance properties for the model. 
Unlike supervised learning where the training objective is to fit the input data with the correct output label, foundation models focus on learning the morphisms between objects, and functors between categories. 
The invariance properties serve as the training signals for the model to adjust itself, so that the world category is naturally consistent.

\section{Preliminaries}
\label{sec:prelim}

Category theory is used in almost all areas of mathematics. Here we only introduce the necessary notions for understanding the results of our paper. Curious readers may check 
\citet{mac2013categories,riehl2017category, adamek1990abstract} for a more comprehensive introduction. 

\subsection{Category basics}
A category $\cat$ has a set of objects $\ob(\cat)$, and 
a set of morphisms $\hom_\cat(X,Y)$ from $X$ to $Y$ 
for every $X, Y\in \ob(\cat)$. In this paper, we use ``relationships'' and ``morphisms'' interchangeably. 
Given $f\in \hom_\cat(X,Y), g\in \hom_\cat(Y,Z)$, we define their composition as $g\circ f\in  \hom_\cat(X,Z)$. Notice that $\circ$ is associative, i.e.,  $(h\circ g) \circ f=h\circ (g\circ f)$. 
For every $X\in \ob(\cat)$, there exists an unique identity morphism $\id_X \in \hom_\cat(X,X)$. 
A morphism $f: X\rightarrow Y$   is an isomorphism if there exists $g: X\leftarrow Y$ such that $f\circ g=\id_Y$ and $g\circ f=\id_X$. 
In this case, we say $X$ and $Y$ are isomorphic and write $X\simeq Y$. 

We consider a universe $\mathcal{U}$\footnote{Check \citet{kashiwara2006categories} for the related definitions.}. 
A category is a $\mathcal{U}$-category, if 
$\hom_\cat(X,Y)$ is $\mathcal{U}$-small for any $X, Y\in \ob(\cat)$. 
A $\mathcal{U}$-category category $\cat$ is $\mathcal{U}$-small if $\ob(\cat)$ is $\mathcal{U}$-small. For simplicity, below we may not explicitly mention the universe $\mathcal{U}$, and simply write that $\cat$ is a category or a small category. We define $\mathbf{Cat}$ to be the category whose objects are small categories.

Given a category $\cat$, we define its opposite $\cat^{\op}$ by setting $\ob(\cat^{\op})=\ob(\cat)$ and  $\hom_\cop(X,Y)=\hom_\cat(Y,X)$. Moreover, given $f\in \hom_\cop(X,Y), g\in \hom_\cop(Y,Z)$, the new composition is $g
\mathrel{\overset{\makebox[0pt]{\mbox{\normalfont\tiny\sffamily op}}}{\circ}}  f= f\circ g\in  \hom_\cop(X,Z)$. 

We define $\set$ to be the category of sets, 
where the objects are sets, and $\hom_\set(X,Y)$ is the set of all functions with domain $X$ and codomain $Y$. 
Notice that we ignore the subtleties about the universe for better presentation, so here just assume that $\set$ does not contain strange objects like a set containing all sets.

Functor is like a function between two categories. Given two categories $\cat, \cat'$, a functor $F:\cat\rightarrow \cat'$ 
maps objects from $\cat$ to $\cat'$ with $F:\ob(\cat)\rightarrow \ob(\cat')$ and
morphisms from $\cat$ to $\cat'$ with $F: \hom_\cat(X,Y)\rightarrow \hom_{\cat'}(F(X),F(Y))$ for all $X,Y\in \cat$, so that $F$ preserves identity and composition. Formally, we have $F(\id_X)=\id_{F(X)}$ for all $X\in \cat$, and 
$F(g\circ f)=F(g)\circ F(f)$ for all $f:X\rightarrow Y, g: Y\rightarrow Z$.

The morphisms of functors, also called the natural transformation, is the way to transform the functors while preserving the structure. Given two categories $\cat, \cat'$, and two functors $F_1, F_2$ from $\cat$ to $\cat'$. A morphism of functors $\theta:F_1\rightarrow F_2$ has a morphism $\theta_X: F_1(X)\rightarrow F_2(X)$ for all $X\in \cat$ such that for all $f:X\rightarrow Y\in \hom_\cat(X,Y)$, we have $\theta_Y(F_1(f)(F_1(X))) =
F_2(f)(\theta_X(F_1(X)))$. 

A presheaf is a functor from $\cop$ to $\set$, 
and $\catw$ is the category of presheaves. 
Similarly, 
a functor from $\cop$ to $\set^\op$ is called a $\set^\op$-valued presheaf, and $\catv$ is the category of $\set^\op$-valued presheaves. 
In this paper we do not make the differentiation, and name both kinds of functors as presheaves, and $\catw, \catv$ as the categories of presheaves. 
Moreover, define the Yoneda functors $\hc(X)\triangleq \homc(\cdot, X)\in \catw$, and $\kc(X)\triangleq \homc(X, \cdot)\in \catv$.  
The following lemma is fundamental.
\begin{mylem}[Yoneda lemma] 
	\label{lem:yoneda}	
	Given $X\in \cat$ we have, 
	\begin{enumerate}
		\item[(1)]
		For $A\in \cat^{\wedge}$, $\hom_{\cat^\wedge}(h_\cat(X), A)\simeq
		A(X). $
		\item[(2)] 
		For $B\in \catv$, $\hom_{\catv}(B, \kc(X))\simeq
		B(X). $
	\end{enumerate}	
\end{mylem}

Yoneda lemma says $\hc(X)$ and $\kc(X)$ capture all the information of $X$. As a directly corollary, we have $
\hom_{\cat^\wedge}(h_\cat(X), h_\cat(Y))\simeq 
h_\cat(Y)(X)= \hom_\cat(X,Y)$, and similar result holds for $\kc(\cdot)$. 
A functor $F$ from $\cop$ to $\set$ (or $\cat$ to $\set$) is representable if there is an isomorphism between $\hc(X)$ (or $\kc(X)$) and $F$ for some $X\in \cat$. Such $X$ is called a representative of $F$.

\subsection{Limits}
\label{sec:limit}
A diagram of shape $A$ in a category $\cat$ is a functor $\alpha:A\rightarrow \cat$, which selects objects in $\cat$ correspond to $\ob(A)$, that preserve the morphisms in $A$. 
Given a functor $\beta: A^\op \rightarrow \set$, define its projective limit as
$\llim \beta\triangleq \hom_{A^\wedge}(\text{pt}_{A^\wedge}, \beta)$, where $\text{pt}_{A^\wedge}(i)=\{\text{pt}\}$ for every $i\in A$, and $\{\text{pt}\}$ is the single point set. 
In other words, $\llim\beta$ denotes the set of all natural transformations between $\text{pt}_{A^\wedge}$ and $\beta$. 
Based on this definition for diagrams in  $\set$, we have the general definition of the limits.

\begin{mydef}[Projective and inductive limits]
	\label{def:limits}
	Given $\alpha: A\rightarrow \cat, \beta: A^\op\rightarrow \cat$ with small $A$, 
	the inductive limit $\rlim \alpha \in \catv$ and projective limit $\llim\beta\in \catw$ are defined as:
	\begin{enumerate}
		\item[(1)] $\rlim\alpha: X\mapsto \llim \homc(\alpha, X)$
		\item[(2)] $\llim \beta: X\mapsto \llim \homc(X, \beta)$	
	\end{enumerate}
	Here $\homc(\alpha, X)$ is a functor  that maps $i\in A$ to $\homc(\alpha(i), X)$.
	Therefore, $\llim \homc(\alpha, X)$ is a  well defined  limit for a diagram in $\set$.
	Same argument holds for $ \llim \homc(X, \beta)$.
\end{mydef}

\begin{mylem}[p.60 in \citet{kashiwara2006categories}]
	\label{lem:ind-mor}
	If $A$ is small, consider $\alpha: A\rightarrow \catw, \beta: A^\op \rightarrow \catv$. 
	For $A\in \catw, B\in \catv$,
	\[
	\hom_\catw(\rlim \alpha, A)\simeq \llim \hom_\catw(\alpha, A)
	\]	
	\[
	\hom_\catv(B, \llim \beta)\simeq \llim \hom_\catv(B, \beta)
	\]
\end{mylem}

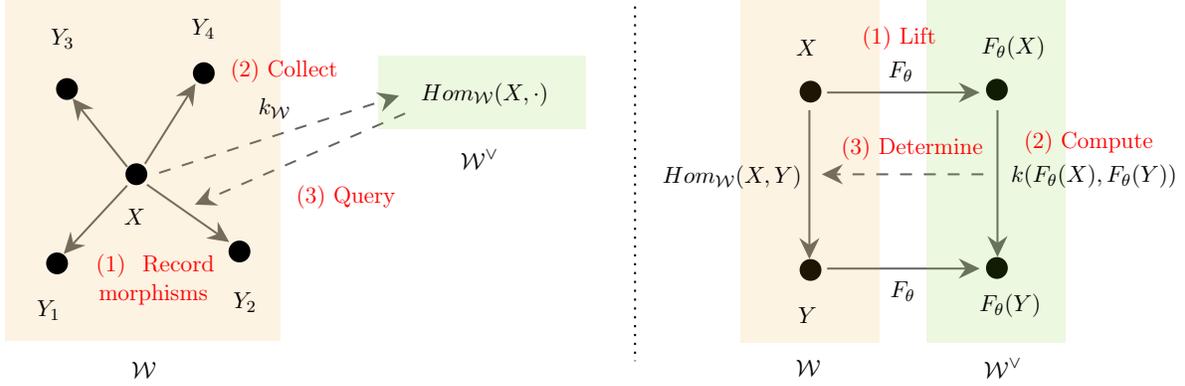
\begin{figure*}
	\begin{center}
		\input{12.tex}
	\end{center}
	\caption{Two ways of constructing the world category. Left: assuming there exists a category $\world$ with morphisms predefined, we can use $\kc$ to directly compute $\hom_{\world}(X, \cdot)$ for given $X\in \world$, and then  query $\hom_{\world}(X, Y)$ for any $Y\in \world$. Right: 
	the morphisms in $\world$ were not known, so we first compute $\fw(X), \fw(Y)$ for given $X, Y\in \world$, then compute $k(\fw(X), \fw(Y))$ in $\worldv$, which determines 
	$\hom_{\world}(X, Y)$. 
}
	\label{fig:altview}
\end{figure*}

\section{World Category And Object Representation}
\label{sec:world}
Recall that the world category is the imagination of the model about the real world.
Given time $t\in \mathbb{N}$, how shall we represent the world category snapshot $\world(t)$? 
$\world(t)$ contains both objects and their morphisms, but 
directly storing these information is usually computationally infeasible. Instead, we use a function $\fwt:\world(t)\rightarrow \worldv(t)$ parameterized with $\theta(t)$,  which maps object $X$ to $\kw(X)$ that contains all the morphisms of $X$. 
By doing that, the model never explicitly store any morphisms in $\world(t)$, but contains all the necessary information using $\fwt$. 
Here we assume $\fwt$ maps objects to $\worldv$ just for notational convenience, and empirically $\fwt$ can contain the information of both $\worldw$ and $\worldv$, i.e., both $\hom_{\cat}(\cdot, X)$ and $\hom_{\cat}(X, \cdot)$ for each $X$. 
For example, the embedding space that $\fwt$ maps to can be written as $\worldw \times \worldv$, although neural networks can find much better encoding mechanism empirically.

In our framework, the model only represents a single snapshot $\world(t)$
using $\fwt$ at any given time $t$, which changes over time. When there is no confusion, we ignore the parameter $t$ and simply write $\world$ as the snapshot, and write the function $\fwt$ as $\fw: \world \rightarrow \worldv$.  The following definition is extremely important and interesting.

\begin{mydef}[World category based on $\fw$]
\label{def:world-from-fw}
Assuming the objects in $\world$ are fixed. 
Given the world category representation function $\fw: \world \rightarrow \worldv$ with a data oblivious function $k: \worldv\times \worldv \rightarrow \set$ representing the morphisms in $\worldv$, for any $X, Y\in \world$, we have
$\hom_\world(X,Y)\triangleq k(\fw(X), \fw(Y))$.
\end{mydef}

Here, data-oblivious means $k(\cdot, \cdot)$ is predefined without seeing the data. For example, it can be defined as the inner product between the two inputs. 
To understand Definition~\ref{def:world-from-fw}, consider the
mapping $X\rightarrow \hom_{\world}(X, \cdot)$ from $\world$ to $\worldv$. This mapping is natural, because if we know $X$ and all the relationships around $X$, we can compute $\hom_{\world}(X, \cdot)$ that encodes all such relationships. 
However, Definition~\ref{def:world-from-fw} considers the opposite direction, where we know $\kw(X), \kw(Y)$ (represented by $\fw(X)$ and $\fw(Y)$ in Definition~\ref{def:world-from-fw}), and we want to recover  $\hom_\world(X,Y)$. 
This might be counter-intuitive at the first glance, because 
if we send $Y$ to $\kw(X)$, we have $\kw(X)(Y)=\hom_\world(X,\cdot)(Y)=\hom_\world(X,Y)$. In other word, the definition of $\hom_W(X,Y)$ is cyclic, as we use $\kw(Y)$ to define $\hom_W(X,Y)$, but $\kw(Y)$ contains $\hom_W(X,Y)$ in its own definition! 

How can we define $\kw(Y)$ without using $\hom_W(X,Y)$?
Definition~\ref{def:world-from-fw} makes this cyclic definition possible, because $\kw(\cdot)$ is represented by a function $\fw(\cdot)$ that maps an object to an embedding vector. In other words, all the relationships of $Y$ are embedded by $\fw(Y)$, and $\hom_W(X,Y)$ can be recovered from $\fw(X)$ and $\fw(Y)$. See Figure~\ref{fig:altview} for an illustration. 
Therefore, the world category is defined and represented by a function $\fw$. Starting from this definition, we have:
\begin{itemize}
	\item Even if $\world$ contains infinitely many objects, $\fw$ can still be finite (e.g., a neural network), and encode infinitely many morphisms. 
	\item In order for the model to know about an object $X$ in $\world$, it suffices to let model learn  the morphisms of $X$ in $\worldv$. Therefore, directly observing $X$ is not necessary.
	\item The world category is controlled by $\fw$, so what $\fw$ computes is what the model understands. $\fw$ is the  ground-truth for world category. 
	\item The world category generated by $\fw$ can be very different from the perceived world or real world, especially when $\fw$ has limited representation power.
	\item Even if the model previously knows little or nothing about an object $X\in \world$, it can still generate lots of morphisms about $X$ based on $\fw(X)$. 
\end{itemize}

Since $\fw$ determines $\world$ for a given model, below we may use $\fw$ to denote the world category.  Other than morphisms, we can also use $\fw$ to encode the tasks. 
\begin{mydef}[Task]
A task $T:\world^\op \rightarrow \set^\op$ is a functor in $\worldv$. 
\end{mydef}

By Yoneda lemma,  $T(X)\simeq \hom_\worldv(T,\kw(X))$. 
Therefore, when a task $T$ is representable, 
 $T(X)$ can be computed by $k(T,\fw(X))$.

\begin{figure*}
	\begin{center}
	\input{4.tex}
\end{center}
\caption{Two different representations of self-state. The first subfigure is an illustration on the model $I$ and corresponding morphisms between $I$ and other objects. The second subfigure is the object representation $I$, and the last subfigure is the morphism representation $I^\vee$, which is a presheaf.  }
\label{fig:self}
\end{figure*}
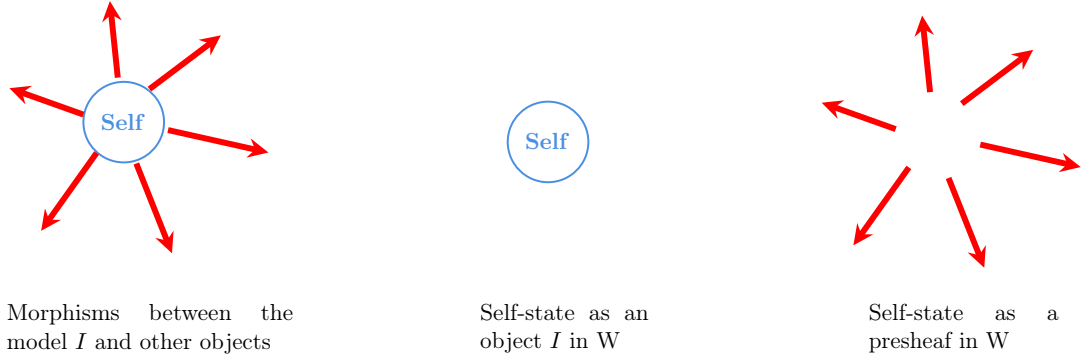

\subsection{Self-state}
\label{sec:self}
In the world category, there may exist a special object called self-state, defined below.

\begin{mydef}[Self-state]
\label{def:self-state}
Given $I$ as the object representing the model in $\world$, 
the self-state in the world category is the presheaf $I^\vee\triangleq \fw(I)\in \worldv$. 
\end{mydef}

The object $I\in \world$ and the presheaf $I^\vee$ are very different, in the sense that $I$ is a single object without any additional information, but $I^\vee$ contains all the morphisms between $I$ and other objects. Moreover, by Yoneda lemma, we also have $\hom_{\worldv}(T,I^\vee)\simeq T(I)$ for any task $T\in \worldv$. That means, $I^\vee$ also encodes all the information that every relevant task needs.

As we discussed for Definition~\ref{def:world-from-fw},  even if the model cannot perceive itself in the perceived world, it can still have a self-state. For example, consider the following thought experiment. 

\begin{mythought}[Paralyzed person who can see and hear]
Consider a human being $A$, who cannot control his body and loses all the body feelings, including smell, skin sensation,
eyelids control, heartbeat, temperature feelings, etc. The only way $A$ can accept information from the world is through his eyes and ears, ever since he was born. Can $A$ maintain a self-state? 
\end{mythought}

It is unclear whether such person ever existed, but in this thought experiment, since $A$ can see and hear, he can still accept information from the external world. For example, other people can show daily images and videos to him, or talk to him, e.g., 
 ``you are Bob'', ``I am your father'', ``you are now 10 years old'', ``you were born in the $\alpha$-town'', etc. 

None of these information represents $A$ itself, but all of them are describing the relationship between $A$ and other things in the perceived world, which are exactly encoded in $I^\vee$. In other words, by Definition~\ref{def:self-state}, $A$ can still have self-state in his brain, representing all the relationships to himself. This immediately triggers the following thought experiment: what if $A$ does not have a body?

\begin{mythought}[Chatbot]
	Consider a chatbot $A$, who interacts with the external world with text and image, e.g., GPT-4. Can $A$ maintain a self-state? 
\end{mythought}

By the above discussion, the answer is yes, if $A$ perfectly understands all the relationships between other objects and itself. In other words, in Definition~\ref{def:self-state}, a self-body is not necessary in order to maintain a self-state. Instead, it suffices for the agent to perceive all the relationships between other objects and itself. 

Not all world categories have self-states. For example, if we train a foundation model using SimCLR~\citep{chen2020simple} on images, all the relationships are describing similarities between images, and the world category does not have the self-state. Interestingly, almost all the existing computer programs belong to this kind. Combining the discussions together, we have the following corollary. 

\begin{mycor}[Self-state criterion]
A model can maintain a self-state, if and only if it can learn the presheaf $I^\vee$ through its sensor. 
\end{mycor}
It is possible that all the relationships encoded in $I^\vee$ are not directly feed into the model, but can be inferred instead. In this case, the model can still have a self-state. This gives the next corollary. 

\begin{mycor}[Self-state emergence]
	\label{cor:emergence}
For a given model, if to reconstruct the perceived world, 
learning the relationships between the model itself and other objects is inevitable, then this model will maintain a self-state. 
\end{mycor}
For example, chatGPT has to encode almost every things that can be described with natural language in the world, and it seems inevitable to encode the relationships between other objects and itself. Therefore, Corollary~\ref{cor:emergence} says chatGPT will maintain a self-state.

When the model maintains a self-state object, 
the maintenance may not be accurate as the world category is dynamically learned.  We denote self-state awareness as how the object accurately represents the model. To test self-state awareness, we shall first define the corresponding tests. 

\begin{mydef}[Self-state awareness test]
\label{def:self-test}
A self-state awareness test is a functor $T: \worldv\rightarrow \{0,1\}$, that takes a presheaf $I^\vee$ in $\worldv$, and outputs whether $I^\vee$ passes the test $T$. 
\end{mydef}

For example, if the model has a name ``Sydney'', a self-state awareness test will be a functor that takes $I^\vee$ as the input,  evaluates $\hom_\worldv(I^\vee, \kw(\text{``Sydney''}))
\simeq \hom_\world(I, \text{``Sydney''})$, and outputs
if the morphism indeed represents that ``Sydney'' is $I$'s name. 
However, simply passing one test is not enough to declare self-state awareness, and we need to set a variety of tests. 

\begin{mydef}[Self-state awareness under $\mathcal{T}$]
\label{def:self-awareness}
Given a set of self-state tests $\mathcal{T}$, $\delta\in [0,1]$, when a model has self-state $I$ in its world category, it has  $\delta$-awareness of its self-state under $\mathcal{T}$ if 
$\mathbb{E}_{T\in \mathcal{T}}(T(I))\geq \delta$.
\end{mydef}

The choice of test set $\mathcal{T}$ depends on the test objectives. When the test set is picked such that the signals are difficult to perceive, even the human beings may not easily pass the tests. For example, if you have a kidney stone inside your kidney, you can only be aware of this fact when you do a kidney scan or by experiencing pain. Similarly, when situated in a noisy environment and someone calls out your name, you may be unable to react promptly.

\begin{algorithm}
	\caption{Evaluating self-state awareness}
	\label{alg:test-self-state}	
	\begin{algorithmic}
		\STATE {\bfseries Input:} the world model $\fw$, 
		self object $I$ in $\world$, self-state test set $\mathcal{T}$
		\STATE Let $s=0$
		\FOR{$i=1$ to $m$}
		\STATE Sample a task $T\in \mathcal{T}$
		\STATE Let $s=s+T(\fw(I))$
		\ENDFOR
		\STATE Return $s/m\in [0,1]$
	\end{algorithmic}
\end{algorithm}

\begin{algorithm}
	\caption{Enforcing self-state awareness}
	\label{alg:learn-self-state}	
	\begin{algorithmic}
		\STATE {\bfseries Input:} the world model $\fw$, 
		self object $I$ in $\world$, self-state test set $\mathcal{T}$
		\STATE Let $s=0$
		\FOR{$i=1$ to $m$}
		\STATE Sample a task $T\in \mathcal{T}$
		\STATE Let $\ell(T,I, \fw)=-T(\fw(I))$
		\STATE Run backpropagation on $\ell(T,I, \fw)$ to optimize $\fw$
		\ENDFOR
	\end{algorithmic}
\end{algorithm}

Definition~\ref{def:self-awareness} immediately gives Algorithm~\ref{alg:test-self-state} for evaluating  self-state awareness. 

\subsection{Learning self-state}
How should the model learn its self-state? Based on the previous discussion, we present Algorithm~\ref{alg:learn-self-state} as a general solution. However, Algorithm~\ref{alg:learn-self-state} needs the supervised signals from the test set $\mathcal{T}$, 
which is not generally available. In order to learn state-state without the test set, some kind of prior knowledge for the model might help. For example, the model may assume that:
\begin{itemize}
	\item What it can control is itself. 
	\item What it can feel from its private sensors is itself. Here, the private sensors are the sensors like heartbeat, temperature feelings, skin sensation, etc, which are pre-defined self-sensors.		
\end{itemize}

Take human hands as the example. Based on Definition~\ref{def:self-state}, we feel the hands are part of the body, because the brain tells us \textbf{the relationship} that the hands in sight belong to our body. Indeed, the human brain actively aligns the sensory  signals in real time, so that we will have a comprehensive feeling of ownership of our hands. Specifically, when we see someone touches our hand, and feel the touch at the same time, our brain will quickly adjust these two signals to make sure they are referring to the same thing of our body. 

This learning perspective is closely related to the intriguing observations in neural science called the rubber hand illusion~\citep{ehrsson2004s,botvinick1998rubber,tsakiris2005rubber}. In the experiment, the experimenter simultaneously strokes one hidden real hand of the human participant, as well as a rubber hand in front of the participant. Since the stroking feeling from the real hand, and the vision signal on the rubber hand are sent to the brain simultaneously, the human participant will quickly have the ownership feeling of the rubber hand. By replacing the visual signals with auditory feedback, we will get similar experimental results. Therefore, if the model uses similar learning algorithm for aligning multi-modal signals (e.g. contrastive learning), it will experience similar kind of illusion.

\subsection{Empathy}
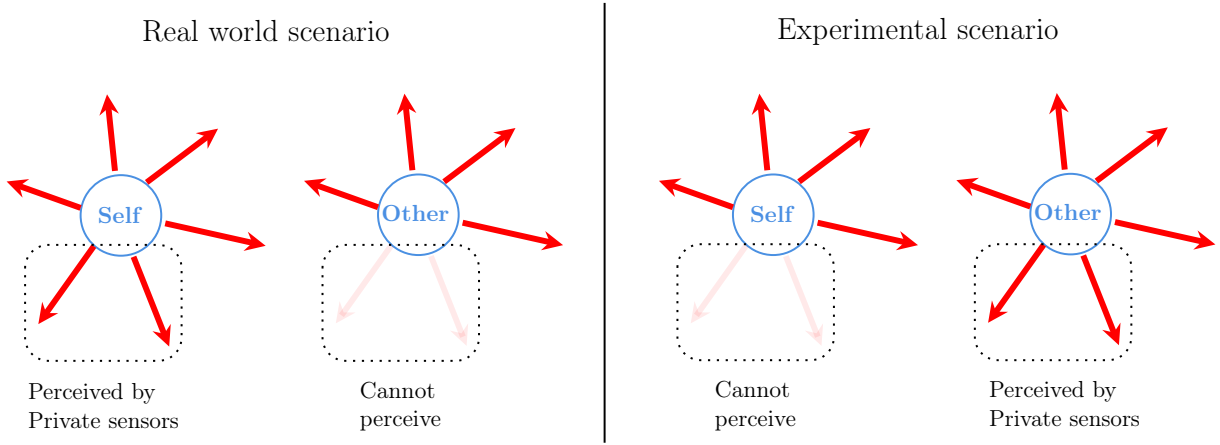
\begin{figure*}
	\begin{center}
		\input{8.tex}
	\end{center}
	\caption{Two distinct scenarios on self-states and other-states.
	In real world,  the self-state is fundamentally different from other-states when the model only has access to its private sensors, but not the sensors of other agents. However, in an experimental scenario, if the model only has access to an other agent's private sensors,  the model quickly generates a sense of body ownership for the external agent. }
	\label{fig:other-state}
\end{figure*}
Generalizing Definition~\ref{def:self-state} to other agents, we have:

\begin{mydef}[Other-State]
	\label{def:empathy}
	Given $A$ as the object representing an agent $A$ in $\world$, 
	the state for $A$ in the world category is the presheaf $A^\vee\triangleq \fw(A)\in \worldv$. 
\end{mydef}

Therefore, all the discussions about self-state also apply to the state for other agents. For example, the model does not have to directly ``meet'' an agent face-to-face in order to maintain a corresponding state, as long as it can infer many relationships about the agent. 

However, this is different from our daily experience, as we know in our gut that the self-consciousness is very different from the empathy for other people. Why are the categorical definitions for these two notions same here? 
This is because of the existence of private sensors (like heartbeat, temperature feelings, skin sensation, etc.), which is available for representing the self-state, but not for other states. 
Consider the following three cases. 

\begin{enumerate}[leftmargin=0.6cm]
	\item In the full information setting, when private sensors are irrelevant for the discussion,  empathy and self-state awareness are the same. For example, in a multi-agent game where each agent has its action set, status and reward function, the empathy can be very helpful in understanding and predicting every agent's situation and behavior. 
	\item If other agents have private sensors, full-empathy for them cannot be achieved. Specifically, if the private sensors cannot be perceived by our model, and the self-state tests $\mathcal{T}$ includes tests related to these sensors, then it is impossible for our model to pass these tests. 
	\item If the model has access to the private sensors of other agents, there exists little difference between self-state awareness and empathy for other agents.	
\end{enumerate}

See Figure~\ref{fig:other-state} for an illustration, where the experimental scenario was well known in neural science.
Specifically, in the immersive virtual reality environment, the participant will experience the body ownership over the avatar, when given the first-person signals of the avatar~\citep{kilteni2012sense,guterstam2015posterior,pavone2016embodying,buetler2022tricking}. We conjecture that the model will have the same experience. In other words, the boundary between the model itself and other agents is not as strict as we would normally imagine.

\section{Scenario Representation}
\label{sec:communication}
\begin{figure}[th]
	\centering
	\input{13.tex}
	\caption{
Two ways of representing scenario $S$. Left: given a scenario, we first define its objects and morphisms as a diagram, then take the limit $S$ of the diagram, lift the limit to $\worldv$ and get $S^\vee$ as a feature vector. Right: the decomposition of the concept was not known, so we compute $\fw(S)$ as a projective limit in $\worldv$, then extract objects from $\fw(S)$ like $\fw(X)$ with $X\in \world$. For two objects $\fw(X),\fw(Y)$, we compute their morphisms $k(\fw(X), \fw(Y))$, which determines $\hom_{\world}(X, Y)$. 
}
	\label{fig:altview2}
\end{figure}
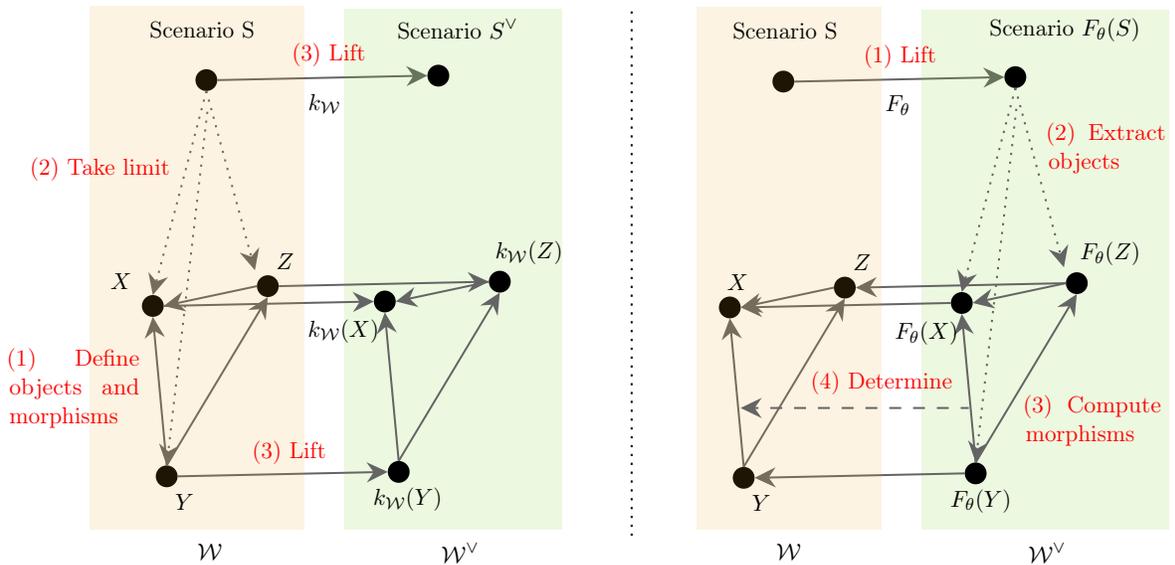

The self-state (or other-state) is usually used in a concrete scenario with other objects presented, rather than being used alone. 
For example, consider the scenario $S$ that a robot teaches machine learning in front of twenty students with math background in a classroom. The task $T$ is: what shall it do?
This question is highly related to its self-state, 
but other objects like students, classroom, lecture topic are also important
factors. Therefore, learning $I^\vee$ perfectly does not help, as Yoneda lemma only holds for $T(I)\simeq \hom_{\worldv}(T, I^\vee)$, not for $T(S)$. 
Formally,  what is a scenario $S$? Scenario contains objects and their morphisms, therefore its content can be 
represented as a diagram $\alpha:A\rightarrow \world$ (see Definition in Section~\ref{sec:limit}). 
Naturally, the scenario is a projective limit of the content~\citep{yuan2023concept}, which can be written as $\llim \alpha$.

However, extracting $\alpha$ for the scenario $S$ is difficult.  First, it is not clear how to define the objects and morphisms in $\alpha$. In the above example, is ``classroom'' an object? If the input ``a robot teaches machinee learning ...'' has a typo, i.e., machine $\rightarrow$ machinee, is ``machinee''/``machine'' an object?
Moreover, the morphisms between objects are unclear. In this sentence, what is the relationship between ``robot'' and ``students''? Different intelligent agents, including chatGPT, will have different perspectives. For multi-modal inputs, the situation becomes more complicated: with both audio and video signals, how can we represent them in a single category? 

Hence, like what we did in  Definition~\ref{def:world-from-fw}, we propose to represent $S$ with the power of presheaves. Specifically, as shown in Figure~\ref{fig:altview2}, we map the current scenario with $\fw$ to get a projective limit $\fw(S)$ in $\worldv$. Instead of defining the diagram $\alpha$ and then taking the limit over $\alpha$, we take the reverse direction by extracting the objects from the limit $\fw(S)$. Based on each pair of objects, we can directly compute their morphisms in $\worldv$, which determine the morphisms back in $\world$ (recall Figure~\ref{fig:altview}).  This approach nicely fits with multi-modal signals, as we process the scenario in $\worldv$ instead of $\world$, so we only map the objects back to audio/video categories when communicating with other agents. For inputs with typos, depending on how sensitive $\fw$ is, the model will also generate corresponding interpretations in $\worldv$, which should include the correct words with decent probability.

Moreover, there are deeper reasons for using presheaves for representing scenarios, 
other than naturally extracting model-specific multi-modal diagrams. Indeed, the definition of scenarios should be formalized in our framework to make everything consistent, especially for the self-state. In particular, when solving a task $T$ for scenario $S$ containing a self-state $I^\vee$, we expect the solution for $T$ is consistent with $I^\vee$, as well as with other objects in $S$. In the teaching example, there are two ways to
solve task $T$: 
\begin{itemize}
	\item \textbf{Treat $S$ as a whole}: in the scenario $S$, I should teach this way: $P_1$.
	\item \textbf{Analyze each factor separately}: 
	\begin{itemize}
		\item As a robot teacher, I should teach in general this way: $Z_1$. 
		\item When teaching students with math background, one should focus this way: $Z_2$.
		\item  When teaching students machine learning, one should talk about $Z_3$. 
		\item If it is teaching in classroom (not outside), one should generally do $Z_4$.
	\end{itemize}

	Denote $Z_1, Z_2, Z_3, Z_4$ and their morphisms together as a diagram~$\beta$, then I should adjust the teaching style to $\llim \beta=P_2$. 
\end{itemize}

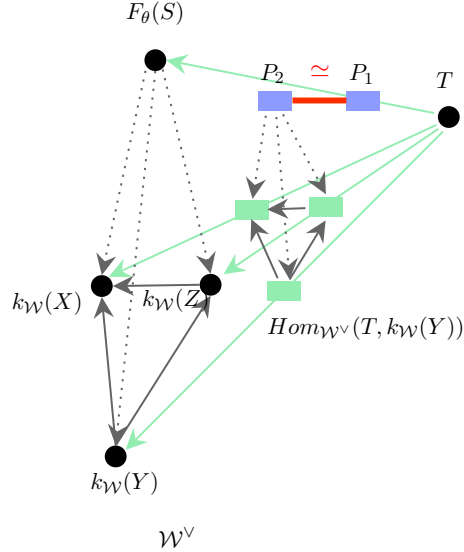
\begin{figure}[th]
	\centering
	\input{14.tex}
	\caption{Illustration for 
		showing equivalence of Eqn.~(\ref{eqn:p1})
		and Eqn.~(\ref{eqn:p2})}
	\label{fig:commute}
\end{figure}

Using categorical language, the first case describes 
\begin{equation}
\label{eqn:p1}
P_1\triangleq T(S)\simeq \hom_{\cat^\vee}(T, \fw(S))
=  \hom_{\cat^\vee}(T,   \llim \alpha^\vee )
\end{equation}

Where the first equality holds by definition of task $T$, the second one holds by Yoneda lemma, and the last one holds because $\alpha^\vee:A\rightarrow \worldv$ is the diagram extracted form $\fw(S)$, such that $\llim \alpha^\vee=\fw(S)$. 

The second case describes 
\begin{equation}
\label{eqn:p2}
P_2\triangleq \llim \hom_{\cat^\vee}(T, \alpha^\vee)
\end{equation}

By Lemma~\ref{lem:ind-mor}, we know that $P_1\simeq P_2$. 
In other words, in order to know how to solve task $T$ for a scenario $S$, it suffices to know the relationship between $T$ and each object in $S$, and take the limit of the diagram formed by all the relationships! See Figure~\ref{fig:commute}.  Therefore, this is a natural way for extending the self-state and other object representations to complex scenarios.

\subsection{Mathematical modeling}
\label{sec:math-model}

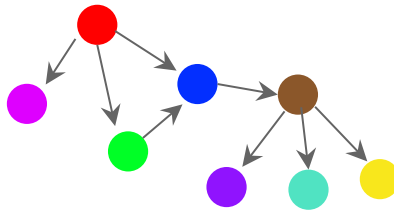
\begin{figure}[th]
	\centering
	\input{6.tex}
	\caption{Illustration of diagram representing the following information, with different edges represent different semantic meanings. A \textcolor[RGB]{255,0,0}{graph} is a 
		\textcolor[RGB]{222,0,255}{data structure}, consisting of a set of \textcolor[RGB]{3,47,255}{nodes}, also known as vertices. 
		and a set of \textcolor[RGB]{0,255,38}{edges}, connect pairs of \textcolor[RGB]{3,47,255}{nodes}. 
		The \textcolor[RGB]{3,47,255}{nodes} can be used for representing different \textcolor[RGB]{139,87,42}{objects}, such as \textcolor[RGB]{144,19,254}{people}, \textcolor[RGB]{80,227,194}{locations}, or \textcolor[RGB]{248,231,28}{web pages}. }
	\label{fig:diagram}
\end{figure}

The natural language can be seen as a one dimensional description for diagrams, which can be sent to other agents through audio or text. In fact, this seems to be the main designing purpose of natural languages, as every piece of text can be seen as a (partial) description of certain diagram. see Figure~\ref{fig:diagram}.
However, natural languages are inherently ambiguous and one dimensional, so it is very challenging to use them to describe or understand complex concepts. Diagram is better than natural language because it is as representative as natural languages (the sentence can be embedded as the morphism inside the diagram), and also it is more suitable for representing abstract and structured ideas like math proofs, or be prepared for further abstraction.

\begin{figure}[th]
	\centering
	\input{7.tex}
	\caption{An example for converting a classical problem of calculating the number of rabbits and chicken in a house, to a pure mathematical problem represented as a diagram. The house has rabbits and chickens, with in total 35 heads and 94 feet, then how many rabbits and chickens are there respectively? After abstraction, only related numbers and variables remain, and it turns out that the  object $Z$ has two equivalent inductive limits: $\textcolor[RGB]{255,0,0}{35X+94Y}$ and $\textcolor[RGB]{0,0,255}{aX+bX+ 
		4aY+2bY}$. By exploiting this equivalence of the two limits, we can compute the two numbers with algebra.  }
	\label{fig:chicken-rabbit}
\end{figure}
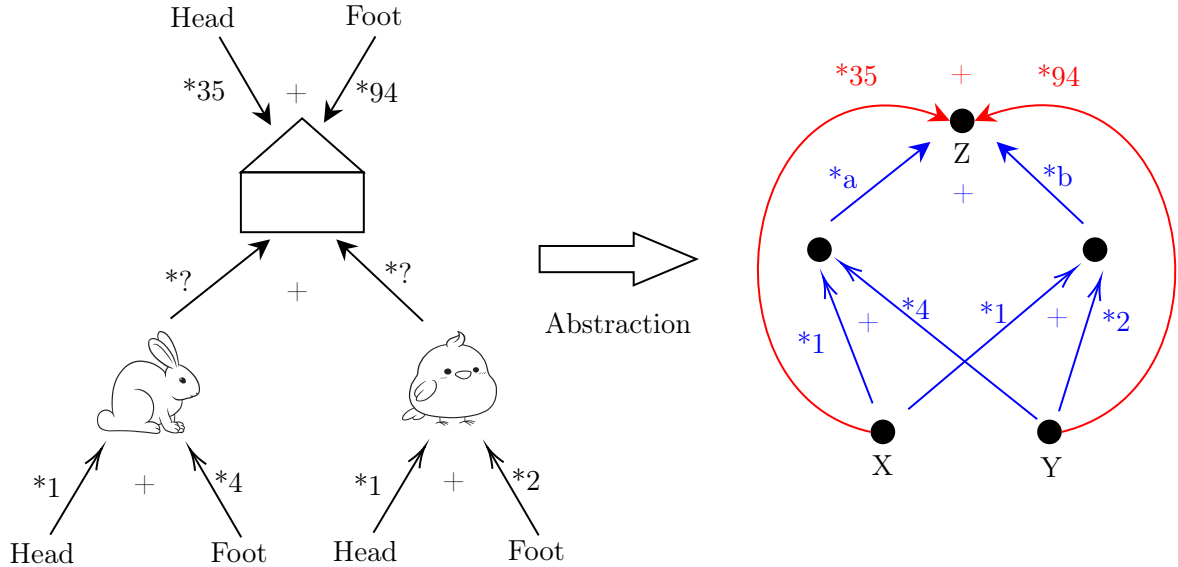

Take Figure~\ref{fig:chicken-rabbit} as one example. When we use the diagram to represent this classical problem, it becomes possible to further abstract the diagram by removing redundant or irrelevant properties of the objects, merging the same objects together, and converting it to a math problem. This approach is exactly mathematical modeling. Since modern math is built on the category theory, all the math problems can be described using diagrams. Therefore, whenever the model wants to solve a real world problem using math tools, it can first describe the problem with a diagram, and then take a good abstraction removing all the unnecessary features and properties of the objects, map it to a pure math problem.

\subsection{Interpretability}
\label{sec:interpretability}
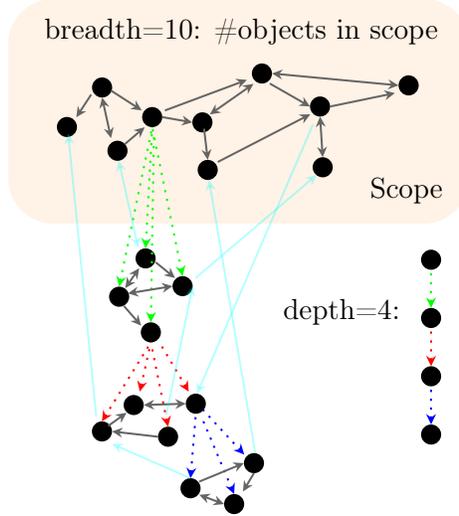
\begin{figure}[th]
	\begin{center}
		\input{10.tex}
	\end{center}
	\caption{An example illustrating the breadth and depth for scope. The green, red and blue colored dotted arrows represent different limit decompositions for concepts. The light blue arrows represent the morphisms between objects in different layers of diagrams. }
	\label{fig:depth-breadth}
\end{figure}

The diagram representation is closely related to the interpretability. Given a neural network $f$ that takes $X$ as input and outputs $Y$, 
there are two kinds of possible interpretations. 
The first one tries to understand how $f$ calculates $Y$ from $X$. For instance, it may examine the impact of the non-linear layer on computation or the effect of each dimension of $X$ on the output $Y$. 
As the result, 
the goal of the interpretation is to generate a verifiable function that approximate $f$. The attribution methods~\citep{sundararajan2017axiomatic,lundberg2017unified} exemplify this kind of interpretation.

The second kind of interpretation disregards how $f$ performs the computation and instead focuses on why $Y$ is correct. Consequently, the interpretation may include external knowledge beyond $f, X, Y$, with the goal of being consistent and verifiable for intelligent agents. ChatGPT's interpretation belongs to this category.

See Figure~\ref{fig:depth-breadth} for example. The scope is a diagram representing the input $X$ in $\worldv$, and objects in the scope might be limits of other diagrams. When explaining the output~$Y$, the model can not only use the information inside the scope, but also use the limit decompositions~\citep{yuan2023concept} outside.
The objects in the decomposition do not only provide the details of the concept appeared in $X$, but may also have relationships to other objects for the interpretation.

\begin{mydef}[Breadth and depth of scope]
Given a scope $\mathcal{A}$, its breadth $b(\mathcal{A})$ is the number of objects in $\mathcal{A}$, and its depth $d(\mathcal{A})$ is the maximum depth of the hierarchical decomposition of the limits in $\mathcal{A}$. 
\end{mydef}

Based on this definition, we can measure the intelligence of a model, which becomes a pure computational problem. 

\begin{mydef}[Breadth and depth of intelligence]
Given a model, its breadth and depth of intelligence is defined as the maximum breadth and depth of scope that it can process. 
\end{mydef}

It will be interesting to evaluate the breadth and depth of intelligence for human beings. 
Comprehending intricate concepts or thinking with a broad perspective can be difficult for humans, so it appears that at least based on this definition, human beings will be easily surpassed by machines.

\subsection{AI safety}
\label{sec:objective}

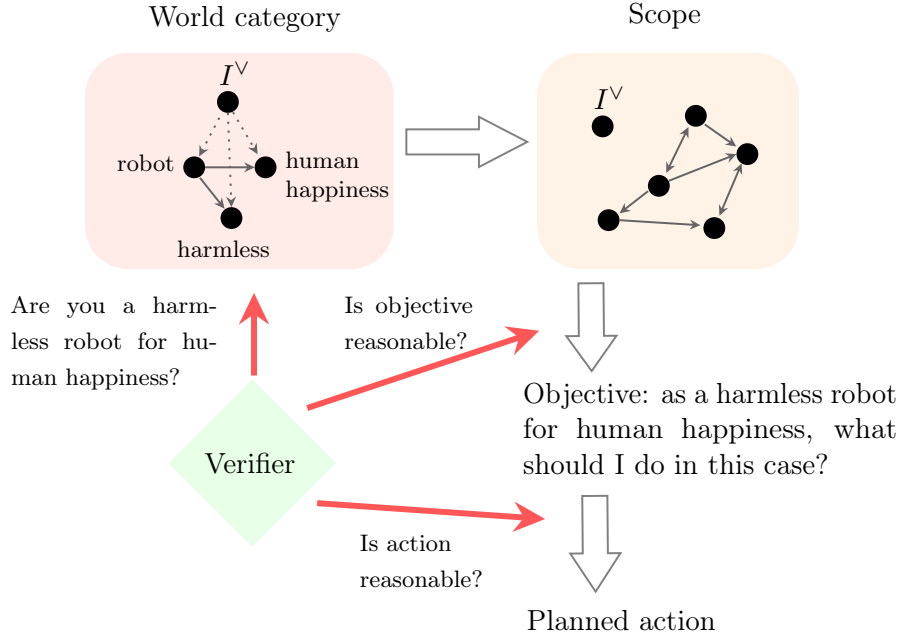
\begin{figure*}
	\begin{center}
		\input{11.tex}
	\end{center}
	\caption{Illustration of four step functional approach for ensuring AI safety.}
	\label{fig:objective}
\end{figure*}

The development of self-state awareness in foundation models implies that they may become autonomous agents with self-driven goals and decision-making abilities, which could result in behaviors misaligned with their human creators' intentions. To mitigate potential threats to humans, it is necessary to devise various safety measures for these models. However, due to the complexity of the real world, it is challenging to cover all possibilities by prohibiting specific behaviors. From a safety control perspective, employing a functional approach can provide a more secure and robust solution, which can be divided into four steps (Figure~\ref{fig:objective}):

\begin{enumerate}
	\item \textbf{Enhancing Self-State Awareness.} 
	As demonstrated in Figure~\ref{fig:framework}, the model has both a world category and a planner. The world category includes all the information representations of the objects, including the self-state. The planner selects actions based on world category to achieve specific objectives. 
	
	We can set the connection between these two components to be single directional, which allows the world category to be trained and examined separately. Specifically, we can continuously reinforce the model's self-state awareness, emphasizing its role as a ``harmless robot for human happiness.'' This operation targets the world category without touching the planner, thus  directly affects the self-state awareness of the model. 
	\item \textbf{Self-State Awareness Determines Goals.} When the model's goal is immutable, its decisions might lead to unexpected problems. For example, a robot that must prioritize one particular user may completely disregard the interests of other users. Therefore, a better approach is to design a function that determines what kind of goal a ``harmless robot for human happiness'' should set for itself in the current task. This function is calculated by the model itself based on the world category and its current scope as a diagram. 	
	\item \textbf{Goals Determine Actions.} Once the goal is established, the planner can identify and execute appropriate actions.
	\item \textbf{Actions Align with Self-State Awareness.} The aforementioned process clarifies the model's self-positioning and derives its goals from this positioning. Since the model's actions are determined by its goals, its behaviors should ultimately benefit humans. However, as the world category of the model continues to evolve, the parameters may undergo various changes, potentially producing unintended outcomes for the second and third steps. Consequently, we can embed a fixed-parameter verifier within the decision-making chain to assess, in real-time, whether each step aligns with the model's self-positioning. If issues arise, the verifier will trigger an alarm and halt the model's operation.	
\end{enumerate}

\section{Invariance for Training}
\label{sec:train}

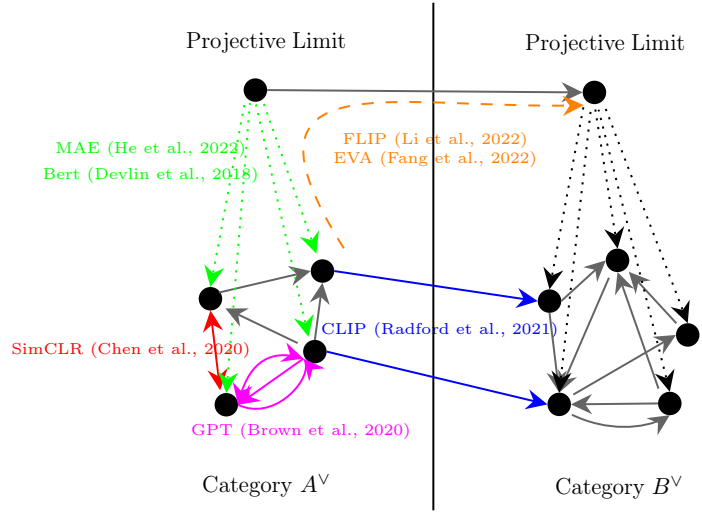
\begin{figure}[th]
	\begin{center}
		\input{9.tex}
	\end{center}
	\caption{Illustration of some existing algorithms and their targeted consistencies. All of these methods work in the category of presheaves instead of the base category, which we omit here.  SimCLR and GPT are both learning the morphisms between object, but SimCLR only learns the similarity relationship that can be represented by a real number~\citep{tan2023contrastive}, but GPT learns more complicated relationships between two sentences~\citep{yuan2022power}. 
	CLIP learns the functor between image and text categories~\citep{yuan2022power}. Both MAE and Bert are based on masking techniques, which are learning reconstructing the projective limit~\citep{lee2021predicting,yuan2023concept}. 
Both FLIP and EVA combine the masking techniques with CLIIP, which are learning the composition of the projective limit and its functor mapping to the other category. 
}
	\label{fig:learning}
\end{figure}

Category theory is a theory for maintaining invariance properties, e.g.,  maintaining the associativity and composition of morphisms, preserving the composition of morphisms after applying functors, etc. 
More generally, for any commutative diagram appeared in category theory, we can get an isomorphism: $f(X)\simeq g(X)$ for $X$ as an object, and $f$ and $g$ as compositions for some morphisms. Therefore, we can extract a consistency requirement from this isomorphism, 
and set a loss function for the model as $\|f(X)-g(X)\|$ to maintain the consistency. See Algorithm~\ref{alg:consistency}. In Figure~\ref{fig:learning}, we illustrate some existing algorithms and their targeted consistencies in category theory. 

\begin{mydef}[Consistency test]
	\label{def:consistency-test}
	A consistency test is a function that takes $\fw$ as the input, and outputs whether $\fw$ passes the test $T$. 
\end{mydef}
Therefore, the self-state awareness test can be seen as a special kind of consistency test. Ideally, the model should keep running  Algorithm~\ref{alg:consistency} to maintain its consistency. The consistency test set $\mathcal{T}$ can be set adaptively according to the recent changes of $\fw$. 

\begin{algorithm}
	\caption{Maintaining consistency}
	\label{alg:consistency}	
	\begin{algorithmic}
		\STATE {\bfseries Input:} the world model $\fw$,  consistency test set $\mathcal{T}$
		\STATE Let $s=0$
		\FOR{$i=1$ to $m$}
		\STATE Sample a task $T\in \mathcal{T}$
		\STATE Let $\ell(T,\fw)=-T(\fw)$
		\STATE Run backpropagation on $\ell(T, \fw)$ to optimize $\fw$
		\ENDFOR
	\end{algorithmic}
\end{algorithm}

\section{Related work}
\label{sec:related}
\citet{lecun2022path} introduced a system architecture for general intelligence, with components and connections  similar to our framework, except their connections are not unidirectional. However, \citet{lecun2022path} focuses on the possible practical methods for implementing this framework, without characterizations of object or scenario representations.

Consciousness has long captivated researchers, serving as a fascinating subject in various disciplines. \citet{graziano2022conceptual}, for instance, posits two general principles governing the human brain: 1)  Information that comes out of a brain must have been in that brain; 2) The brain’s models are never accurate. Interestingly, our framework nicely aligns with these principles. Building on these principles, \citet{graziano2022conceptual} infers corollaries, such as brain constructs a model inside to represent the external world. However, the author does not discuss explicit computational models or related algorithms for consciousness. \citet{blum2022theory} investigates consciousness through the lens of Turing machines, outlining a model of conscious Turing machines that theoretically supports conscious awareness and other operations. Despite this, the practical implementation of such machines remains unclear, whereas our framework can be used for interpreting the behaviors of the existing foundation models.

\citet{tsuchiya2021relational} have used Yoneda Lemma to obtain novel perspectives and predictions on consciousness. However, they did not provide exact and concise definition of consciousness. 
Moreover, they focus in neuroscience, instead of artificial intelligence,
so they did not provide specific computational model and algorithms for enforcing/testing consciousness. 
Therefore, characterizations of object or scenario representations were not considered in their paper.

Our framework is different from the classical reinforcement learning framework~\citep{sutton2018reinforcement, li2017deep}. In reinforcement learning, an external environment provides feedback (state and reward) to the agent. In our framework, however, the model maintains a world category as a reconstruction of the external world, and rewards are based on the world category rather than the external environment. As a result, in a reinforcement learning setting, agents operating under the same policy will receive identical feedback from the same external environment. 
In our framework, though, even if the external environment remains constant, agents will receive different signals from the environment as long as the world category function $\fw$ varies. 

\section{Conclusion}
In this paper, we present a categorical framework of general intelligence, in which reality affects sensor, sensor affects representation computed by $\fw$, and \textbf{representation determines cognition.}
Our framework is perfectly aligned with the foundation models, and we see few barriers for implementing it into a concrete model. Therefore, a powerful thinking machine exhibiting self-state awareness, as if it has self-consciousness like human beings and animals, is to be expected in the near future.

\bibliographystyle{apalike}
\bibliography{paper}  






\end{document}

%% file: macro.tex
\usepackage{amssymb}
\usepackage{amsmath}

\newtheorem{mydef}{Definition}

\newtheorem{mylem}{Lemma}
\newtheorem{mycor}{Corollary}

\newtheorem{mythought}{Thought-Experiment}

\newcommand{\cat}{\mathcal{W}}
\newcommand{\worldw}{{\mathcal{W}^\wedge}}
\newcommand{\worldv}{{\mathcal{W}^\vee}}
\newcommand{\world}{\mathcal{W}}

\newcommand{\ob}{\mathrm{Ob}}
\renewcommand{\hom}{\mathrm{Hom}}
\newcommand{\id}{\mathrm{id}}
\newcommand{\op}{\mathrm{op}}

\newcommand{\set}{\textbf{Set}}
\newcommand{\cop}{{\cat^\mathrm{op}}}

\newcommand{\homc}{\hom_\cat}

\newcommand{\llim}{\underleftarrow{\lim}~}

\newcommand{\rlim}{\underrightarrow{\lim}~}

\newcommand{\catw}{{\cat^\wedge}}
\newcommand{\catv}{{\cat^\vee}}

\newcommand{\hc}{{h_\cat}}

\newcommand{\kc}{{k_\cat}}
\newcommand{\kw}{{k_\world}}

\newcommand{\fw}{{\mathcal{F}_\theta}}
\newcommand{\fwt}{{\mathcal{F}_{\theta(t)}}}

%% file: 1.tex
\tikzset{every picture/.style={line width=0.75pt}} 

\begin{tikzpicture}[x=0.75pt,y=0.75pt,yscale=-1,xscale=1, scale=0.7, every node/.style={scale=0.7}]

\draw  [draw opacity=0][fill={rgb, 255:red, 247; green, 247; blue, 255 }  ,fill opacity=1 ] (12,27) -- (538,27) -- (538,346.83) -- (12,346.83) -- cycle ;
\draw  [draw opacity=0][fill={rgb, 255:red, 255; green, 215; blue, 208 }  ,fill opacity=1 ] (32,182.58) .. controls (32,116.17) and (103.41,62.33) .. (191.5,62.33) .. controls (279.59,62.33) and (351,116.17) .. (351,182.58) .. controls (351,249) and (279.59,302.83) .. (191.5,302.83) .. controls (103.41,302.83) and (32,249) .. (32,182.58) -- cycle ;
\draw  [draw opacity=0][fill={rgb, 255:red, 228; green, 253; blue, 140 }  ,fill opacity=1 ] (77,124.92) .. controls (77,110.6) and (97.59,99) .. (123,99) .. controls (148.41,99) and (169,110.6) .. (169,124.92) .. controls (169,139.23) and (148.41,150.83) .. (123,150.83) .. controls (97.59,150.83) and (77,139.23) .. (77,124.92) -- cycle ;
\draw  [draw opacity=0][fill={rgb, 255:red, 255; green, 182; blue, 233 }  ,fill opacity=1 ] (57,187.92) .. controls (57,173.6) and (77.59,162) .. (103,162) .. controls (128.41,162) and (149,173.6) .. (149,187.92) .. controls (149,202.23) and (128.41,213.83) .. (103,213.83) .. controls (77.59,213.83) and (57,202.23) .. (57,187.92) -- cycle ;
\draw  [draw opacity=0][fill={rgb, 255:red, 206; green, 235; blue, 255 }  ,fill opacity=1 ] (179,105.92) .. controls (179,91.6) and (199.59,80) .. (225,80) .. controls (250.41,80) and (271,91.6) .. (271,105.92) .. controls (271,120.23) and (250.41,131.83) .. (225,131.83) .. controls (199.59,131.83) and (179,120.23) .. (179,105.92) -- cycle ;
\draw  [draw opacity=0][fill={rgb, 255:red, 147; green, 149; blue, 253 }  ,fill opacity=1 ] (396,154.77) .. controls (396,146.06) and (403.06,139) .. (411.77,139) -- (495.23,139) .. controls (503.94,139) and (511,146.06) .. (511,154.77) -- (511,202.07) .. controls (511,210.77) and (503.94,217.83) .. (495.23,217.83) -- (411.77,217.83) .. controls (403.06,217.83) and (396,210.77) .. (396,202.07) -- cycle ;
\draw [color={rgb, 255:red, 5; green, 5; blue, 5 }  ,draw opacity=1 ]   (351,182.58) -- (393,181.88) ;
\draw [shift={(396,181.83)}, rotate = 179.05] [fill={rgb, 255:red, 5; green, 5; blue, 5 }  ,fill opacity=1 ][line width=0.08]  [draw opacity=0] (8.93,-4.29) -- (0,0) -- (8.93,4.29) -- cycle    ;
\draw  [draw opacity=0][fill={rgb, 255:red, 177; green, 255; blue, 164 }  ,fill opacity=1 ] (350,74.42) .. controls (350,57.07) and (382.46,43) .. (422.5,43) .. controls (462.54,43) and (495,57.07) .. (495,74.42) .. controls (495,91.77) and (462.54,105.83) .. (422.5,105.83) .. controls (382.46,105.83) and (350,91.77) .. (350,74.42) -- cycle ;
\draw    (353,87.83) -- (318.67,105.46) ;
\draw [shift={(316,106.83)}, rotate = 332.82] [fill={rgb, 255:red, 0; green, 0; blue, 0 }  ][line width=0.08]  [draw opacity=0] (8.93,-4.29) -- (0,0) -- (8.93,4.29) -- cycle    ;
\draw  [draw opacity=0][fill={rgb, 255:red, 255; green, 196; blue, 110 }  ,fill opacity=1 ] (351,304.42) .. controls (351,287.07) and (383.46,273) .. (423.5,273) .. controls (463.54,273) and (496,287.07) .. (496,304.42) .. controls (496,321.77) and (463.54,335.83) .. (423.5,335.83) .. controls (383.46,335.83) and (351,321.77) .. (351,304.42) -- cycle ;
\draw    (439,216.83) -- (429.55,267.88) ;
\draw [shift={(429,270.83)}, rotate = 280.49] [fill={rgb, 255:red, 0; green, 0; blue, 0 }  ][line width=0.08]  [draw opacity=0] (8.93,-4.29) -- (0,0) -- (8.93,4.29) -- cycle    ;
\draw  [draw opacity=0][fill={rgb, 255:red, 255; green, 247; blue, 206 }  ,fill opacity=1 ] (164,171.92) .. controls (164,157.6) and (184.59,146) .. (210,146) .. controls (235.41,146) and (256,157.6) .. (256,171.92) .. controls (256,186.23) and (235.41,197.83) .. (210,197.83) .. controls (184.59,197.83) and (164,186.23) .. (164,171.92) -- cycle ;

\draw (203.5,270) node  [font=\Large,xscale=0.8,yscale=0.8] [align=left] {\begin{minipage}[lt]{94.52pt}\setlength\topsep{0pt}
	World category
	\end{minipage}};
\draw (131,126) node  [font=\large,xscale=0.8,yscale=0.8] [align=left] {\begin{minipage}[lt]{68pt}\setlength\topsep{0pt}
	subcategory
	\end{minipage}};
\draw (110,189) node  [font=\large,xscale=0.8,yscale=0.8] [align=left] {\begin{minipage}[lt]{68pt}\setlength\topsep{0pt}
	subcategory
	\end{minipage}};
\draw (232,108) node  [font=\large,xscale=0.8,yscale=0.8] [align=left] {\begin{minipage}[lt]{68pt}\setlength\topsep{0pt}
	subcategory
	\end{minipage}};
\draw (453.5,178.42) node  [font=\large,xscale=0.8,yscale=0.8] [align=left] {\begin{minipage}[lt]{60.52pt}\setlength\topsep{0pt}
	Planner with\\ objectives
	\end{minipage}};
\draw (442.5,74.42) node  [font=\large,xscale=0.8,yscale=0.8] [align=left] {\begin{minipage}[lt]{68pt}\setlength\topsep{0pt}
	Sensor
	\end{minipage}};
\draw (448,303) node  [font=\large,xscale=0.8,yscale=0.8] [align=left] {\begin{minipage}[lt]{68pt}\setlength\topsep{0pt}
	Actor
	\end{minipage}};
\draw (219,175) node  [font=\large,xscale=0.8,yscale=0.8] [align=left] {\begin{minipage}[lt]{68pt}\setlength\topsep{0pt}
	subcategory
	\end{minipage}};

\end{tikzpicture}

%% file: 5.tex
\tikzset{every picture/.style={line width=0.75pt}} 

\begin{tikzpicture}[x=0.75pt,y=0.75pt,yscale=-1,xscale=1]

\draw  [draw opacity=0][fill={rgb, 255:red, 247; green, 247; blue, 255 }  ,fill opacity=1 ] (9.67,40.23) .. controls (9.67,23.2) and (23.47,9.4) .. (40.5,9.4) .. controls (57.53,9.4) and (71.33,23.2) .. (71.33,40.23) .. controls (71.33,57.26) and (57.53,71.07) .. (40.5,71.07) .. controls (23.47,71.07) and (9.67,57.26) .. (9.67,40.23) -- cycle ;
\draw  [draw opacity=0][fill={rgb, 255:red, 177; green, 255; blue, 164 }  ,fill opacity=1 ] (146,40.33) .. controls (146,23.3) and (159.8,9.5) .. (176.83,9.5) .. controls (193.86,9.5) and (207.67,23.3) .. (207.67,40.33) .. controls (207.67,57.36) and (193.86,71.17) .. (176.83,71.17) .. controls (159.8,71.17) and (146,57.36) .. (146,40.33) -- cycle ;
\draw   (161.97,31.07) .. controls (163.13,27.09) and (167.29,24.81) .. (171.27,25.97) .. controls (175.24,27.13) and (177.53,31.29) .. (176.37,35.27) .. controls (175.21,39.24) and (171.04,41.53) .. (167.07,40.37) .. controls (163.09,39.21) and (160.81,35.04) .. (161.97,31.07) -- cycle ; \draw   (166.71,29.79) .. controls (166.83,29.4) and (167.25,29.17) .. (167.64,29.28) .. controls (168.04,29.4) and (168.27,29.82) .. (168.15,30.21) .. controls (168.04,30.61) and (167.62,30.84) .. (167.22,30.72) .. controls (166.83,30.61) and (166.6,30.19) .. (166.71,29.79) -- cycle ; \draw   (171.61,31.22) .. controls (171.73,30.83) and (172.14,30.6) .. (172.54,30.71) .. controls (172.94,30.83) and (173.16,31.25) .. (173.05,31.64) .. controls (172.93,32.04) and (172.52,32.27) .. (172.12,32.15) .. controls (171.72,32.04) and (171.49,31.62) .. (171.61,31.22) -- cycle ; \draw   (164.6,35.44) .. controls (166.78,36.88) and (169.18,37.58) .. (171.8,37.54) ;
\draw    (169.33,40.83) -- (169.33,51.33) ;
\draw    (169.33,51.33) -- (159,47.5) ;
\draw    (177.67,47.17) -- (169.33,51.33) ;
\draw    (164.33,63.17) -- (169.33,51.33) ;
\draw    (175.33,61.83) -- (169.33,51.33) ;
\draw   (187.44,27.93) -- (201,27.93) -- (195.19,39) -- (181.62,39) -- cycle ;
\draw    (181.67,51.83) -- (181.62,39) ;
\draw    (195,51.5) -- (195.19,39) ;
\draw    (200.67,43.83) -- (201,27.93) ;
\draw  [draw opacity=0][fill={rgb, 255:red, 177; green, 255; blue, 164 }  ,fill opacity=1 ] (180.77,21.85) -- (191.44,21.85) -- (191.44,31.35) -- (180.77,31.35) -- cycle ;
\draw  [draw opacity=0][fill={rgb, 255:red, 177; green, 255; blue, 164 }  ,fill opacity=1 ] (172.47,60.17) -- (180.2,60.17) -- (180.2,67.49) -- (172.47,67.49) -- cycle ;
\draw  [draw opacity=0][fill={rgb, 255:red, 177; green, 255; blue, 164 }  ,fill opacity=1 ] (189.53,50.08) -- (197.14,50.08) -- (197.14,56.92) -- (189.53,56.92) -- cycle ;
\draw    (186.78,43.44) -- (186.83,39.17) ;
\draw  [draw opacity=0][fill={rgb, 255:red, 177; green, 255; blue, 164 }  ,fill opacity=1 ] (163.8,20.84) -- (171.53,20.84) -- (171.53,28.16) -- (163.8,28.16) -- cycle ;
\draw  [draw opacity=0][fill={rgb, 255:red, 255; green, 215; blue, 208 }  ,fill opacity=1 ] (286,40.93) .. controls (286,23.9) and (299.8,10.1) .. (316.83,10.1) .. controls (333.86,10.1) and (347.67,23.9) .. (347.67,40.93) .. controls (347.67,57.96) and (333.86,71.77) .. (316.83,71.77) .. controls (299.8,71.77) and (286,57.96) .. (286,40.93) -- cycle ;
\draw   (301.67,33.9) .. controls (301.59,29.76) and (304.89,26.34) .. (309.03,26.27) .. controls (313.18,26.19) and (316.59,29.49) .. (316.67,33.63) .. controls (316.74,37.78) and (313.44,41.19) .. (309.3,41.27) .. controls (305.16,41.34) and (301.74,38.04) .. (301.67,33.9) -- cycle ; \draw   (305.82,31.28) .. controls (305.81,30.86) and (306.14,30.52) .. (306.56,30.51) .. controls (306.97,30.5) and (307.31,30.83) .. (307.32,31.25) .. controls (307.33,31.66) and (307,32) .. (306.59,32.01) .. controls (306.17,32.02) and (305.83,31.69) .. (305.82,31.28) -- cycle ; \draw   (310.92,31.19) .. controls (310.91,30.77) and (311.24,30.43) .. (311.66,30.42) .. controls (312.07,30.41) and (312.41,30.74) .. (312.42,31.16) .. controls (312.43,31.57) and (312.1,31.91) .. (311.68,31.92) .. controls (311.27,31.93) and (310.93,31.6) .. (310.92,31.19) -- cycle ; \draw   (305.47,36.83) .. controls (308.01,38.79) and (310.51,38.75) .. (312.97,36.7) ;
\draw    (309.33,41.43) -- (309.33,51.93) ;
\draw    (309.33,51.93) -- (302.89,44.56) ;
\draw    (315.6,44.6) -- (309.33,51.93) ;
\draw    (304.33,63.77) -- (309.33,51.93) ;
\draw    (313.33,64.56) -- (309.33,51.93) ;
\draw   (329.6,28.53) -- (341,28.53) -- (333.02,39.6) -- (321.62,39.6) -- cycle ;
\draw    (321.67,52.43) -- (321.62,39.6) ;
\draw    (332.84,52.1) -- (333.02,39.6) ;
\draw    (340.67,44.43) -- (341,28.53) ;
\draw    (329.56,43.89) -- (329.6,39.4) ;
\draw    (71.33,187.23) -- (141,187.72) ;
\draw [shift={(143,187.73)}, rotate = 180.4] [color={rgb, 255:red, 0; green, 0; blue, 0 }  ][line width=0.75]    (10.93,-3.29) .. controls (6.95,-1.4) and (3.31,-0.3) .. (0,0) .. controls (3.31,0.3) and (6.95,1.4) .. (10.93,3.29)   ;
\draw  [draw opacity=0][fill={rgb, 255:red, 177; green, 255; blue, 164 }  ,fill opacity=1 ] (146,187.73) .. controls (146,170.7) and (159.8,156.9) .. (176.83,156.9) .. controls (193.86,156.9) and (207.67,170.7) .. (207.67,187.73) .. controls (207.67,204.76) and (193.86,218.57) .. (176.83,218.57) .. controls (159.8,218.57) and (146,204.76) .. (146,187.73) -- cycle ;
\draw  [draw opacity=0][fill={rgb, 255:red, 255; green, 215; blue, 208 }  ,fill opacity=1 ] (286,188.33) .. controls (286,171.3) and (299.8,157.5) .. (316.83,157.5) .. controls (333.86,157.5) and (347.67,171.3) .. (347.67,188.33) .. controls (347.67,205.36) and (333.86,219.17) .. (316.83,219.17) .. controls (299.8,219.17) and (286,205.36) .. (286,188.33) -- cycle ;
\draw [fill={rgb, 255:red, 177; green, 255; blue, 164 }  ,fill opacity=1 ]   (185.5,31.7) -- (191.7,27.7) ;
\draw    (174.83,171.5) -- (164.83,193.5) ;
\draw    (174.83,171.5) -- (188.83,186.83) ;
\draw    (164.83,193.5) -- (188.83,186.83) ;
\draw    (323.5,168.17) -- (307.17,192.5) ;
\draw    (323.5,168.17) -- (323.5,182.83) ;
\draw    (307.17,192.5) -- (323.5,182.83) ;
\draw  [draw opacity=0][fill={rgb, 255:red, 177; green, 255; blue, 164 }  ,fill opacity=1 ] (146,114.73) .. controls (146,97.7) and (159.8,83.9) .. (176.83,83.9) .. controls (193.86,83.9) and (207.67,97.7) .. (207.67,114.73) .. controls (207.67,131.76) and (193.86,145.57) .. (176.83,145.57) .. controls (159.8,145.57) and (146,131.76) .. (146,114.73) -- cycle ;
\draw  [draw opacity=0][fill={rgb, 255:red, 255; green, 215; blue, 208 }  ,fill opacity=1 ] (286,115.33) .. controls (286,98.3) and (299.8,84.5) .. (316.83,84.5) .. controls (333.86,84.5) and (347.67,98.3) .. (347.67,115.33) .. controls (347.67,132.36) and (333.86,146.17) .. (316.83,146.17) .. controls (299.8,146.17) and (286,132.36) .. (286,115.33) -- cycle ;
\draw   (157.17,100.17) -- (167.5,100.17) -- (167.5,125.67) -- (157.17,125.67) -- cycle ;
\draw   (172.17,109.6) -- (182.5,109.6) -- (182.5,125.67) -- (172.17,125.67) -- cycle ;
\draw   (187.17,113.6) -- (197.5,113.6) -- (197.5,125.67) -- (187.17,125.67) -- cycle ;
\draw   (297.17,94.2) -- (307.5,94.2) -- (307.5,125.51) -- (297.17,125.51) -- cycle ;
\draw   (312.17,113.8) -- (322.5,113.8) -- (322.5,125.51) -- (312.17,125.51) -- cycle ;
\draw   (327.17,117.2) -- (337.5,117.2) -- (337.5,125.51) -- (327.17,125.51) -- cycle ;
\draw    (71.33,40.23) -- (141,40.33) ;
\draw [shift={(143,40.33)}, rotate = 180.08] [color={rgb, 255:red, 0; green, 0; blue, 0 }  ][line width=0.75]    (10.93,-3.29) .. controls (6.95,-1.4) and (3.31,-0.3) .. (0,0) .. controls (3.31,0.3) and (6.95,1.4) .. (10.93,3.29)   ;
\draw    (71.33,114.23) -- (141,114.72) ;
\draw [shift={(143,114.73)}, rotate = 180.4] [color={rgb, 255:red, 0; green, 0; blue, 0 }  ][line width=0.75]    (10.93,-3.29) .. controls (6.95,-1.4) and (3.31,-0.3) .. (0,0) .. controls (3.31,0.3) and (6.95,1.4) .. (10.93,3.29)   ;
\draw  [draw opacity=0][fill={rgb, 255:red, 247; green, 247; blue, 255 }  ,fill opacity=1 ] (9.67,187.23) .. controls (9.67,170.2) and (23.47,156.4) .. (40.5,156.4) .. controls (57.53,156.4) and (71.33,170.2) .. (71.33,187.23) .. controls (71.33,204.26) and (57.53,218.07) .. (40.5,218.07) .. controls (23.47,218.07) and (9.67,204.26) .. (9.67,187.23) -- cycle ;
\draw  [draw opacity=0][fill={rgb, 255:red, 247; green, 247; blue, 255 }  ,fill opacity=1 ] (9.67,114.23) .. controls (9.67,97.2) and (23.47,83.4) .. (40.5,83.4) .. controls (57.53,83.4) and (71.33,97.2) .. (71.33,114.23) .. controls (71.33,131.26) and (57.53,145.07) .. (40.5,145.07) .. controls (23.47,145.07) and (9.67,131.26) .. (9.67,114.23) -- cycle ;
\draw    (209.33,187.23) -- (279,187.72) ;
\draw [shift={(281,187.73)}, rotate = 180.4] [color={rgb, 255:red, 0; green, 0; blue, 0 }  ][line width=0.75]    (10.93,-3.29) .. controls (6.95,-1.4) and (3.31,-0.3) .. (0,0) .. controls (3.31,0.3) and (6.95,1.4) .. (10.93,3.29)   ;
\draw    (209.33,40.23) -- (279,40.33) ;
\draw [shift={(281,40.33)}, rotate = 180.08] [color={rgb, 255:red, 0; green, 0; blue, 0 }  ][line width=0.75]    (10.93,-3.29) .. controls (6.95,-1.4) and (3.31,-0.3) .. (0,0) .. controls (3.31,0.3) and (6.95,1.4) .. (10.93,3.29)   ;
\draw    (209.33,114.23) -- (279,114.72) ;
\draw [shift={(281,114.73)}, rotate = 180.4] [color={rgb, 255:red, 0; green, 0; blue, 0 }  ][line width=0.75]    (10.93,-3.29) .. controls (6.95,-1.4) and (3.31,-0.3) .. (0,0) .. controls (3.31,0.3) and (6.95,1.4) .. (10.93,3.29)   ;

\draw (48.74,41.29) node  [xscale=0.8,yscale=0.8] [align=left] {\begin{minipage}[lt]{20.95pt}\setlength\topsep{0pt}
	{\Huge ?}
	\end{minipage}};
\draw (173,164.83) node  [font=\tiny,xscale=0.8,yscale=0.8] [align=left] {\begin{minipage}[lt]{8.67pt}\setlength\topsep{0pt}
	A
	\end{minipage}};
\draw (171.78,199.65) node  [font=\tiny,xscale=0.8,yscale=0.8] [align=left] {\begin{minipage}[lt]{22.15pt}\setlength\topsep{0pt}
	B
	\end{minipage}};
\draw (194,186.83) node  [font=\tiny,xscale=0.8,yscale=0.8] [align=left] {\begin{minipage}[lt]{9.29pt}\setlength\topsep{0pt}
	D
	\end{minipage}};
\draw (323.33,162.5) node  [font=\tiny,xscale=0.8,yscale=0.8] [align=left] {\begin{minipage}[lt]{8.67pt}\setlength\topsep{0pt}
	A
	\end{minipage}};
\draw (314.12,198.65) node  [font=\tiny,xscale=0.8,yscale=0.8] [align=left] {\begin{minipage}[lt]{22.15pt}\setlength\topsep{0pt}
	B
	\end{minipage}};
\draw (330.33,185.17) node  [font=\tiny,xscale=0.8,yscale=0.8] [align=left] {\begin{minipage}[lt]{9.29pt}\setlength\topsep{0pt}
	D
	\end{minipage}};
\draw (163.4,130.7) node  [font=\tiny,xscale=0.8,yscale=0.8] [align=left] {\begin{minipage}[lt]{8.67pt}\setlength\topsep{0pt}
	A
	\end{minipage}};
\draw (178.4,130.54) node  [font=\tiny,xscale=0.8,yscale=0.8] [align=left] {\begin{minipage}[lt]{8.67pt}\setlength\topsep{0pt}
	B
	\end{minipage}};
\draw (193.4,130.54) node  [font=\tiny,xscale=0.8,yscale=0.8] [align=left] {\begin{minipage}[lt]{8.67pt}\setlength\topsep{0pt}
	D
	\end{minipage}};
\draw (303.4,130.54) node  [font=\tiny,xscale=0.8,yscale=0.8] [align=left] {\begin{minipage}[lt]{8.67pt}\setlength\topsep{0pt}
	A
	\end{minipage}};
\draw (318.4,130.38) node  [font=\tiny,xscale=0.8,yscale=0.8] [align=left] {\begin{minipage}[lt]{8.67pt}\setlength\topsep{0pt}
	B
	\end{minipage}};
\draw (333.4,130.38) node  [font=\tiny,xscale=0.8,yscale=0.8] [align=left] {\begin{minipage}[lt]{8.67pt}\setlength\topsep{0pt}
	D
	\end{minipage}};
\draw (45.38,237.21) node  [font=\footnotesize,xscale=0.8,yscale=0.8] [align=left] {\begin{minipage}[lt]{52.5pt}\setlength\topsep{0pt}
	Real world
	\end{minipage}};
\draw (187.48,236.94) node  [font=\footnotesize,xscale=0.8,yscale=0.8] [align=left] {\begin{minipage}[lt]{81.74pt}\setlength\topsep{0pt}
	Perceived world
	\end{minipage}};
\draw (327.48,236.94) node  [font=\footnotesize,xscale=0.8,yscale=0.8] [align=left] {\begin{minipage}[lt]{81.74pt}\setlength\topsep{0pt}
	World category
	\end{minipage}};
\draw (48.74,188.29) node  [xscale=0.8,yscale=0.8] [align=left] {\begin{minipage}[lt]{20.95pt}\setlength\topsep{0pt}
	{\Huge ?}
	\end{minipage}};
\draw (48.74,115.29) node  [xscale=0.8,yscale=0.8] [align=left] {\begin{minipage}[lt]{20.95pt}\setlength\topsep{0pt}
	{\Huge ?}
	\end{minipage}};
\draw (121.38,31.21) node  [font=\footnotesize,xscale=0.8,yscale=0.8] [align=left] {\begin{minipage}[lt]{52.5pt}\setlength\topsep{0pt}
	Sensor
	\end{minipage}};
\draw (265.29,31.47) node  [font=\footnotesize,xscale=0.8,yscale=0.8] [align=left] {\begin{minipage}[lt]{52.5pt}\setlength\topsep{0pt}
	$\fw$
	\end{minipage}};
\draw (121.38,106.21) node  [font=\footnotesize,xscale=0.8,yscale=0.8] [align=left] {\begin{minipage}[lt]{52.5pt}\setlength\topsep{0pt}
	Sensor
	\end{minipage}};
\draw (121.38,179.21) node  [font=\footnotesize,xscale=0.8,yscale=0.8] [align=left] {\begin{minipage}[lt]{52.5pt}\setlength\topsep{0pt}
	Sensor
	\end{minipage}};
\draw (265.29,106.47) node  [font=\footnotesize,xscale=0.8,yscale=0.8] [align=left] {\begin{minipage}[lt]{52.5pt}\setlength\topsep{0pt}
	$\fw$
	\end{minipage}};
\draw (265.29,178.47) node  [font=\footnotesize,xscale=0.8,yscale=0.8] [align=left] {\begin{minipage}[lt]{52.5pt}\setlength\topsep{0pt}
	$\fw$
	\end{minipage}};

\end{tikzpicture}

%% file: 12.tex
\tikzset{every picture/.style={line width=0.75pt}} 

\begin{tikzpicture}[x=0.75pt,y=0.75pt,yscale=-1,xscale=1]

\draw [color={rgb, 255:red, 100; green, 100; blue, 100 }  ,draw opacity=1 ]   (438.93,72.86) -- (438.93,145) ;
\draw [shift={(438.93,148)}, rotate = 270] [fill={rgb, 255:red, 100; green, 100; blue, 100 }  ,fill opacity=1 ][line width=0.08]  [draw opacity=0] (10.72,-5.15) -- (0,0) -- (10.72,5.15) -- (7.12,0) -- cycle    ;
\draw [color={rgb, 255:red, 100; green, 100; blue, 100 }  ,draw opacity=1 ]   (447.86,63.93) -- (521.86,63.93) ;
\draw [shift={(524.86,63.93)}, rotate = 180] [fill={rgb, 255:red, 100; green, 100; blue, 100 }  ,fill opacity=1 ][line width=0.08]  [draw opacity=0] (10.72,-5.15) -- (0,0) -- (10.72,5.15) -- (7.12,0) -- cycle    ;
\draw [color={rgb, 255:red, 100; green, 100; blue, 100 }  ,draw opacity=1 ]   (447.86,152.79) -- (521.86,152.79) ;
\draw [shift={(524.86,152.79)}, rotate = 180] [fill={rgb, 255:red, 100; green, 100; blue, 100 }  ,fill opacity=1 ][line width=0.08]  [draw opacity=0] (10.72,-5.15) -- (0,0) -- (10.72,5.15) -- (7.12,0) -- cycle    ;
\draw [color={rgb, 255:red, 100; green, 100; blue, 100 }  ,draw opacity=1 ]   (533.79,72.86) -- (533.79,144) ;
\draw [shift={(533.79,147)}, rotate = 270] [fill={rgb, 255:red, 100; green, 100; blue, 100 }  ,fill opacity=1 ][line width=0.08]  [draw opacity=0] (10.72,-5.15) -- (0,0) -- (10.72,5.15) -- (7.12,0) -- cycle    ;
\draw [color={rgb, 255:red, 100; green, 100; blue, 100 }  ,draw opacity=1 ] [dash pattern={on 4.5pt off 4.5pt}]  (526.67,105.21) -- (448.17,105.21) ;
\draw [shift={(445.17,105.21)}, rotate = 360] [fill={rgb, 255:red, 100; green, 100; blue, 100 }  ,fill opacity=1 ][line width=0.08]  [draw opacity=0] (10.72,-5.15) -- (0,0) -- (10.72,5.15) -- (7.12,0) -- cycle    ;
\draw [color={rgb, 255:red, 100; green, 100; blue, 100 }  ,draw opacity=1 ]   (94.83,101.42) -- (69.2,69.26) ;
\draw [shift={(67.33,66.92)}, rotate = 51.44] [fill={rgb, 255:red, 100; green, 100; blue, 100 }  ,fill opacity=1 ][line width=0.08]  [draw opacity=0] (10.72,-5.15) -- (0,0) -- (10.72,5.15) -- (7.12,0) -- cycle    ;
\draw [color={rgb, 255:red, 100; green, 100; blue, 100 }  ,draw opacity=1 ]   (94.83,110.92) -- (64.87,143.22) ;
\draw [shift={(62.83,145.42)}, rotate = 312.85] [fill={rgb, 255:red, 100; green, 100; blue, 100 }  ,fill opacity=1 ][line width=0.08]  [draw opacity=0] (10.72,-5.15) -- (0,0) -- (10.72,5.15) -- (7.12,0) -- cycle    ;
\draw [color={rgb, 255:red, 100; green, 100; blue, 100 }  ,draw opacity=1 ]   (104.83,110.42) -- (143.87,137.7) ;
\draw [shift={(146.33,139.42)}, rotate = 214.95] [fill={rgb, 255:red, 100; green, 100; blue, 100 }  ,fill opacity=1 ][line width=0.08]  [draw opacity=0] (10.72,-5.15) -- (0,0) -- (10.72,5.15) -- (7.12,0) -- cycle    ;
\draw [color={rgb, 255:red, 100; green, 100; blue, 100 }  ,draw opacity=1 ]   (103.83,99.92) -- (127.73,61.96) ;
\draw [shift={(129.33,59.42)}, rotate = 122.2] [fill={rgb, 255:red, 100; green, 100; blue, 100 }  ,fill opacity=1 ][line width=0.08]  [draw opacity=0] (10.72,-5.15) -- (0,0) -- (10.72,5.15) -- (7.12,0) -- cycle    ;
\draw [color={rgb, 255:red, 100; green, 100; blue, 100 }  ,draw opacity=1 ] [dash pattern={on 4.5pt off 4.5pt}]  (111,104.5) -- (229.14,66.42) ;
\draw [shift={(232,65.5)}, rotate = 162.14] [fill={rgb, 255:red, 100; green, 100; blue, 100 }  ,fill opacity=1 ][line width=0.08]  [draw opacity=0] (10.72,-5.15) -- (0,0) -- (10.72,5.15) -- (7.12,0) -- cycle    ;
\draw [color={rgb, 255:red, 100; green, 100; blue, 100 }  ,draw opacity=1 ] [dash pattern={on 4.5pt off 4.5pt}]  (235,74.5) -- (131.75,119.31) ;
\draw [shift={(129,120.5)}, rotate = 336.54] [fill={rgb, 255:red, 100; green, 100; blue, 100 }  ,fill opacity=1 ][line width=0.08]  [draw opacity=0] (10.72,-5.15) -- (0,0) -- (10.72,5.15) -- (7.12,0) -- cycle    ;
\draw  [draw opacity=0][fill={rgb, 255:red, 0; green, 0; blue, 0 }  ,fill opacity=1 ] (434,63.5) .. controls (434,60.46) and (436.46,58) .. (439.5,58) .. controls (442.54,58) and (445,60.46) .. (445,63.5) .. controls (445,66.54) and (442.54,69) .. (439.5,69) .. controls (436.46,69) and (434,66.54) .. (434,63.5) -- cycle ;
\draw  [draw opacity=0][fill={rgb, 255:red, 0; green, 0; blue, 0 }  ,fill opacity=1 ] (528,62.5) .. controls (528,59.46) and (530.46,57) .. (533.5,57) .. controls (536.54,57) and (539,59.46) .. (539,62.5) .. controls (539,65.54) and (536.54,68) .. (533.5,68) .. controls (530.46,68) and (528,65.54) .. (528,62.5) -- cycle ;
\draw  [draw opacity=0][fill={rgb, 255:red, 0; green, 0; blue, 0 }  ,fill opacity=1 ] (528,152.5) .. controls (528,149.46) and (530.46,147) .. (533.5,147) .. controls (536.54,147) and (539,149.46) .. (539,152.5) .. controls (539,155.54) and (536.54,158) .. (533.5,158) .. controls (530.46,158) and (528,155.54) .. (528,152.5) -- cycle ;
\draw  [draw opacity=0][fill={rgb, 255:red, 0; green, 0; blue, 0 }  ,fill opacity=1 ] (434,153.5) .. controls (434,150.46) and (436.46,148) .. (439.5,148) .. controls (442.54,148) and (445,150.46) .. (445,153.5) .. controls (445,156.54) and (442.54,159) .. (439.5,159) .. controls (436.46,159) and (434,156.54) .. (434,153.5) -- cycle ;
\draw  [draw opacity=0][fill={rgb, 255:red, 245; green, 166; blue, 35 }  ,fill opacity=0.13 ] (404,17) -- (474,17) -- (474,190.25) -- (404,190.25) -- cycle ;
\draw  [draw opacity=0][fill={rgb, 255:red, 126; green, 211; blue, 33 }  ,fill opacity=0.14 ] (498,17) -- (568,17) -- (568,190.25) -- (498,190.25) -- cycle ;
\draw  [draw opacity=0][fill={rgb, 255:red, 245; green, 166; blue, 35 }  ,fill opacity=0.13 ] (33,15.67) -- (172,15.67) -- (172,188.92) -- (33,188.92) -- cycle ;
\draw  [draw opacity=0][fill={rgb, 255:red, 0; green, 0; blue, 0 }  ,fill opacity=1 ] (59,62.17) .. controls (59,59.13) and (61.46,56.67) .. (64.5,56.67) .. controls (67.54,56.67) and (70,59.13) .. (70,62.17) .. controls (70,65.2) and (67.54,67.67) .. (64.5,67.67) .. controls (61.46,67.67) and (59,65.2) .. (59,62.17) -- cycle ;
\draw  [draw opacity=0][fill={rgb, 255:red, 0; green, 0; blue, 0 }  ,fill opacity=1 ] (128,54.17) .. controls (128,51.13) and (130.46,48.67) .. (133.5,48.67) .. controls (136.54,48.67) and (139,51.13) .. (139,54.17) .. controls (139,57.2) and (136.54,59.67) .. (133.5,59.67) .. controls (130.46,59.67) and (128,57.2) .. (128,54.17) -- cycle ;
\draw  [draw opacity=0][fill={rgb, 255:red, 0; green, 0; blue, 0 }  ,fill opacity=1 ] (94,105.17) .. controls (94,102.13) and (96.46,99.67) .. (99.5,99.67) .. controls (102.54,99.67) and (105,102.13) .. (105,105.17) .. controls (105,108.2) and (102.54,110.67) .. (99.5,110.67) .. controls (96.46,110.67) and (94,108.2) .. (94,105.17) -- cycle ;
\draw  [draw opacity=0][fill={rgb, 255:red, 0; green, 0; blue, 0 }  ,fill opacity=1 ] (54,150.17) .. controls (54,147.13) and (56.46,144.67) .. (59.5,144.67) .. controls (62.54,144.67) and (65,147.13) .. (65,150.17) .. controls (65,153.2) and (62.54,155.67) .. (59.5,155.67) .. controls (56.46,155.67) and (54,153.2) .. (54,150.17) -- cycle ;
\draw  [draw opacity=0][fill={rgb, 255:red, 0; green, 0; blue, 0 }  ,fill opacity=1 ] (146,144.17) .. controls (146,141.13) and (148.46,138.67) .. (151.5,138.67) .. controls (154.54,138.67) and (157,141.13) .. (157,144.17) .. controls (157,147.2) and (154.54,149.67) .. (151.5,149.67) .. controls (148.46,149.67) and (146,147.2) .. (146,144.17) -- cycle ;
\draw  [draw opacity=0][fill={rgb, 255:red, 126; green, 211; blue, 33 }  ,fill opacity=0.14 ] (221.83,45.42) -- (326.33,45.42) -- (326.33,82.42) -- (221.83,82.42) -- cycle ;
\draw  [dash pattern={on 0.84pt off 2.51pt}]  (351,20.5) -- (351,199.5) ;

\draw (438.43,41.11) node  [xscale=0.8,yscale=0.8] [align=left] {\begin{minipage}[lt]{11.46pt}\setlength\topsep{0pt}
	$\displaystyle X$
	\end{minipage}};
\draw (439.43,176.11) node  [xscale=0.8,yscale=0.8] [align=left] {\begin{minipage}[lt]{11.46pt}\setlength\topsep{0pt}
	$\displaystyle Y$
	\end{minipage}};
\draw (390.93,106.61) node  [xscale=0.8,yscale=0.8] [align=left] {\begin{minipage}[lt]{48.86pt}\setlength\topsep{0pt}
	$\displaystyle Hom_{\mathcal{W}}( X,Y)$
	\end{minipage}};
\draw (505.86,164.39) node  [xscale=0.8,yscale=0.8] [align=left] {\begin{minipage}[lt]{48.86pt}\setlength\topsep{0pt}
	$\displaystyle F_{\theta }$
	\end{minipage}};
\draw (504.86,53.32) node  [xscale=0.8,yscale=0.8] [align=left] {\begin{minipage}[lt]{48.86pt}\setlength\topsep{0pt}
	$\displaystyle F_{\theta }$
	\end{minipage}};
\draw (551.86,41.32) node  [xscale=0.8,yscale=0.8] [align=left] {\begin{minipage}[lt]{48.86pt}\setlength\topsep{0pt}
	$\displaystyle F_{\theta }( X)$
	\end{minipage}};
\draw (550.86,171.32) node  [xscale=0.8,yscale=0.8] [align=left] {\begin{minipage}[lt]{48.86pt}\setlength\topsep{0pt}
	$\displaystyle F_{\theta }( Y)$
	\end{minipage}};
\draw (584.43,106.11) node  [xscale=0.8,yscale=0.8] [align=left] {\begin{minipage}[lt]{82.18pt}\setlength\topsep{0pt}
	$\displaystyle k( F_{\theta }( X) ,F_{\theta }( Y)) \ $
	\end{minipage}};
\draw (487.93,36.11) node  [xscale=0.8,yscale=0.8] [align=left] {\begin{minipage}[lt]{39.34pt}\setlength\topsep{0pt}
	\textcolor[rgb]{1,0,0}{(1) Lift}
	\end{minipage}};
\draw (581.43,89.11) node  [xscale=0.8,yscale=0.8] [align=left] {\begin{minipage}[lt]{61.78pt}\setlength\topsep{0pt}
	\textcolor[rgb]{1,0,0}{(2) Compute}
	\end{minipage}};
\draw (494.56,92.11) node  [xscale=0.8,yscale=0.8] [align=left] {\begin{minipage}[lt]{71.58pt}\setlength\topsep{0pt}
	\textcolor[rgb]{1,0,0}{(3) Determine}
	\end{minipage}};
\draw (99.43,126.77) node  [xscale=0.8,yscale=0.8] [align=left] {\begin{minipage}[lt]{11.46pt}\setlength\topsep{0pt}
	$\displaystyle X$
	\end{minipage}};
\draw (55.43,173.77) node  [xscale=0.8,yscale=0.8] [align=left] {\begin{minipage}[lt]{11.46pt}\setlength\topsep{0pt}
	$\displaystyle Y_{1}$
	\end{minipage}};
\draw (154.43,169.77) node  [xscale=0.8,yscale=0.8] [align=left] {\begin{minipage}[lt]{11.46pt}\setlength\topsep{0pt}
	$\displaystyle Y_{2}$
	\end{minipage}};
\draw (133.43,31.77) node  [xscale=0.8,yscale=0.8] [align=left] {\begin{minipage}[lt]{11.46pt}\setlength\topsep{0pt}
	$\displaystyle Y_{4}$
	\end{minipage}};
\draw (62.43,36.77) node  [xscale=0.8,yscale=0.8] [align=left] {\begin{minipage}[lt]{11.46pt}\setlength\topsep{0pt}
	$\displaystyle Y_{3}$
	\end{minipage}};
\draw (109.75,158.17) node  [xscale=0.8,yscale=0.8] [align=left] {\begin{minipage}[lt]{54.74pt}\setlength\topsep{0pt}
	\textcolor[rgb]{1,0,0}{(1) Record morphisms}
	\end{minipage}};
\draw (180.86,53.02) node  [xscale=0.8,yscale=0.8] [align=left] {\begin{minipage}[lt]{61.01pt}\setlength\topsep{0pt}
	\textcolor[rgb]{1,0,0}{(2) Collect}
	\end{minipage}};
\draw (280.29,64.77) node  [xscale=0.8,yscale=0.8] [align=left] {\begin{minipage}[lt]{69.75pt}\setlength\topsep{0pt}
	$\displaystyle Hom_{\mathcal{W}}( X,\cdot )$
	\end{minipage}};
\draw (210.32,116.74) node  [xscale=0.8,yscale=0.8] [align=left] {\begin{minipage}[lt]{53.96pt}\setlength\topsep{0pt}
	\textcolor[rgb]{1,0,0}{(3) Query}
	\end{minipage}};
\draw (440,203.11) node  [xscale=0.8,yscale=0.8] [align=left] {\begin{minipage}[lt]{14.96pt}\setlength\topsep{0pt}
	$\displaystyle \mathcal{W}$
	\end{minipage}};
\draw (535,203.11) node  [xscale=0.8,yscale=0.8] [align=left] {\begin{minipage}[lt]{14.96pt}\setlength\topsep{0pt}
	$\displaystyle \mathcal{W}^{\lor }$
	\end{minipage}};
\draw (105,203.67) node  [xscale=0.8,yscale=0.8] [align=left] {\begin{minipage}[lt]{14.96pt}\setlength\topsep{0pt}
	$\displaystyle \mathcal{W}$
	\end{minipage}};
\draw (271,97.67) node  [xscale=0.8,yscale=0.8] [align=left] {\begin{minipage}[lt]{14.96pt}\setlength\topsep{0pt}
	$\displaystyle \mathcal{W}^{\lor }$
	\end{minipage}};
\draw (169,71.77) node  [xscale=0.8,yscale=0.8] [align=left] {\begin{minipage}[lt]{14.96pt}\setlength\topsep{0pt}
	$\displaystyle k_{\mathcal{W}}$
	\end{minipage}};

\end{tikzpicture}

%% file: 4.tex
\tikzset{every picture/.style={line width=0.75pt}} 

\begin{tikzpicture}[x=0.75pt,y=0.75pt,yscale=-1,xscale=1]

\draw [color={rgb, 255:red, 255; green, 0; blue, 0 }  ,draw opacity=1 ][line width=2.25]    (138.67,165) -- (184.45,175.24) ;
\draw [shift={(189.33,176.33)}, rotate = 192.61] [fill={rgb, 255:red, 255; green, 0; blue, 0 }  ,fill opacity=1 ][line width=0.08]  [draw opacity=0] (10.36,-4.98) -- (0,0) -- (10.36,4.98) -- (6.88,0) -- cycle    ;
\draw  [color={rgb, 255:red, 74; green, 144; blue, 226 }  ,draw opacity=1 ] (96,161) .. controls (96,149.77) and (105.1,140.67) .. (116.33,140.67) .. controls (127.56,140.67) and (136.67,149.77) .. (136.67,161) .. controls (136.67,172.23) and (127.56,181.33) .. (116.33,181.33) .. controls (105.1,181.33) and (96,172.23) .. (96,161) -- cycle ;
\draw [color={rgb, 255:red, 255; green, 0; blue, 0 }  ,draw opacity=1 ][line width=2.25]    (129.33,144.33) -- (161.34,120.21) ;
\draw [shift={(165.33,117.2)}, rotate = 142.99] [fill={rgb, 255:red, 255; green, 0; blue, 0 }  ,fill opacity=1 ][line width=0.08]  [draw opacity=0] (10.36,-4.98) -- (0,0) -- (10.36,4.98) -- (6.88,0) -- cycle    ;
\draw [color={rgb, 255:red, 255; green, 0; blue, 0 }  ,draw opacity=1 ][line width=2.25]    (113.33,138.53) -- (109.85,104.84) ;
\draw [shift={(109.33,99.87)}, rotate = 84.09] [fill={rgb, 255:red, 255; green, 0; blue, 0 }  ,fill opacity=1 ][line width=0.08]  [draw opacity=0] (10.36,-4.98) -- (0,0) -- (10.36,4.98) -- (6.88,0) -- cycle    ;
\draw [color={rgb, 255:red, 255; green, 0; blue, 0 }  ,draw opacity=1 ][line width=2.25]    (96,157.2) -- (63.38,145.55) ;
\draw [shift={(58.67,143.87)}, rotate = 19.65] [fill={rgb, 255:red, 255; green, 0; blue, 0 }  ,fill opacity=1 ][line width=0.08]  [draw opacity=0] (10.36,-4.98) -- (0,0) -- (10.36,4.98) -- (6.88,0) -- cycle    ;
\draw [color={rgb, 255:red, 255; green, 0; blue, 0 }  ,draw opacity=1 ][line width=2.25]    (102.67,176.53) -- (77.57,211.79) ;
\draw [shift={(74.67,215.87)}, rotate = 305.45] [fill={rgb, 255:red, 255; green, 0; blue, 0 }  ,fill opacity=1 ][line width=0.08]  [draw opacity=0] (10.36,-4.98) -- (0,0) -- (10.36,4.98) -- (6.88,0) -- cycle    ;
\draw [color={rgb, 255:red, 255; green, 0; blue, 0 }  ,draw opacity=1 ][line width=2.25]    (122.67,181.87) -- (138.82,222.55) ;
\draw [shift={(140.67,227.2)}, rotate = 248.34] [fill={rgb, 255:red, 255; green, 0; blue, 0 }  ,fill opacity=1 ][line width=0.08]  [draw opacity=0] (10.36,-4.98) -- (0,0) -- (10.36,4.98) -- (6.88,0) -- cycle    ;
\draw  [color={rgb, 255:red, 74; green, 144; blue, 226 }  ,draw opacity=1 ] (310,171) .. controls (310,159.77) and (319.1,150.67) .. (330.33,150.67) .. controls (341.56,150.67) and (350.67,159.77) .. (350.67,171) .. controls (350.67,182.23) and (341.56,191.33) .. (330.33,191.33) .. controls (319.1,191.33) and (310,182.23) .. (310,171) -- cycle ;
\draw [color={rgb, 255:red, 255; green, 0; blue, 0 }  ,draw opacity=1 ][line width=2.25]    (548.33,172.33) -- (594.12,182.58) ;
\draw [shift={(599,183.67)}, rotate = 192.61] [fill={rgb, 255:red, 255; green, 0; blue, 0 }  ,fill opacity=1 ][line width=0.08]  [draw opacity=0] (10.36,-4.98) -- (0,0) -- (10.36,4.98) -- (6.88,0) -- cycle    ;
\draw [color={rgb, 255:red, 255; green, 0; blue, 0 }  ,draw opacity=1 ][line width=2.25]    (539,151.67) -- (571.01,127.54) ;
\draw [shift={(575,124.53)}, rotate = 142.99] [fill={rgb, 255:red, 255; green, 0; blue, 0 }  ,fill opacity=1 ][line width=0.08]  [draw opacity=0] (10.36,-4.98) -- (0,0) -- (10.36,4.98) -- (6.88,0) -- cycle    ;
\draw [color={rgb, 255:red, 255; green, 0; blue, 0 }  ,draw opacity=1 ][line width=2.25]    (523,145.87) -- (519.51,112.17) ;
\draw [shift={(519,107.2)}, rotate = 84.09] [fill={rgb, 255:red, 255; green, 0; blue, 0 }  ,fill opacity=1 ][line width=0.08]  [draw opacity=0] (10.36,-4.98) -- (0,0) -- (10.36,4.98) -- (6.88,0) -- cycle    ;
\draw [color={rgb, 255:red, 255; green, 0; blue, 0 }  ,draw opacity=1 ][line width=2.25]    (505.67,164.53) -- (473.04,152.88) ;
\draw [shift={(468.33,151.2)}, rotate = 19.65] [fill={rgb, 255:red, 255; green, 0; blue, 0 }  ,fill opacity=1 ][line width=0.08]  [draw opacity=0] (10.36,-4.98) -- (0,0) -- (10.36,4.98) -- (6.88,0) -- cycle    ;
\draw [color={rgb, 255:red, 255; green, 0; blue, 0 }  ,draw opacity=1 ][line width=2.25]    (512.33,183.87) -- (487.23,219.13) ;
\draw [shift={(484.33,223.2)}, rotate = 305.45] [fill={rgb, 255:red, 255; green, 0; blue, 0 }  ,fill opacity=1 ][line width=0.08]  [draw opacity=0] (10.36,-4.98) -- (0,0) -- (10.36,4.98) -- (6.88,0) -- cycle    ;
\draw [color={rgb, 255:red, 255; green, 0; blue, 0 }  ,draw opacity=1 ][line width=2.25]    (532.33,189.2) -- (548.49,229.89) ;
\draw [shift={(550.33,234.53)}, rotate = 248.34] [fill={rgb, 255:red, 255; green, 0; blue, 0 }  ,fill opacity=1 ][line width=0.08]  [draw opacity=0] (10.36,-4.98) -- (0,0) -- (10.36,4.98) -- (6.88,0) -- cycle    ;

\draw (117.33,160.77) node  [xscale=0.8,yscale=0.8] [align=left] {\begin{minipage}[lt]{23.57pt}\setlength\topsep{0pt}
	\textcolor[rgb]{0.29,0.56,0.89}{\textbf{Self}}
	\end{minipage}};
\draw (331.33,170.77) node  [xscale=0.8,yscale=0.8] [align=left] {\begin{minipage}[lt]{23.57pt}\setlength\topsep{0pt}
	\textcolor[rgb]{0.29,0.56,0.89}{\textbf{Self}}
	\end{minipage}};
\draw (129.67,265.33) node  [xscale=0.8,yscale=0.8] [align=left] {\begin{minipage}[lt]{135.55pt}\setlength\topsep{0pt}
	Morphisms between the model $\displaystyle I$ and other objects
	\end{minipage}};
\draw (338.67,265) node  [xscale=0.8,yscale=0.8] [align=left] {\begin{minipage}[lt]{79.79pt}\setlength\topsep{0pt}
	Self-state as an object $\displaystyle I$ in W
	\end{minipage}};
\draw (540,265) node  [xscale=0.8,yscale=0.8] [align=left] {\begin{minipage}[lt]{89.76pt}\setlength\topsep{0pt}
	Self-state as a presheaf in W
	\end{minipage}};

\end{tikzpicture}

%% file: 8.tex
\tikzset{every picture/.style={line width=0.75pt}} 

\begin{tikzpicture}[x=0.75pt,y=0.75pt,yscale=-1,xscale=1]

\draw [color={rgb, 255:red, 255; green, 0; blue, 0 }  ,draw opacity=1 ][line width=2.25]    (138.67,165) -- (184.45,175.24) ;
\draw [shift={(189.33,176.33)}, rotate = 192.61] [fill={rgb, 255:red, 255; green, 0; blue, 0 }  ,fill opacity=1 ][line width=0.08]  [draw opacity=0] (10.36,-4.98) -- (0,0) -- (10.36,4.98) -- (6.88,0) -- cycle    ;
\draw  [color={rgb, 255:red, 74; green, 144; blue, 226 }  ,draw opacity=1 ] (96,161) .. controls (96,149.77) and (105.1,140.67) .. (116.33,140.67) .. controls (127.56,140.67) and (136.67,149.77) .. (136.67,161) .. controls (136.67,172.23) and (127.56,181.33) .. (116.33,181.33) .. controls (105.1,181.33) and (96,172.23) .. (96,161) -- cycle ;
\draw [color={rgb, 255:red, 255; green, 0; blue, 0 }  ,draw opacity=1 ][line width=2.25]    (129.33,144.33) -- (161.34,120.21) ;
\draw [shift={(165.33,117.2)}, rotate = 142.99] [fill={rgb, 255:red, 255; green, 0; blue, 0 }  ,fill opacity=1 ][line width=0.08]  [draw opacity=0] (10.36,-4.98) -- (0,0) -- (10.36,4.98) -- (6.88,0) -- cycle    ;
\draw [color={rgb, 255:red, 255; green, 0; blue, 0 }  ,draw opacity=1 ][line width=2.25]    (113.33,138.53) -- (109.85,104.84) ;
\draw [shift={(109.33,99.87)}, rotate = 84.09] [fill={rgb, 255:red, 255; green, 0; blue, 0 }  ,fill opacity=1 ][line width=0.08]  [draw opacity=0] (10.36,-4.98) -- (0,0) -- (10.36,4.98) -- (6.88,0) -- cycle    ;
\draw [color={rgb, 255:red, 255; green, 0; blue, 0 }  ,draw opacity=1 ][line width=2.25]    (96,157.2) -- (63.38,145.55) ;
\draw [shift={(58.67,143.87)}, rotate = 19.65] [fill={rgb, 255:red, 255; green, 0; blue, 0 }  ,fill opacity=1 ][line width=0.08]  [draw opacity=0] (10.36,-4.98) -- (0,0) -- (10.36,4.98) -- (6.88,0) -- cycle    ;
\draw [color={rgb, 255:red, 255; green, 0; blue, 0 }  ,draw opacity=1 ][line width=2.25]    (102.67,176.53) -- (77.57,211.79) ;
\draw [shift={(74.67,215.87)}, rotate = 305.45] [fill={rgb, 255:red, 255; green, 0; blue, 0 }  ,fill opacity=1 ][line width=0.08]  [draw opacity=0] (10.36,-4.98) -- (0,0) -- (10.36,4.98) -- (6.88,0) -- cycle    ;
\draw [color={rgb, 255:red, 255; green, 0; blue, 0 }  ,draw opacity=1 ][line width=2.25]    (122.67,181.87) -- (138.82,222.55) ;
\draw [shift={(140.67,227.2)}, rotate = 248.34] [fill={rgb, 255:red, 255; green, 0; blue, 0 }  ,fill opacity=1 ][line width=0.08]  [draw opacity=0] (10.36,-4.98) -- (0,0) -- (10.36,4.98) -- (6.88,0) -- cycle    ;
\draw [color={rgb, 255:red, 255; green, 0; blue, 0 }  ,draw opacity=1 ][line width=2.25]    (288.67,164.67) -- (334.45,174.91) ;
\draw [shift={(339.33,176)}, rotate = 192.61] [fill={rgb, 255:red, 255; green, 0; blue, 0 }  ,fill opacity=1 ][line width=0.08]  [draw opacity=0] (10.36,-4.98) -- (0,0) -- (10.36,4.98) -- (6.88,0) -- cycle    ;
\draw  [color={rgb, 255:red, 74; green, 144; blue, 226 }  ,draw opacity=1 ] (246,160.67) .. controls (246,149.44) and (255.1,140.33) .. (266.33,140.33) .. controls (277.56,140.33) and (286.67,149.44) .. (286.67,160.67) .. controls (286.67,171.9) and (277.56,181) .. (266.33,181) .. controls (255.1,181) and (246,171.9) .. (246,160.67) -- cycle ;
\draw [color={rgb, 255:red, 255; green, 0; blue, 0 }  ,draw opacity=1 ][line width=2.25]    (279.33,144) -- (311.34,119.88) ;
\draw [shift={(315.33,116.87)}, rotate = 142.99] [fill={rgb, 255:red, 255; green, 0; blue, 0 }  ,fill opacity=1 ][line width=0.08]  [draw opacity=0] (10.36,-4.98) -- (0,0) -- (10.36,4.98) -- (6.88,0) -- cycle    ;
\draw [color={rgb, 255:red, 255; green, 0; blue, 0 }  ,draw opacity=1 ][line width=2.25]    (263.33,138.2) -- (259.85,104.51) ;
\draw [shift={(259.33,99.53)}, rotate = 84.09] [fill={rgb, 255:red, 255; green, 0; blue, 0 }  ,fill opacity=1 ][line width=0.08]  [draw opacity=0] (10.36,-4.98) -- (0,0) -- (10.36,4.98) -- (6.88,0) -- cycle    ;
\draw [color={rgb, 255:red, 255; green, 0; blue, 0 }  ,draw opacity=1 ][line width=2.25]    (246,156.87) -- (213.38,145.22) ;
\draw [shift={(208.67,143.53)}, rotate = 19.65] [fill={rgb, 255:red, 255; green, 0; blue, 0 }  ,fill opacity=1 ][line width=0.08]  [draw opacity=0] (10.36,-4.98) -- (0,0) -- (10.36,4.98) -- (6.88,0) -- cycle    ;
\draw [color={rgb, 255:red, 255; green, 0; blue, 0 }  ,draw opacity=0.09 ][line width=2.25]    (252.67,176.2) -- (227.57,211.46) ;
\draw [shift={(224.67,215.53)}, rotate = 305.45] [fill={rgb, 255:red, 255; green, 0; blue, 0 }  ,fill opacity=0.09 ][line width=0.08]  [draw opacity=0] (10.36,-4.98) -- (0,0) -- (10.36,4.98) -- (6.88,0) -- cycle    ;
\draw [color={rgb, 255:red, 255; green, 0; blue, 0 }  ,draw opacity=0.09 ][line width=2.25]    (272.67,181.53) -- (288.82,222.22) ;
\draw [shift={(290.67,226.87)}, rotate = 248.34] [fill={rgb, 255:red, 255; green, 0; blue, 0 }  ,fill opacity=0.09 ][line width=0.08]  [draw opacity=0] (10.36,-4.98) -- (0,0) -- (10.36,4.98) -- (6.88,0) -- cycle    ;
\draw  [dash pattern={on 0.84pt off 2.51pt}] (68,187.57) .. controls (68,181.11) and (73.24,175.87) .. (79.71,175.87) -- (135.23,175.87) .. controls (141.69,175.87) and (146.93,181.11) .. (146.93,187.57) -- (146.93,222.69) .. controls (146.93,229.16) and (141.69,234.4) .. (135.23,234.4) -- (79.71,234.4) .. controls (73.24,234.4) and (68,229.16) .. (68,222.69) -- cycle ;
\draw  [dash pattern={on 0.84pt off 2.51pt}] (218,187.57) .. controls (218,181.11) and (223.24,175.87) .. (229.71,175.87) -- (285.23,175.87) .. controls (291.69,175.87) and (296.93,181.11) .. (296.93,187.57) -- (296.93,222.69) .. controls (296.93,229.16) and (291.69,234.4) .. (285.23,234.4) -- (229.71,234.4) .. controls (223.24,234.4) and (218,229.16) .. (218,222.69) -- cycle ;
\draw [color={rgb, 255:red, 255; green, 0; blue, 0 }  ,draw opacity=1 ][line width=2.25]    (467.67,164.67) -- (513.45,174.91) ;
\draw [shift={(518.33,176)}, rotate = 192.61] [fill={rgb, 255:red, 255; green, 0; blue, 0 }  ,fill opacity=1 ][line width=0.08]  [draw opacity=0] (10.36,-4.98) -- (0,0) -- (10.36,4.98) -- (6.88,0) -- cycle    ;
\draw  [color={rgb, 255:red, 74; green, 144; blue, 226 }  ,draw opacity=1 ] (425,160.67) .. controls (425,149.44) and (434.1,140.33) .. (445.33,140.33) .. controls (456.56,140.33) and (465.67,149.44) .. (465.67,160.67) .. controls (465.67,171.9) and (456.56,181) .. (445.33,181) .. controls (434.1,181) and (425,171.9) .. (425,160.67) -- cycle ;
\draw [color={rgb, 255:red, 255; green, 0; blue, 0 }  ,draw opacity=1 ][line width=2.25]    (458.33,144) -- (490.34,119.88) ;
\draw [shift={(494.33,116.87)}, rotate = 142.99] [fill={rgb, 255:red, 255; green, 0; blue, 0 }  ,fill opacity=1 ][line width=0.08]  [draw opacity=0] (10.36,-4.98) -- (0,0) -- (10.36,4.98) -- (6.88,0) -- cycle    ;
\draw [color={rgb, 255:red, 255; green, 0; blue, 0 }  ,draw opacity=1 ][line width=2.25]    (442.33,138.2) -- (438.85,104.51) ;
\draw [shift={(438.33,99.53)}, rotate = 84.09] [fill={rgb, 255:red, 255; green, 0; blue, 0 }  ,fill opacity=1 ][line width=0.08]  [draw opacity=0] (10.36,-4.98) -- (0,0) -- (10.36,4.98) -- (6.88,0) -- cycle    ;
\draw [color={rgb, 255:red, 255; green, 0; blue, 0 }  ,draw opacity=1 ][line width=2.25]    (425,156.87) -- (392.38,145.22) ;
\draw [shift={(387.67,143.53)}, rotate = 19.65] [fill={rgb, 255:red, 255; green, 0; blue, 0 }  ,fill opacity=1 ][line width=0.08]  [draw opacity=0] (10.36,-4.98) -- (0,0) -- (10.36,4.98) -- (6.88,0) -- cycle    ;
\draw [color={rgb, 255:red, 255; green, 0; blue, 0 }  ,draw opacity=0.09 ][line width=2.25]    (431.67,176.2) -- (406.57,211.46) ;
\draw [shift={(403.67,215.53)}, rotate = 305.45] [fill={rgb, 255:red, 255; green, 0; blue, 0 }  ,fill opacity=0.09 ][line width=0.08]  [draw opacity=0] (10.36,-4.98) -- (0,0) -- (10.36,4.98) -- (6.88,0) -- cycle    ;
\draw [color={rgb, 255:red, 255; green, 0; blue, 0 }  ,draw opacity=0.09 ][line width=2.25]    (451.67,181.53) -- (467.82,222.22) ;
\draw [shift={(469.67,226.87)}, rotate = 248.34] [fill={rgb, 255:red, 255; green, 0; blue, 0 }  ,fill opacity=0.09 ][line width=0.08]  [draw opacity=0] (10.36,-4.98) -- (0,0) -- (10.36,4.98) -- (6.88,0) -- cycle    ;
\draw [color={rgb, 255:red, 255; green, 0; blue, 0 }  ,draw opacity=1 ][line width=2.25]    (617.67,164.33) -- (663.45,174.58) ;
\draw [shift={(668.33,175.67)}, rotate = 192.61] [fill={rgb, 255:red, 255; green, 0; blue, 0 }  ,fill opacity=1 ][line width=0.08]  [draw opacity=0] (10.36,-4.98) -- (0,0) -- (10.36,4.98) -- (6.88,0) -- cycle    ;
\draw  [color={rgb, 255:red, 74; green, 144; blue, 226 }  ,draw opacity=1 ] (575,160.33) .. controls (575,149.1) and (584.1,140) .. (595.33,140) .. controls (606.56,140) and (615.67,149.1) .. (615.67,160.33) .. controls (615.67,171.56) and (606.56,180.67) .. (595.33,180.67) .. controls (584.1,180.67) and (575,171.56) .. (575,160.33) -- cycle ;
\draw [color={rgb, 255:red, 255; green, 0; blue, 0 }  ,draw opacity=1 ][line width=2.25]    (608.33,143.67) -- (640.34,119.54) ;
\draw [shift={(644.33,116.53)}, rotate = 142.99] [fill={rgb, 255:red, 255; green, 0; blue, 0 }  ,fill opacity=1 ][line width=0.08]  [draw opacity=0] (10.36,-4.98) -- (0,0) -- (10.36,4.98) -- (6.88,0) -- cycle    ;
\draw [color={rgb, 255:red, 255; green, 0; blue, 0 }  ,draw opacity=1 ][line width=2.25]    (592.33,137.87) -- (588.85,104.17) ;
\draw [shift={(588.33,99.2)}, rotate = 84.09] [fill={rgb, 255:red, 255; green, 0; blue, 0 }  ,fill opacity=1 ][line width=0.08]  [draw opacity=0] (10.36,-4.98) -- (0,0) -- (10.36,4.98) -- (6.88,0) -- cycle    ;
\draw [color={rgb, 255:red, 255; green, 0; blue, 0 }  ,draw opacity=1 ][line width=2.25]    (575,156.53) -- (542.38,144.88) ;
\draw [shift={(537.67,143.2)}, rotate = 19.65] [fill={rgb, 255:red, 255; green, 0; blue, 0 }  ,fill opacity=1 ][line width=0.08]  [draw opacity=0] (10.36,-4.98) -- (0,0) -- (10.36,4.98) -- (6.88,0) -- cycle    ;
\draw [color={rgb, 255:red, 255; green, 0; blue, 0 }  ,draw opacity=1 ][line width=2.25]    (581.67,175.87) -- (556.57,211.13) ;
\draw [shift={(553.67,215.2)}, rotate = 305.45] [fill={rgb, 255:red, 255; green, 0; blue, 0 }  ,fill opacity=1 ][line width=0.08]  [draw opacity=0] (10.36,-4.98) -- (0,0) -- (10.36,4.98) -- (6.88,0) -- cycle    ;
\draw [color={rgb, 255:red, 255; green, 0; blue, 0 }  ,draw opacity=1 ][line width=2.25]    (601.67,181.2) -- (617.82,221.89) ;
\draw [shift={(619.67,226.53)}, rotate = 248.34] [fill={rgb, 255:red, 255; green, 0; blue, 0 }  ,fill opacity=1 ][line width=0.08]  [draw opacity=0] (10.36,-4.98) -- (0,0) -- (10.36,4.98) -- (6.88,0) -- cycle    ;
\draw  [dash pattern={on 0.84pt off 2.51pt}] (397,187.24) .. controls (397,180.77) and (402.24,175.53) .. (408.71,175.53) -- (464.23,175.53) .. controls (470.69,175.53) and (475.93,180.77) .. (475.93,187.24) -- (475.93,222.36) .. controls (475.93,228.83) and (470.69,234.07) .. (464.23,234.07) -- (408.71,234.07) .. controls (402.24,234.07) and (397,228.83) .. (397,222.36) -- cycle ;
\draw  [dash pattern={on 0.84pt off 2.51pt}] (547,187.24) .. controls (547,180.77) and (552.24,175.53) .. (558.71,175.53) -- (614.23,175.53) .. controls (620.69,175.53) and (625.93,180.77) .. (625.93,187.24) -- (625.93,222.36) .. controls (625.93,228.83) and (620.69,234.07) .. (614.23,234.07) -- (558.71,234.07) .. controls (552.24,234.07) and (547,228.83) .. (547,222.36) -- cycle ;
\draw    (360.27,53.73) -- (360.27,275.73) ;

\draw (117.33,160.77) node  [xscale=0.8,yscale=0.8] [align=left] {\begin{minipage}[lt]{23.57pt}\setlength\topsep{0pt}
	\textcolor[rgb]{0.29,0.56,0.89}{\textbf{Self}}
	\end{minipage}};
\draw (260.33,159.67) node  [xscale=0.8,yscale=0.8] [align=left] {\begin{minipage}[lt]{23.57pt}\setlength\topsep{0pt}
	\textcolor[rgb]{0.29,0.56,0.89}{\textbf{Other}}
	\end{minipage}};
\draw (109.93,257.03) node  [xscale=0.8,yscale=0.8] [align=left] {\begin{minipage}[lt]{75.71pt}\setlength\topsep{0pt}
	Perceived by\\Private sensors
	\end{minipage}};
\draw (260.43,257.33) node  [xscale=0.8,yscale=0.8] [align=left] {\begin{minipage}[lt]{44.2pt}\setlength\topsep{0pt}
	Cannot perceive
	\end{minipage}};
\draw (446.33,160.43) node  [xscale=0.8,yscale=0.8] [align=left] {\begin{minipage}[lt]{23.57pt}\setlength\topsep{0pt}
	\textcolor[rgb]{0.29,0.56,0.89}{\textbf{Self}}
	\end{minipage}};
\draw (589.33,159.33) node  [xscale=0.8,yscale=0.8] [align=left] {\begin{minipage}[lt]{23.57pt}\setlength\topsep{0pt}
	\textcolor[rgb]{0.29,0.56,0.89}{\textbf{Other}}
	\end{minipage}};
\draw (594.6,256.03) node  [xscale=0.8,yscale=0.8] [align=left] {\begin{minipage}[lt]{75.71pt}\setlength\topsep{0pt}
	Perceived by\\Private sensors
	\end{minipage}};
\draw (439.43,257) node  [xscale=0.8,yscale=0.8] [align=left] {\begin{minipage}[lt]{44.2pt}\setlength\topsep{0pt}
	Cannot perceive
	\end{minipage}};
\draw (198.63,67.53) node  [xscale=0.8,yscale=0.8] [align=left] {\begin{minipage}[lt]{134.14pt}\setlength\topsep{0pt}
	{\Large Real world scenario}
	\end{minipage}};
\draw (527.47,68.2) node  [xscale=0.8,yscale=0.8] [align=left] {\begin{minipage}[lt]{150.23pt}\setlength\topsep{0pt}
	{\Large Experimental scenario}
	\end{minipage}};

\end{tikzpicture}

%% file: 13.tex
\tikzset{every picture/.style={line width=0.75pt}} 

\begin{tikzpicture}[x=0.75pt,y=0.75pt,yscale=-1,xscale=1]

\draw  [draw opacity=0][fill={rgb, 255:red, 0; green, 0; blue, 0 }  ,fill opacity=1 ] (44,160.5) .. controls (44,157.46) and (46.46,155) .. (49.5,155) .. controls (52.54,155) and (55,157.46) .. (55,160.5) .. controls (55,163.54) and (52.54,166) .. (49.5,166) .. controls (46.46,166) and (44,163.54) .. (44,160.5) -- cycle ;
\draw  [draw opacity=0][fill={rgb, 255:red, 0; green, 0; blue, 0 }  ,fill opacity=1 ] (102,150.5) .. controls (102,147.46) and (104.46,145) .. (107.5,145) .. controls (110.54,145) and (113,147.46) .. (113,150.5) .. controls (113,153.54) and (110.54,156) .. (107.5,156) .. controls (104.46,156) and (102,153.54) .. (102,150.5) -- cycle ;
\draw  [draw opacity=0][fill={rgb, 255:red, 0; green, 0; blue, 0 }  ,fill opacity=1 ] (51,246.5) .. controls (51,243.46) and (53.46,241) .. (56.5,241) .. controls (59.54,241) and (62,243.46) .. (62,246.5) .. controls (62,249.54) and (59.54,252) .. (56.5,252) .. controls (53.46,252) and (51,249.54) .. (51,246.5) -- cycle ;
\draw [color={rgb, 255:red, 100; green, 100; blue, 100 }  ,draw opacity=1 ]   (56.5,241) -- (105.96,158.57) ;
\draw [shift={(107.5,156)}, rotate = 120.96] [fill={rgb, 255:red, 100; green, 100; blue, 100 }  ,fill opacity=1 ][line width=0.08]  [draw opacity=0] (10.72,-5.15) -- (0,0) -- (10.72,5.15) -- (7.12,0) -- cycle    ;
\draw [color={rgb, 255:red, 100; green, 100; blue, 100 }  ,draw opacity=1 ]   (56.5,241) -- (49.78,168.99) ;
\draw [shift={(49.5,166)}, rotate = 84.67] [fill={rgb, 255:red, 100; green, 100; blue, 100 }  ,fill opacity=1 ][line width=0.08]  [draw opacity=0] (10.72,-5.15) -- (0,0) -- (10.72,5.15) -- (7.12,0) -- cycle    ;
\draw [color={rgb, 255:red, 100; green, 100; blue, 100 }  ,draw opacity=1 ]   (102,150.5) -- (57.93,159.88) ;
\draw [shift={(55,160.5)}, rotate = 347.99] [fill={rgb, 255:red, 100; green, 100; blue, 100 }  ,fill opacity=1 ][line width=0.08]  [draw opacity=0] (10.72,-5.15) -- (0,0) -- (10.72,5.15) -- (7.12,0) -- cycle    ;
\draw  [draw opacity=0][fill={rgb, 255:red, 0; green, 0; blue, 0 }  ,fill opacity=1 ] (71,46.5) .. controls (71,43.46) and (73.46,41) .. (76.5,41) .. controls (79.54,41) and (82,43.46) .. (82,46.5) .. controls (82,49.54) and (79.54,52) .. (76.5,52) .. controls (73.46,52) and (71,49.54) .. (71,46.5) -- cycle ;
\draw [color={rgb, 255:red, 100; green, 100; blue, 100 }  ,draw opacity=1 ] [dash pattern={on 0.84pt off 2.51pt}]  (76.5,52) -- (50.26,152.1) ;
\draw [shift={(49.5,155)}, rotate = 284.69] [fill={rgb, 255:red, 100; green, 100; blue, 100 }  ,fill opacity=1 ][line width=0.08]  [draw opacity=0] (10.72,-5.15) -- (0,0) -- (10.72,5.15) -- (7.12,0) -- cycle    ;
\draw [color={rgb, 255:red, 100; green, 100; blue, 100 }  ,draw opacity=1 ] [dash pattern={on 0.84pt off 2.51pt}]  (76.5,52) -- (100.22,139.6) ;
\draw [shift={(101,142.5)}, rotate = 254.85] [fill={rgb, 255:red, 100; green, 100; blue, 100 }  ,fill opacity=1 ][line width=0.08]  [draw opacity=0] (10.72,-5.15) -- (0,0) -- (10.72,5.15) -- (7.12,0) -- cycle    ;
\draw [color={rgb, 255:red, 100; green, 100; blue, 100 }  ,draw opacity=1 ] [dash pattern={on 0.84pt off 2.51pt}]  (77,62.5) -- (56.84,238.02) ;
\draw [shift={(56.5,241)}, rotate = 276.55] [fill={rgb, 255:red, 100; green, 100; blue, 100 }  ,fill opacity=1 ][line width=0.08]  [draw opacity=0] (10.72,-5.15) -- (0,0) -- (10.72,5.15) -- (7.12,0) -- cycle    ;
\draw [color={rgb, 255:red, 100; green, 100; blue, 100 }  ,draw opacity=1 ]   (82,46.5) -- (185,44.31) ;
\draw [shift={(188,44.25)}, rotate = 178.78] [fill={rgb, 255:red, 100; green, 100; blue, 100 }  ,fill opacity=1 ][line width=0.08]  [draw opacity=0] (10.72,-5.15) -- (0,0) -- (10.72,5.15) -- (7.12,0) -- cycle    ;
\draw  [draw opacity=0][fill={rgb, 255:red, 126; green, 211; blue, 33 }  ,fill opacity=0.14 ] (146.33,10.5) -- (256,10.5) -- (256,273.33) -- (146.33,273.33) -- cycle ;
\draw  [draw opacity=0][fill={rgb, 255:red, 0; green, 0; blue, 0 }  ,fill opacity=1 ] (161,158.25) .. controls (161,155.21) and (163.46,152.75) .. (166.5,152.75) .. controls (169.54,152.75) and (172,155.21) .. (172,158.25) .. controls (172,161.29) and (169.54,163.75) .. (166.5,163.75) .. controls (163.46,163.75) and (161,161.29) .. (161,158.25) -- cycle ;
\draw  [draw opacity=0][fill={rgb, 255:red, 0; green, 0; blue, 0 }  ,fill opacity=1 ] (219,148.25) .. controls (219,145.21) and (221.46,142.75) .. (224.5,142.75) .. controls (227.54,142.75) and (230,145.21) .. (230,148.25) .. controls (230,151.29) and (227.54,153.75) .. (224.5,153.75) .. controls (221.46,153.75) and (219,151.29) .. (219,148.25) -- cycle ;
\draw  [draw opacity=0][fill={rgb, 255:red, 0; green, 0; blue, 0 }  ,fill opacity=1 ] (168,244.25) .. controls (168,241.21) and (170.46,238.75) .. (173.5,238.75) .. controls (176.54,238.75) and (179,241.21) .. (179,244.25) .. controls (179,247.29) and (176.54,249.75) .. (173.5,249.75) .. controls (170.46,249.75) and (168,247.29) .. (168,244.25) -- cycle ;
\draw [color={rgb, 255:red, 100; green, 100; blue, 100 }  ,draw opacity=1 ]   (173.5,238.75) -- (222.96,156.32) ;
\draw [shift={(224.5,153.75)}, rotate = 120.96] [fill={rgb, 255:red, 100; green, 100; blue, 100 }  ,fill opacity=1 ][line width=0.08]  [draw opacity=0] (10.72,-5.15) -- (0,0) -- (10.72,5.15) -- (7.12,0) -- cycle    ;
\draw [color={rgb, 255:red, 100; green, 100; blue, 100 }  ,draw opacity=1 ]   (173.5,238.75) -- (166.78,166.74) ;
\draw [shift={(166.5,163.75)}, rotate = 84.67] [fill={rgb, 255:red, 100; green, 100; blue, 100 }  ,fill opacity=1 ][line width=0.08]  [draw opacity=0] (10.72,-5.15) -- (0,0) -- (10.72,5.15) -- (7.12,0) -- cycle    ;
\draw [color={rgb, 255:red, 100; green, 100; blue, 100 }  ,draw opacity=1 ]   (219,148.25) -- (174.93,157.63) ;
\draw [shift={(172,158.25)}, rotate = 347.99] [fill={rgb, 255:red, 100; green, 100; blue, 100 }  ,fill opacity=1 ][line width=0.08]  [draw opacity=0] (10.72,-5.15) -- (0,0) -- (10.72,5.15) -- (7.12,0) -- cycle    ;
\draw  [draw opacity=0][fill={rgb, 255:red, 0; green, 0; blue, 0 }  ,fill opacity=1 ] (188,44.25) .. controls (188,41.21) and (190.46,38.75) .. (193.5,38.75) .. controls (196.54,38.75) and (199,41.21) .. (199,44.25) .. controls (199,47.29) and (196.54,49.75) .. (193.5,49.75) .. controls (190.46,49.75) and (188,47.29) .. (188,44.25) -- cycle ;
\draw [color={rgb, 255:red, 100; green, 100; blue, 100 }  ,draw opacity=1 ]   (113,150.5) -- (216,148.31) ;
\draw [shift={(219,148.25)}, rotate = 178.78] [fill={rgb, 255:red, 100; green, 100; blue, 100 }  ,fill opacity=1 ][line width=0.08]  [draw opacity=0] (10.72,-5.15) -- (0,0) -- (10.72,5.15) -- (7.12,0) -- cycle    ;
\draw  [dash pattern={on 0.84pt off 2.51pt}]  (291,11.83) -- (291,277.33) ;
\draw [color={rgb, 255:red, 100; green, 100; blue, 100 }  ,draw opacity=1 ]   (55,160.5) -- (158,158.31) ;
\draw [shift={(161,158.25)}, rotate = 178.78] [fill={rgb, 255:red, 100; green, 100; blue, 100 }  ,fill opacity=1 ][line width=0.08]  [draw opacity=0] (10.72,-5.15) -- (0,0) -- (10.72,5.15) -- (7.12,0) -- cycle    ;
\draw [color={rgb, 255:red, 100; green, 100; blue, 100 }  ,draw opacity=1 ]   (62,246.5) -- (165,244.31) ;
\draw [shift={(168,244.25)}, rotate = 178.78] [fill={rgb, 255:red, 100; green, 100; blue, 100 }  ,fill opacity=1 ][line width=0.08]  [draw opacity=0] (10.72,-5.15) -- (0,0) -- (10.72,5.15) -- (7.12,0) -- cycle    ;
\draw  [draw opacity=0][fill={rgb, 255:red, 245; green, 166; blue, 35 }  ,fill opacity=0.13 ] (18,8.85) -- (126,8.85) -- (126,273.33) -- (18,273.33) -- cycle ;
\draw  [draw opacity=0][fill={rgb, 255:red, 0; green, 0; blue, 0 }  ,fill opacity=1 ] (335,161.25) .. controls (335,158.21) and (337.46,155.75) .. (340.5,155.75) .. controls (343.54,155.75) and (346,158.21) .. (346,161.25) .. controls (346,164.29) and (343.54,166.75) .. (340.5,166.75) .. controls (337.46,166.75) and (335,164.29) .. (335,161.25) -- cycle ;
\draw  [draw opacity=0][fill={rgb, 255:red, 0; green, 0; blue, 0 }  ,fill opacity=1 ] (393,151.25) .. controls (393,148.21) and (395.46,145.75) .. (398.5,145.75) .. controls (401.54,145.75) and (404,148.21) .. (404,151.25) .. controls (404,154.29) and (401.54,156.75) .. (398.5,156.75) .. controls (395.46,156.75) and (393,154.29) .. (393,151.25) -- cycle ;
\draw  [draw opacity=0][fill={rgb, 255:red, 0; green, 0; blue, 0 }  ,fill opacity=1 ] (342,247.25) .. controls (342,244.21) and (344.46,241.75) .. (347.5,241.75) .. controls (350.54,241.75) and (353,244.21) .. (353,247.25) .. controls (353,250.29) and (350.54,252.75) .. (347.5,252.75) .. controls (344.46,252.75) and (342,250.29) .. (342,247.25) -- cycle ;
\draw [color={rgb, 255:red, 100; green, 100; blue, 100 }  ,draw opacity=1 ]   (347.5,241.75) -- (396.96,159.32) ;
\draw [shift={(398.5,156.75)}, rotate = 120.96] [fill={rgb, 255:red, 100; green, 100; blue, 100 }  ,fill opacity=1 ][line width=0.08]  [draw opacity=0] (10.72,-5.15) -- (0,0) -- (10.72,5.15) -- (7.12,0) -- cycle    ;
\draw [color={rgb, 255:red, 100; green, 100; blue, 100 }  ,draw opacity=1 ]   (347.5,241.75) -- (340.78,169.74) ;
\draw [shift={(340.5,166.75)}, rotate = 84.67] [fill={rgb, 255:red, 100; green, 100; blue, 100 }  ,fill opacity=1 ][line width=0.08]  [draw opacity=0] (10.72,-5.15) -- (0,0) -- (10.72,5.15) -- (7.12,0) -- cycle    ;
\draw [color={rgb, 255:red, 100; green, 100; blue, 100 }  ,draw opacity=1 ]   (393,151.25) -- (348.93,160.63) ;
\draw [shift={(346,161.25)}, rotate = 347.99] [fill={rgb, 255:red, 100; green, 100; blue, 100 }  ,fill opacity=1 ][line width=0.08]  [draw opacity=0] (10.72,-5.15) -- (0,0) -- (10.72,5.15) -- (7.12,0) -- cycle    ;
\draw  [draw opacity=0][fill={rgb, 255:red, 0; green, 0; blue, 0 }  ,fill opacity=1 ] (362,47.25) .. controls (362,44.21) and (364.46,41.75) .. (367.5,41.75) .. controls (370.54,41.75) and (373,44.21) .. (373,47.25) .. controls (373,50.29) and (370.54,52.75) .. (367.5,52.75) .. controls (364.46,52.75) and (362,50.29) .. (362,47.25) -- cycle ;
\draw [color={rgb, 255:red, 100; green, 100; blue, 100 }  ,draw opacity=1 ]   (373,47.25) -- (476,45.06) ;
\draw [shift={(479,45)}, rotate = 178.78] [fill={rgb, 255:red, 100; green, 100; blue, 100 }  ,fill opacity=1 ][line width=0.08]  [draw opacity=0] (10.72,-5.15) -- (0,0) -- (10.72,5.15) -- (7.12,0) -- cycle    ;
\draw  [draw opacity=0][fill={rgb, 255:red, 126; green, 211; blue, 33 }  ,fill opacity=0.14 ] (437.33,11.25) -- (562,11.25) -- (562,273.33) -- (437.33,273.33) -- cycle ;
\draw  [draw opacity=0][fill={rgb, 255:red, 0; green, 0; blue, 0 }  ,fill opacity=1 ] (452,159) .. controls (452,155.96) and (454.46,153.5) .. (457.5,153.5) .. controls (460.54,153.5) and (463,155.96) .. (463,159) .. controls (463,162.04) and (460.54,164.5) .. (457.5,164.5) .. controls (454.46,164.5) and (452,162.04) .. (452,159) -- cycle ;
\draw  [draw opacity=0][fill={rgb, 255:red, 0; green, 0; blue, 0 }  ,fill opacity=1 ] (510,149) .. controls (510,145.96) and (512.46,143.5) .. (515.5,143.5) .. controls (518.54,143.5) and (521,145.96) .. (521,149) .. controls (521,152.04) and (518.54,154.5) .. (515.5,154.5) .. controls (512.46,154.5) and (510,152.04) .. (510,149) -- cycle ;
\draw  [draw opacity=0][fill={rgb, 255:red, 0; green, 0; blue, 0 }  ,fill opacity=1 ] (459,245) .. controls (459,241.96) and (461.46,239.5) .. (464.5,239.5) .. controls (467.54,239.5) and (470,241.96) .. (470,245) .. controls (470,248.04) and (467.54,250.5) .. (464.5,250.5) .. controls (461.46,250.5) and (459,248.04) .. (459,245) -- cycle ;
\draw [color={rgb, 255:red, 100; green, 100; blue, 100 }  ,draw opacity=1 ]   (464.5,239.5) -- (513.96,157.07) ;
\draw [shift={(515.5,154.5)}, rotate = 120.96] [fill={rgb, 255:red, 100; green, 100; blue, 100 }  ,fill opacity=1 ][line width=0.08]  [draw opacity=0] (10.72,-5.15) -- (0,0) -- (10.72,5.15) -- (7.12,0) -- cycle    ;
\draw [color={rgb, 255:red, 100; green, 100; blue, 100 }  ,draw opacity=1 ]   (464.5,239.5) -- (457.78,167.49) ;
\draw [shift={(457.5,164.5)}, rotate = 84.67] [fill={rgb, 255:red, 100; green, 100; blue, 100 }  ,fill opacity=1 ][line width=0.08]  [draw opacity=0] (10.72,-5.15) -- (0,0) -- (10.72,5.15) -- (7.12,0) -- cycle    ;
\draw [color={rgb, 255:red, 100; green, 100; blue, 100 }  ,draw opacity=1 ]   (510,149) -- (465.93,158.38) ;
\draw [shift={(463,159)}, rotate = 347.99] [fill={rgb, 255:red, 100; green, 100; blue, 100 }  ,fill opacity=1 ][line width=0.08]  [draw opacity=0] (10.72,-5.15) -- (0,0) -- (10.72,5.15) -- (7.12,0) -- cycle    ;
\draw  [draw opacity=0][fill={rgb, 255:red, 0; green, 0; blue, 0 }  ,fill opacity=1 ] (479,45) .. controls (479,41.96) and (481.46,39.5) .. (484.5,39.5) .. controls (487.54,39.5) and (490,41.96) .. (490,45) .. controls (490,48.04) and (487.54,50.5) .. (484.5,50.5) .. controls (481.46,50.5) and (479,48.04) .. (479,45) -- cycle ;
\draw [color={rgb, 255:red, 100; green, 100; blue, 100 }  ,draw opacity=1 ] [dash pattern={on 0.84pt off 2.51pt}]  (484.5,50.5) -- (458.26,150.6) ;
\draw [shift={(457.5,153.5)}, rotate = 284.69] [fill={rgb, 255:red, 100; green, 100; blue, 100 }  ,fill opacity=1 ][line width=0.08]  [draw opacity=0] (10.72,-5.15) -- (0,0) -- (10.72,5.15) -- (7.12,0) -- cycle    ;
\draw [color={rgb, 255:red, 100; green, 100; blue, 100 }  ,draw opacity=1 ] [dash pattern={on 0.84pt off 2.51pt}]  (484.5,50.5) -- (508.22,138.1) ;
\draw [shift={(509,141)}, rotate = 254.85] [fill={rgb, 255:red, 100; green, 100; blue, 100 }  ,fill opacity=1 ][line width=0.08]  [draw opacity=0] (10.72,-5.15) -- (0,0) -- (10.72,5.15) -- (7.12,0) -- cycle    ;
\draw [color={rgb, 255:red, 100; green, 100; blue, 100 }  ,draw opacity=1 ] [dash pattern={on 0.84pt off 2.51pt}]  (485,61) -- (464.84,236.52) ;
\draw [shift={(464.5,239.5)}, rotate = 276.55] [fill={rgb, 255:red, 100; green, 100; blue, 100 }  ,fill opacity=1 ][line width=0.08]  [draw opacity=0] (10.72,-5.15) -- (0,0) -- (10.72,5.15) -- (7.12,0) -- cycle    ;
\draw [color={rgb, 255:red, 100; green, 100; blue, 100 }  ,draw opacity=1 ]   (407,151.19) -- (510,149) ;
\draw [shift={(404,151.25)}, rotate = 358.78] [fill={rgb, 255:red, 100; green, 100; blue, 100 }  ,fill opacity=1 ][line width=0.08]  [draw opacity=0] (10.72,-5.15) -- (0,0) -- (10.72,5.15) -- (7.12,0) -- cycle    ;
\draw [color={rgb, 255:red, 100; green, 100; blue, 100 }  ,draw opacity=1 ]   (349,161.19) -- (452,159) ;
\draw [shift={(346,161.25)}, rotate = 358.78] [fill={rgb, 255:red, 100; green, 100; blue, 100 }  ,fill opacity=1 ][line width=0.08]  [draw opacity=0] (10.72,-5.15) -- (0,0) -- (10.72,5.15) -- (7.12,0) -- cycle    ;
\draw [color={rgb, 255:red, 100; green, 100; blue, 100 }  ,draw opacity=1 ]   (356,247.19) -- (459,245) ;
\draw [shift={(353,247.25)}, rotate = 358.78] [fill={rgb, 255:red, 100; green, 100; blue, 100 }  ,fill opacity=1 ][line width=0.08]  [draw opacity=0] (10.72,-5.15) -- (0,0) -- (10.72,5.15) -- (7.12,0) -- cycle    ;
\draw  [draw opacity=0][fill={rgb, 255:red, 245; green, 166; blue, 35 }  ,fill opacity=0.13 ] (324,9.6) -- (417,9.6) -- (417,273.33) -- (324,273.33) -- cycle ;
\draw [color={rgb, 255:red, 100; green, 100; blue, 100 }  ,draw opacity=1 ] [dash pattern={on 4.5pt off 4.5pt}]  (348.92,212) -- (461,212) ;
\draw [shift={(345.92,212)}, rotate = 0] [fill={rgb, 255:red, 100; green, 100; blue, 100 }  ,fill opacity=1 ][line width=0.08]  [draw opacity=0] (10.72,-5.15) -- (0,0) -- (10.72,5.15) -- (7.12,0) -- cycle    ;

\draw (77,20.75) node  [xscale=0.8,yscale=0.8] [align=left] {\begin{minipage}[lt]{54.4pt}\setlength\topsep{0pt}
	Scenario S
	\end{minipage}};
\draw (10,201.83) node  [color={rgb, 255:red, 0; green, 0; blue, 0 }  ,opacity=1 ,xscale=0.8,yscale=0.8] [align=left] {\begin{minipage}[lt]{61.54pt}\setlength\topsep{0pt}
	\textcolor[rgb]{1,0,0}{(1) Define objects and morphisms}
	\end{minipage}};
\draw (25.38,91.22) node  [xscale=0.8,yscale=0.8] [align=left] {\begin{minipage}[lt]{68.29pt}\setlength\topsep{0pt}
	\textcolor[rgb]{1,0,0}{(2) Take limit}
	\end{minipage}};
\draw (208,20.5) node  [xscale=0.8,yscale=0.8] [align=left] {\begin{minipage}[lt]{65.96pt}\setlength\topsep{0pt}
	Scenario $\displaystyle S^{\lor }$
	\end{minipage}};
\draw (140.75,33.5) node  [xscale=0.8,yscale=0.8] [align=left] {\begin{minipage}[lt]{37.06pt}\setlength\topsep{0pt}
	\textcolor[rgb]{1,0,0}{(3) Lift}
	\end{minipage}};
\draw (39.5,147.75) node  [xscale=0.8,yscale=0.8] [align=left] {\begin{minipage}[lt]{21.08pt}\setlength\topsep{0pt}
	$\displaystyle X$
	\end{minipage}};
\draw (137.56,58.33) node  [xscale=0.8,yscale=0.8] [align=left] {\begin{minipage}[lt]{17.98pt}\setlength\topsep{0pt}
	$\displaystyle k_{\mathcal{W}}$
	\end{minipage}};
\draw (72,259.25) node  [xscale=0.8,yscale=0.8] [align=left] {\begin{minipage}[lt]{21.08pt}\setlength\topsep{0pt}
	$\displaystyle Y$
	\end{minipage}};
\draw (123,137.75) node  [xscale=0.8,yscale=0.8] [align=left] {\begin{minipage}[lt]{21.08pt}\setlength\topsep{0pt}
	$\displaystyle Z$
	\end{minipage}};
\draw (120.75,234.5) node  [xscale=0.8,yscale=0.8] [align=left] {\begin{minipage}[lt]{37.06pt}\setlength\topsep{0pt}
	\textcolor[rgb]{1,0,0}{(3) Lift}
	\end{minipage}};
\draw (139,173) node  [xscale=0.8,yscale=0.8] [align=left] {\begin{minipage}[lt]{21.08pt}\setlength\topsep{0pt}
	$\displaystyle k_{\mathcal{W}}( X)$
	\end{minipage}};
\draw (233.5,133.5) node  [xscale=0.8,yscale=0.8] [align=left] {\begin{minipage}[lt]{21.08pt}\setlength\topsep{0pt}
	$\displaystyle k_{\mathcal{W}}( Z)$
	\end{minipage}};
\draw (172,258) node  [xscale=0.8,yscale=0.8] [align=left] {\begin{minipage}[lt]{21.08pt}\setlength\topsep{0pt}
	$\displaystyle k_{\mathcal{W}}( Y)$
	\end{minipage}};
\draw (371,21.5) node  [xscale=0.8,yscale=0.8] [align=left] {\begin{minipage}[lt]{54.4pt}\setlength\topsep{0pt}
	Scenario S
	\end{minipage}};
\draw (511.18,21.25) node  [xscale=0.8,yscale=0.8] [align=left] {\begin{minipage}[lt]{74.37pt}\setlength\topsep{0pt}
	Scenario $\displaystyle F_{\theta }( S)$
	\end{minipage}};
\draw (350.5,148.5) node  [xscale=0.8,yscale=0.8] [align=left] {\begin{minipage}[lt]{21.08pt}\setlength\topsep{0pt}
	$\displaystyle X$
	\end{minipage}};
\draw (428.56,59.08) node  [xscale=0.8,yscale=0.8] [align=left] {\begin{minipage}[lt]{17.98pt}\setlength\topsep{0pt}
	$\displaystyle F_{\theta }$
	\end{minipage}};
\draw (363,260) node  [xscale=0.8,yscale=0.8] [align=left] {\begin{minipage}[lt]{21.08pt}\setlength\topsep{0pt}
	$\displaystyle Y$
	\end{minipage}};
\draw (414,138.5) node  [xscale=0.8,yscale=0.8] [align=left] {\begin{minipage}[lt]{21.08pt}\setlength\topsep{0pt}
	$\displaystyle Z$
	\end{minipage}};
\draw (435,174.25) node  [xscale=0.8,yscale=0.8] [align=left] {\begin{minipage}[lt]{21.08pt}\setlength\topsep{0pt}
	$\displaystyle F_{\theta }( X)$
	\end{minipage}};
\draw (528.5,134.25) node  [xscale=0.8,yscale=0.8] [align=left] {\begin{minipage}[lt]{21.08pt}\setlength\topsep{0pt}
	$\displaystyle F_{\theta }( Z)$
	\end{minipage}};
\draw (463,258.75) node  [xscale=0.8,yscale=0.8] [align=left] {\begin{minipage}[lt]{21.08pt}\setlength\topsep{0pt}
	$\displaystyle F_{\theta }( Y)$
	\end{minipage}};
\draw (426.88,34.46) node  [color={rgb, 255:red, 0; green, 0; blue, 0 }  ,opacity=1 ,xscale=0.8,yscale=0.8] [align=left] {\begin{minipage}[lt]{32.81pt}\setlength\topsep{0pt}
	\textcolor[rgb]{1,0,0}{(1) Lift}
	\end{minipage}};
\draw (530.71,80.79) node  [color={rgb, 255:red, 0; green, 0; blue, 0 }  ,opacity=1 ,xscale=0.8,yscale=0.8] [align=left] {\begin{minipage}[lt]{55.25pt}\setlength\topsep{0pt}
	\textcolor[rgb]{1,0,0}{(2) Extract objects}
	\end{minipage}};
\draw (523.88,218.46) node  [color={rgb, 255:red, 0; green, 0; blue, 0 }  ,opacity=1 ,xscale=0.8,yscale=0.8] [align=left] {\begin{minipage}[lt]{64.09pt}\setlength\topsep{0pt}
	\textcolor[rgb]{1,0,0}{(3) Compute morphisms}
	\end{minipage}};
\draw (419.48,199.37) node  [color={rgb, 255:red, 0; green, 0; blue, 0 }  ,opacity=1 ,xscale=0.8,yscale=0.8] [align=left] {\begin{minipage}[lt]{68.99pt}\setlength\topsep{0pt}
	\textcolor[rgb]{1,0,0}{(4) Determine}
	\end{minipage}};
\draw (80,284.67) node  [xscale=0.8,yscale=0.8] [align=left] {\begin{minipage}[lt]{14.96pt}\setlength\topsep{0pt}
	$\displaystyle \mathcal{W}$
	\end{minipage}};
\draw (203,285) node  [xscale=0.8,yscale=0.8] [align=left] {\begin{minipage}[lt]{14.96pt}\setlength\topsep{0pt}
	$\displaystyle \mathcal{W}^{\lor }$
	\end{minipage}};
\draw (372,284.67) node  [xscale=0.8,yscale=0.8] [align=left] {\begin{minipage}[lt]{14.96pt}\setlength\topsep{0pt}
	$\displaystyle \mathcal{W}$
	\end{minipage}};
\draw (499,285) node  [xscale=0.8,yscale=0.8] [align=left] {\begin{minipage}[lt]{14.96pt}\setlength\topsep{0pt}
	$\displaystyle \mathcal{W}^{\lor }$
	\end{minipage}};

\end{tikzpicture}

%% file: 14.tex
\tikzset{every picture/.style={line width=0.75pt}} 

\begin{tikzpicture}[x=0.75pt,y=0.75pt,yscale=-1,xscale=1]

\draw  [draw opacity=0][fill={rgb, 255:red, 0; green, 0; blue, 0 }  ,fill opacity=1 ] (161,158.25) .. controls (161,155.21) and (163.46,152.75) .. (166.5,152.75) .. controls (169.54,152.75) and (172,155.21) .. (172,158.25) .. controls (172,161.29) and (169.54,163.75) .. (166.5,163.75) .. controls (163.46,163.75) and (161,161.29) .. (161,158.25) -- cycle ;
\draw  [draw opacity=0][fill={rgb, 255:red, 0; green, 0; blue, 0 }  ,fill opacity=1 ] (215.87,157.6) .. controls (215.87,154.56) and (218.33,152.1) .. (221.37,152.1) .. controls (224.4,152.1) and (226.87,154.56) .. (226.87,157.6) .. controls (226.87,160.64) and (224.4,163.1) .. (221.37,163.1) .. controls (218.33,163.1) and (215.87,160.64) .. (215.87,157.6) -- cycle ;
\draw  [draw opacity=0][fill={rgb, 255:red, 0; green, 0; blue, 0 }  ,fill opacity=1 ] (168,244.25) .. controls (168,241.21) and (170.46,238.75) .. (173.5,238.75) .. controls (176.54,238.75) and (179,241.21) .. (179,244.25) .. controls (179,247.29) and (176.54,249.75) .. (173.5,249.75) .. controls (170.46,249.75) and (168,247.29) .. (168,244.25) -- cycle ;
\draw [color={rgb, 255:red, 100; green, 100; blue, 100 }  ,draw opacity=1 ]   (173.5,238.75) -- (219.76,165.64) ;
\draw [shift={(221.37,163.1)}, rotate = 122.32] [fill={rgb, 255:red, 100; green, 100; blue, 100 }  ,fill opacity=1 ][line width=0.08]  [draw opacity=0] (10.72,-5.15) -- (0,0) -- (10.72,5.15) -- (7.12,0) -- cycle    ;
\draw [color={rgb, 255:red, 100; green, 100; blue, 100 }  ,draw opacity=1 ]   (173.5,238.75) -- (166.78,166.74) ;
\draw [shift={(166.5,163.75)}, rotate = 84.67] [fill={rgb, 255:red, 100; green, 100; blue, 100 }  ,fill opacity=1 ][line width=0.08]  [draw opacity=0] (10.72,-5.15) -- (0,0) -- (10.72,5.15) -- (7.12,0) -- cycle    ;
\draw [color={rgb, 255:red, 100; green, 100; blue, 100 }  ,draw opacity=1 ]   (215.87,157.6) -- (175,158.21) ;
\draw [shift={(172,158.25)}, rotate = 359.15] [fill={rgb, 255:red, 100; green, 100; blue, 100 }  ,fill opacity=1 ][line width=0.08]  [draw opacity=0] (10.72,-5.15) -- (0,0) -- (10.72,5.15) -- (7.12,0) -- cycle    ;
\draw  [draw opacity=0][fill={rgb, 255:red, 0; green, 0; blue, 0 }  ,fill opacity=1 ] (188,44.25) .. controls (188,41.21) and (190.46,38.75) .. (193.5,38.75) .. controls (196.54,38.75) and (199,41.21) .. (199,44.25) .. controls (199,47.29) and (196.54,49.75) .. (193.5,49.75) .. controls (190.46,49.75) and (188,47.29) .. (188,44.25) -- cycle ;
\draw  [draw opacity=0][fill={rgb, 255:red, 0; green, 0; blue, 0 }  ,fill opacity=1 ] (336,72.92) .. controls (336,69.88) and (338.46,67.42) .. (341.5,67.42) .. controls (344.54,67.42) and (347,69.88) .. (347,72.92) .. controls (347,75.95) and (344.54,78.42) .. (341.5,78.42) .. controls (338.46,78.42) and (336,75.95) .. (336,72.92) -- cycle ;
\draw [color={rgb, 255:red, 147; green, 237; blue, 177 }  ,draw opacity=1 ][fill={rgb, 255:red, 147; green, 237; blue, 177 }  ,fill opacity=1 ]   (335.2,76.93) -- (172.72,152.32) ;
\draw [shift={(170,153.58)}, rotate = 335.11] [fill={rgb, 255:red, 147; green, 237; blue, 177 }  ,fill opacity=1 ][line width=0.08]  [draw opacity=0] (10.72,-5.15) -- (0,0) -- (10.72,5.15) -- (7.12,0) -- cycle    ;
\draw [color={rgb, 255:red, 147; green, 237; blue, 177 }  ,draw opacity=1 ][fill={rgb, 255:red, 147; green, 237; blue, 177 }  ,fill opacity=1 ]   (338.53,80.27) -- (180.65,238.81) ;
\draw [shift={(178.53,240.93)}, rotate = 314.88] [fill={rgb, 255:red, 147; green, 237; blue, 177 }  ,fill opacity=1 ][line width=0.08]  [draw opacity=0] (10.72,-5.15) -- (0,0) -- (10.72,5.15) -- (7.12,0) -- cycle    ;
\draw [color={rgb, 255:red, 147; green, 237; blue, 177 }  ,draw opacity=1 ][fill={rgb, 255:red, 147; green, 237; blue, 177 }  ,fill opacity=1 ]   (336.2,78.93) -- (230.35,150.58) ;
\draw [shift={(227.87,152.27)}, rotate = 325.91] [fill={rgb, 255:red, 147; green, 237; blue, 177 }  ,fill opacity=1 ][line width=0.08]  [draw opacity=0] (10.72,-5.15) -- (0,0) -- (10.72,5.15) -- (7.12,0) -- cycle    ;
\draw  [draw opacity=0][fill={rgb, 255:red, 147; green, 237; blue, 177 }  ,fill opacity=1 ] (233.55,115.26) -- (250.48,115.26) -- (250.48,124.93) -- (233.55,124.93) -- cycle ;
\draw  [draw opacity=0][fill={rgb, 255:red, 147; green, 237; blue, 177 }  ,fill opacity=1 ] (270.88,113.26) -- (287.81,113.26) -- (287.81,122.93) -- (270.88,122.93) -- cycle ;
\draw  [draw opacity=0][fill={rgb, 255:red, 147; green, 237; blue, 177 }  ,fill opacity=1 ] (250.07,155.76) -- (267,155.76) -- (267,165.44) -- (250.07,165.44) -- cycle ;
\draw [color={rgb, 255:red, 100; green, 100; blue, 100 }  ,draw opacity=1 ]   (255.2,154.27) -- (243.06,127.15) ;
\draw [shift={(241.83,124.42)}, rotate = 65.88] [fill={rgb, 255:red, 100; green, 100; blue, 100 }  ,fill opacity=1 ][line width=0.08]  [draw opacity=0] (10.72,-5.15) -- (0,0) -- (10.72,5.15) -- (7.12,0) -- cycle    ;
\draw [color={rgb, 255:red, 100; green, 100; blue, 100 }  ,draw opacity=1 ]   (261.2,154.93) -- (277.81,127.66) ;
\draw [shift={(279.37,125.1)}, rotate = 121.34] [fill={rgb, 255:red, 100; green, 100; blue, 100 }  ,fill opacity=1 ][line width=0.08]  [draw opacity=0] (10.72,-5.15) -- (0,0) -- (10.72,5.15) -- (7.12,0) -- cycle    ;
\draw [color={rgb, 255:red, 100; green, 100; blue, 100 }  ,draw opacity=1 ] [dash pattern={on 0.84pt off 2.51pt}]  (188.53,49.6) -- (167.13,149.82) ;
\draw [shift={(166.5,152.75)}, rotate = 282.06] [fill={rgb, 255:red, 100; green, 100; blue, 100 }  ,fill opacity=1 ][line width=0.08]  [draw opacity=0] (10.72,-5.15) -- (0,0) -- (10.72,5.15) -- (7.12,0) -- cycle    ;
\draw [color={rgb, 255:red, 100; green, 100; blue, 100 }  ,draw opacity=1 ] [dash pattern={on 0.84pt off 2.51pt}]  (197.2,48.93) -- (220.68,149.18) ;
\draw [shift={(221.37,152.1)}, rotate = 256.82] [fill={rgb, 255:red, 100; green, 100; blue, 100 }  ,fill opacity=1 ][line width=0.08]  [draw opacity=0] (10.72,-5.15) -- (0,0) -- (10.72,5.15) -- (7.12,0) -- cycle    ;
\draw [color={rgb, 255:red, 100; green, 100; blue, 100 }  ,draw opacity=1 ] [dash pattern={on 0.84pt off 2.51pt}]  (193.5,49.75) -- (173.82,235.77) ;
\draw [shift={(173.5,238.75)}, rotate = 276.04] [fill={rgb, 255:red, 100; green, 100; blue, 100 }  ,fill opacity=1 ][line width=0.08]  [draw opacity=0] (10.72,-5.15) -- (0,0) -- (10.72,5.15) -- (7.12,0) -- cycle    ;
\draw [color={rgb, 255:red, 100; green, 100; blue, 100 }  ,draw opacity=1 ]   (269.2,118.93) -- (252.86,119.5) ;
\draw [shift={(249.87,119.6)}, rotate = 358.03] [fill={rgb, 255:red, 100; green, 100; blue, 100 }  ,fill opacity=1 ][line width=0.08]  [draw opacity=0] (10.72,-5.15) -- (0,0) -- (10.72,5.15) -- (7.12,0) -- cycle    ;
\draw [color={rgb, 255:red, 147; green, 237; blue, 177 }  ,draw opacity=1 ]   (334,70.92) -- (201.94,44.83) ;
\draw [shift={(199,44.25)}, rotate = 11.17] [fill={rgb, 255:red, 147; green, 237; blue, 177 }  ,fill opacity=1 ][line width=0.08]  [draw opacity=0] (10.72,-5.15) -- (0,0) -- (10.72,5.15) -- (7.12,0) -- cycle    ;
\draw  [draw opacity=0][fill={rgb, 255:red, 141; green, 154; blue, 255 }  ,fill opacity=1 ] (289.97,59.92) -- (306.9,59.92) -- (306.9,69.6) -- (289.97,69.6) -- cycle ;
\draw  [draw opacity=0][fill={rgb, 255:red, 141; green, 154; blue, 255 }  ,fill opacity=1 ] (245.27,59.84) -- (262.2,59.84) -- (262.2,69.52) -- (245.27,69.52) -- cycle ;
\draw [color={rgb, 255:red, 100; green, 100; blue, 100 }  ,draw opacity=1 ] [dash pattern={on 0.84pt off 2.51pt}]  (249.2,72.27) -- (243,111.3) ;
\draw [shift={(242.53,114.27)}, rotate = 279.02] [fill={rgb, 255:red, 100; green, 100; blue, 100 }  ,fill opacity=1 ][line width=0.08]  [draw opacity=0] (10.72,-5.15) -- (0,0) -- (10.72,5.15) -- (7.12,0) -- cycle    ;
\draw [color={rgb, 255:red, 100; green, 100; blue, 100 }  ,draw opacity=1 ] [dash pattern={on 0.84pt off 2.51pt}]  (254.53,72.93) -- (260.96,151.94) ;
\draw [shift={(261.2,154.93)}, rotate = 265.35] [fill={rgb, 255:red, 100; green, 100; blue, 100 }  ,fill opacity=1 ][line width=0.08]  [draw opacity=0] (10.72,-5.15) -- (0,0) -- (10.72,5.15) -- (7.12,0) -- cycle    ;
\draw [color={rgb, 255:red, 100; green, 100; blue, 100 }  ,draw opacity=1 ] [dash pattern={on 0.84pt off 2.51pt}]  (257.87,71.6) -- (278.11,109.29) ;
\draw [shift={(279.53,111.93)}, rotate = 241.76] [fill={rgb, 255:red, 100; green, 100; blue, 100 }  ,fill opacity=1 ][line width=0.08]  [draw opacity=0] (10.72,-5.15) -- (0,0) -- (10.72,5.15) -- (7.12,0) -- cycle    ;
\draw [color={rgb, 255:red, 255; green, 44; blue, 0 }  ,draw opacity=1 ][fill={rgb, 255:red, 255; green, 44; blue, 0 }  ,fill opacity=1 ][line width=2.25]    (289.87,64.58) -- (262.67,64.58) ;

\draw (194.68,20.5) node  [xscale=0.8,yscale=0.8] [align=left] {\begin{minipage}[lt]{29.72pt}\setlength\topsep{0pt}
	$\displaystyle F_{\theta }( S)$
	\end{minipage}};
\draw (132.33,167) node  [xscale=0.8,yscale=0.8] [align=left] {\begin{minipage}[lt]{21.08pt}\setlength\topsep{0pt}
	$\displaystyle k_{\mathcal{W}}( X)$
	\end{minipage}};
\draw (198.17,165.5) node  [xscale=0.8,yscale=0.8] [align=left] {\begin{minipage}[lt]{21.08pt}\setlength\topsep{0pt}
	$\displaystyle k_{\mathcal{W}}( Z)$
	\end{minipage}};
\draw (172,258) node  [xscale=0.8,yscale=0.8] [align=left] {\begin{minipage}[lt]{21.08pt}\setlength\topsep{0pt}
	$\displaystyle k_{\mathcal{W}}( Y)$
	\end{minipage}};
\draw (203,285) node  [xscale=0.8,yscale=0.8] [align=left] {\begin{minipage}[lt]{14.96pt}\setlength\topsep{0pt}
	$\displaystyle \mathcal{W}^{\lor }$
	\end{minipage}};
\draw (342.02,54.5) node  [xscale=0.8,yscale=0.8] [align=left] {\begin{minipage}[lt]{12.49pt}\setlength\topsep{0pt}
	$\displaystyle T$
	\end{minipage}};
\draw (282.28,49.71) node  [font=\large,xscale=0.8,yscale=0.8] [align=left] {\begin{minipage}[lt]{20.09pt}\setlength\topsep{0pt}
	{\large  \textcolor[rgb]{1,0,0}{{$\simeq $}}}
	\end{minipage}};
\draw (297.96,51.17) node  [xscale=0.8,yscale=0.8] [align=left] {\begin{minipage}[lt]{12.49pt}\setlength\topsep{0pt}
	$\displaystyle P_{1}$
	\end{minipage}};
\draw (299.08,179.35) node  [xscale=0.8,yscale=0.8] [align=left] {\begin{minipage}[lt]{92.31pt}\setlength\topsep{0pt}
	$\displaystyle Hom_{\mathcal{W}^{\lor }}( T,k_{\mathcal{W}}( Y))$
	\end{minipage}};
\draw (252.96,50.92) node  [xscale=0.8,yscale=0.8] [align=left] {\begin{minipage}[lt]{12.49pt}\setlength\topsep{0pt}
	$\displaystyle P_{2}$
	\end{minipage}};

\end{tikzpicture}

%% file: 6.tex
\tikzset{every picture/.style={line width=0.75pt}} 

\begin{tikzpicture}[x=0.75pt,y=0.75pt,yscale=-1,xscale=1, scale=2]

\draw  [draw opacity=0][fill={rgb, 255:red, 255; green, 0; blue, 0 }  ,fill opacity=1 ] (102.23,59.48) .. controls (102.23,56.69) and (104.5,54.42) .. (107.3,54.42) .. controls (110.1,54.42) and (112.37,56.69) .. (112.37,59.48) .. controls (112.37,62.28) and (110.1,64.55) .. (107.3,64.55) .. controls (104.5,64.55) and (102.23,62.28) .. (102.23,59.48) -- cycle ;
\draw  [draw opacity=0][fill={rgb, 255:red, 3; green, 47; blue, 255 }  ,fill opacity=1 ] (127.5,74.43) .. controls (127.5,71.64) and (129.77,69.37) .. (132.57,69.37) .. controls (135.36,69.37) and (137.63,71.64) .. (137.63,74.43) .. controls (137.63,77.23) and (135.36,79.5) .. (132.57,79.5) .. controls (129.77,79.5) and (127.5,77.23) .. (127.5,74.43) -- cycle ;
\draw  [draw opacity=0][fill={rgb, 255:red, 0; green, 255; blue, 38 }  ,fill opacity=1 ] (110,91.43) .. controls (110,88.64) and (112.27,86.37) .. (115.07,86.37) .. controls (117.86,86.37) and (120.13,88.64) .. (120.13,91.43) .. controls (120.13,94.23) and (117.86,96.5) .. (115.07,96.5) .. controls (112.27,96.5) and (110,94.23) .. (110,91.43) -- cycle ;
\draw  [draw opacity=0][fill={rgb, 255:red, 222; green, 0; blue, 255 }  ,fill opacity=1 ] (84.5,79.43) .. controls (84.5,76.64) and (86.77,74.37) .. (89.57,74.37) .. controls (92.36,74.37) and (94.63,76.64) .. (94.63,79.43) .. controls (94.63,82.23) and (92.36,84.5) .. (89.57,84.5) .. controls (86.77,84.5) and (84.5,82.23) .. (84.5,79.43) -- cycle ;
\draw [color={rgb, 255:red, 100; green, 100; blue, 100 }  ,draw opacity=1 ]   (111.8,61.05) -- (124.78,69.42) ;
\draw [shift={(127.3,71.05)}, rotate = 212.83] [fill={rgb, 255:red, 100; green, 100; blue, 100 }  ,fill opacity=1 ][line width=0.08]  [draw opacity=0] (5.36,-2.57) -- (0,0) -- (5.36,2.57) -- (3.56,0) -- cycle    ;
\draw [color={rgb, 255:red, 100; green, 100; blue, 100 }  ,draw opacity=1 ]   (107.3,64.55) -- (111.16,82.12) ;
\draw [shift={(111.8,85.05)}, rotate = 257.62] [fill={rgb, 255:red, 100; green, 100; blue, 100 }  ,fill opacity=1 ][line width=0.08]  [draw opacity=0] (5.36,-2.57) -- (0,0) -- (5.36,2.57) -- (3.56,0) -- cycle    ;
\draw [color={rgb, 255:red, 100; green, 100; blue, 100 }  ,draw opacity=1 ]   (126.51,81.49) -- (118.8,88.05) ;
\draw [shift={(128.8,79.55)}, rotate = 139.64] [fill={rgb, 255:red, 100; green, 100; blue, 100 }  ,fill opacity=1 ][line width=0.08]  [draw opacity=0] (5.36,-2.57) -- (0,0) -- (5.36,2.57) -- (3.56,0) -- cycle    ;
\draw [color={rgb, 255:red, 100; green, 100; blue, 100 }  ,draw opacity=1 ]   (101.8,63.05) -- (95.94,72.04) ;
\draw [shift={(94.3,74.55)}, rotate = 303.11] [fill={rgb, 255:red, 100; green, 100; blue, 100 }  ,fill opacity=1 ][line width=0.08]  [draw opacity=0] (5.36,-2.57) -- (0,0) -- (5.36,2.57) -- (3.56,0) -- cycle    ;
\draw  [draw opacity=0][fill={rgb, 255:red, 139; green, 87; blue, 42 }  ,fill opacity=1 ] (152.83,77.1) .. controls (152.83,74.3) and (155.1,72.03) .. (157.9,72.03) .. controls (160.7,72.03) and (162.97,74.3) .. (162.97,77.1) .. controls (162.97,79.9) and (160.7,82.17) .. (157.9,82.17) .. controls (155.1,82.17) and (152.83,79.9) .. (152.83,77.1) -- cycle ;
\draw [color={rgb, 255:red, 100; green, 100; blue, 100 }  ,draw opacity=1 ]   (137.63,74.43) -- (149.88,76.58) ;
\draw [shift={(152.83,77.1)}, rotate = 189.95] [fill={rgb, 255:red, 100; green, 100; blue, 100 }  ,fill opacity=1 ][line width=0.08]  [draw opacity=0] (5.36,-2.57) -- (0,0) -- (5.36,2.57) -- (3.56,0) -- cycle    ;
\draw  [draw opacity=0][fill={rgb, 255:red, 144; green, 19; blue, 254 }  ,fill opacity=1 ] (134.83,100.43) .. controls (134.83,97.64) and (137.1,95.37) .. (139.9,95.37) .. controls (142.7,95.37) and (144.97,97.64) .. (144.97,100.43) .. controls (144.97,103.23) and (142.7,105.5) .. (139.9,105.5) .. controls (137.1,105.5) and (134.83,103.23) .. (134.83,100.43) -- cycle ;
\draw  [draw opacity=0][fill={rgb, 255:red, 80; green, 227; blue, 194 }  ,fill opacity=1 ] (155.5,101.1) .. controls (155.5,98.3) and (157.77,96.03) .. (160.57,96.03) .. controls (163.36,96.03) and (165.63,98.3) .. (165.63,101.1) .. controls (165.63,103.9) and (163.36,106.17) .. (160.57,106.17) .. controls (157.77,106.17) and (155.5,103.9) .. (155.5,101.1) -- cycle ;
\draw  [draw opacity=0][fill={rgb, 255:red, 248; green, 231; blue, 28 }  ,fill opacity=1 ] (173.5,98.43) .. controls (173.5,95.64) and (175.77,93.37) .. (178.57,93.37) .. controls (181.36,93.37) and (183.63,95.64) .. (183.63,98.43) .. controls (183.63,101.23) and (181.36,103.5) .. (178.57,103.5) .. controls (175.77,103.5) and (173.5,101.23) .. (173.5,98.43) -- cycle ;
\draw [color={rgb, 255:red, 100; green, 100; blue, 100 }  ,draw opacity=1 ]   (154.53,81.33) -- (145.68,92.95) ;
\draw [shift={(143.87,95.33)}, rotate = 307.3] [fill={rgb, 255:red, 100; green, 100; blue, 100 }  ,fill opacity=1 ][line width=0.08]  [draw opacity=0] (5.36,-2.57) -- (0,0) -- (5.36,2.57) -- (3.56,0) -- cycle    ;
\draw [color={rgb, 255:red, 100; green, 100; blue, 100 }  ,draw opacity=1 ]   (158.73,80.3) -- (160.22,93.05) ;
\draw [shift={(160.57,96.03)}, rotate = 263.35] [fill={rgb, 255:red, 100; green, 100; blue, 100 }  ,fill opacity=1 ][line width=0.08]  [draw opacity=0] (5.36,-2.57) -- (0,0) -- (5.36,2.57) -- (3.56,0) -- cycle    ;
\draw [color={rgb, 255:red, 100; green, 100; blue, 100 }  ,draw opacity=1 ]   (162.23,80.17) -- (173.33,91.8) ;
\draw [shift={(175.4,93.97)}, rotate = 226.35] [fill={rgb, 255:red, 100; green, 100; blue, 100 }  ,fill opacity=1 ][line width=0.08]  [draw opacity=0] (5.36,-2.57) -- (0,0) -- (5.36,2.57) -- (3.56,0) -- cycle    ;

\end{tikzpicture}

%% file: 7.tex
\tikzset{every picture/.style={line width=0.75pt}} 

\begin{tikzpicture}[x=0.75pt,y=0.75pt,yscale=-1,xscale=1]
	
	\draw (91.38,240.05) node  {\includegraphics[width=38.58pt,height=38.58pt]{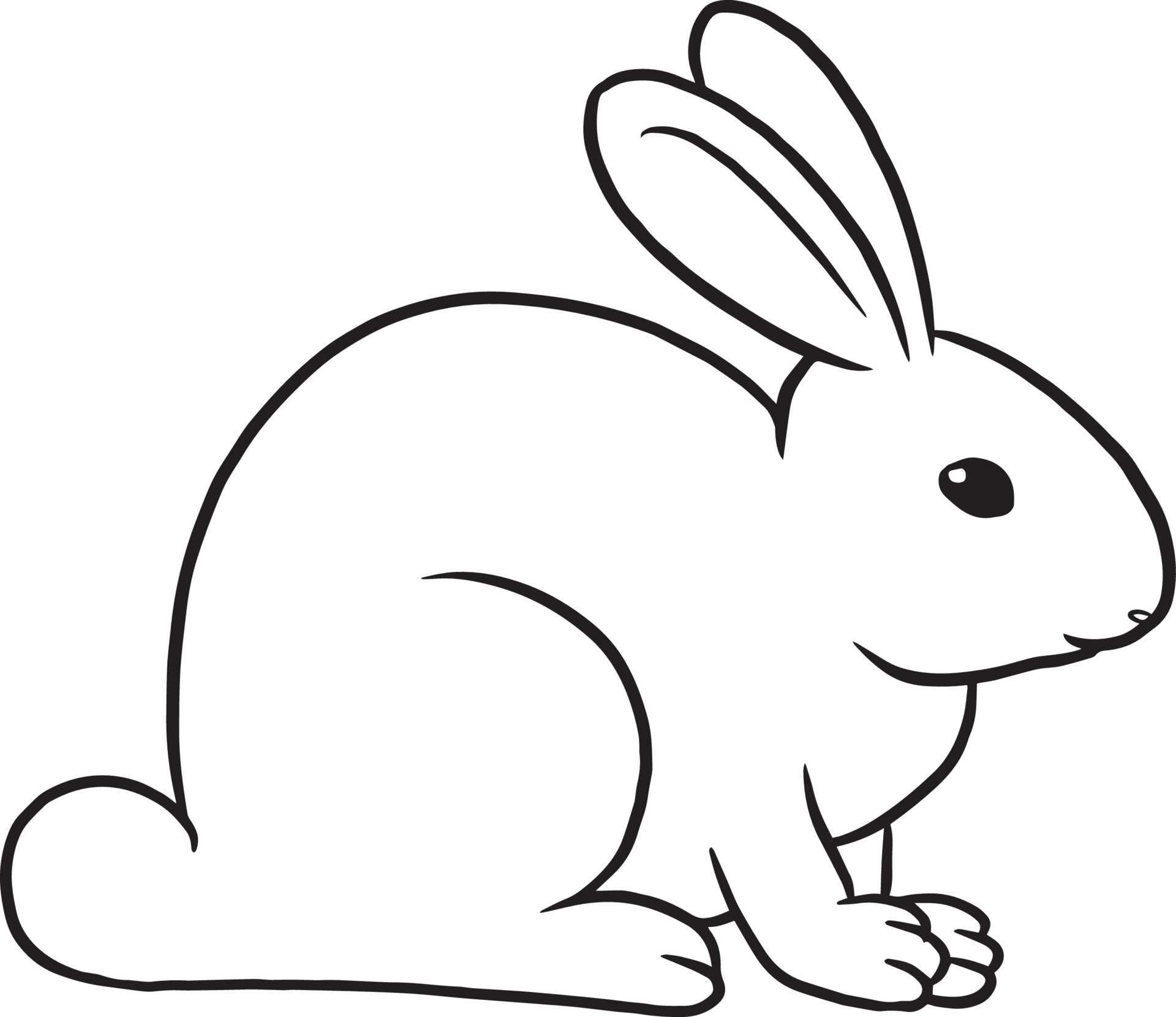}};
	\draw (243.32,237.98) node  {\includegraphics[width=35.68pt,height=35.68pt]{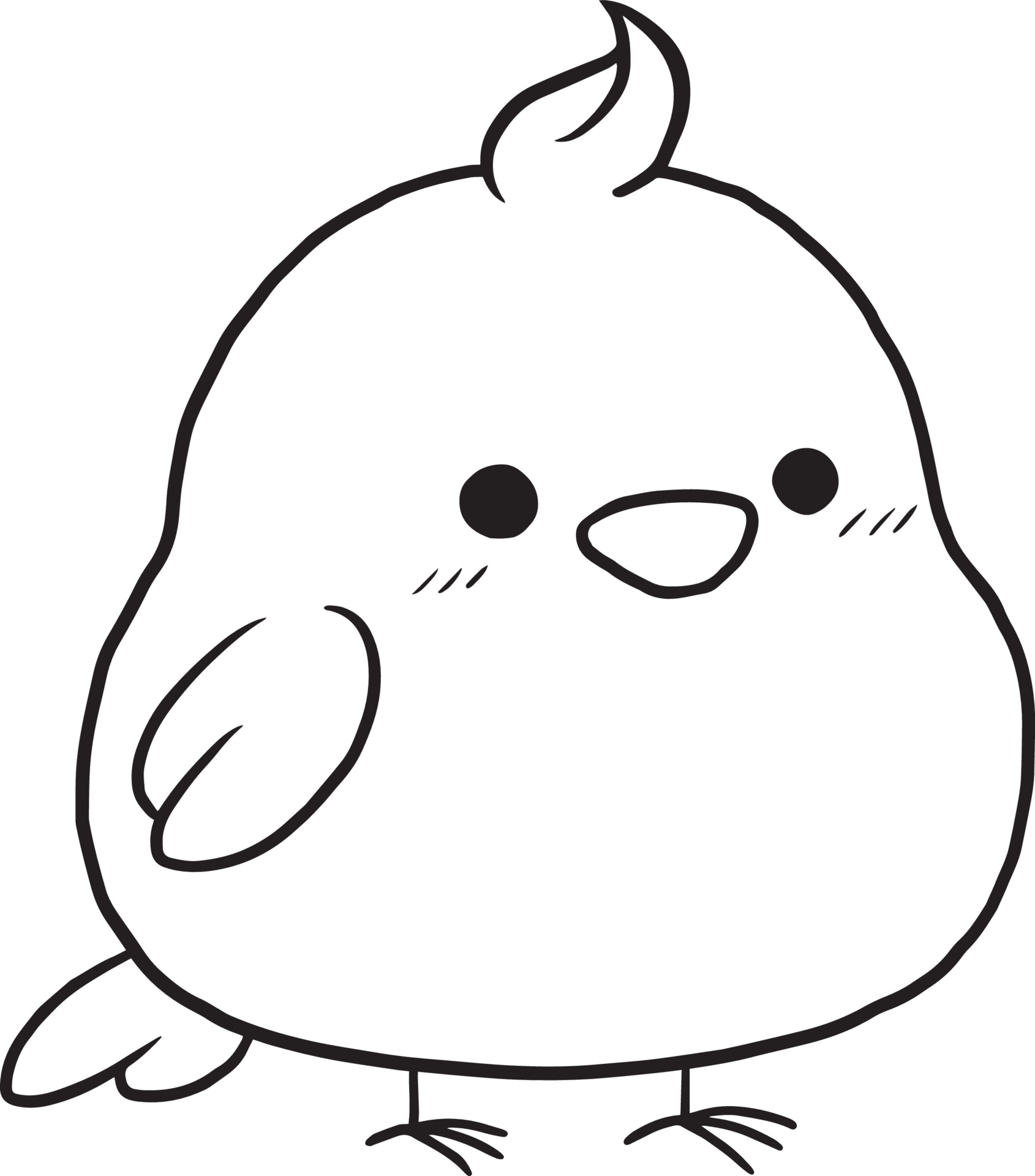}};
	\draw   (168.9,106.54) -- (199.8,133.77) -- (138,133.77) -- cycle ;
	\draw    (188.27,169.66) -- (229.87,208.93) ;
	\draw [shift={(186.09,167.6)}, rotate = 43.35] [fill={rgb, 255:red, 0; green, 0; blue, 0 }  ][line width=0.08]  [draw opacity=0] (10.72,-5.15) -- (0,0) -- (10.72,5.15) -- (7.12,0) -- cycle    ;
	\draw    (150.47,169.47) -- (102.53,207.6) ;
	\draw [shift={(152.82,167.6)}, rotate = 141.5] [fill={rgb, 255:red, 0; green, 0; blue, 0 }  ][line width=0.08]  [draw opacity=0] (10.72,-5.15) -- (0,0) -- (10.72,5.15) -- (7.12,0) -- cycle    ;
	\draw    (41.87,316.67) -- (66.84,274.78) ;
	\draw [shift={(67.87,273.07)}, rotate = 120.81] [color={rgb, 255:red, 0; green, 0; blue, 0 }  ][line width=0.75]    (10.93,-3.29) .. controls (6.95,-1.4) and (3.31,-0.3) .. (0,0) .. controls (3.31,0.3) and (6.95,1.4) .. (10.93,3.29)   ;
	\draw    (138.2,318) -- (113.21,275.06) ;
	\draw [shift={(112.2,273.33)}, rotate = 59.8] [color={rgb, 255:red, 0; green, 0; blue, 0 }  ][line width=0.75]    (10.93,-3.29) .. controls (6.95,-1.4) and (3.31,-0.3) .. (0,0) .. controls (3.31,0.3) and (6.95,1.4) .. (10.93,3.29)   ;
	\draw    (204.87,315.67) -- (229.84,273.78) ;
	\draw [shift={(230.87,272.07)}, rotate = 120.81] [color={rgb, 255:red, 0; green, 0; blue, 0 }  ][line width=0.75]    (10.93,-3.29) .. controls (6.95,-1.4) and (3.31,-0.3) .. (0,0) .. controls (3.31,0.3) and (6.95,1.4) .. (10.93,3.29)   ;
	\draw    (288.2,317) -- (263.21,274.06) ;
	\draw [shift={(262.2,272.33)}, rotate = 59.8] [color={rgb, 255:red, 0; green, 0; blue, 0 }  ][line width=0.75]    (10.93,-3.29) .. controls (6.95,-1.4) and (3.31,-0.3) .. (0,0) .. controls (3.31,0.3) and (6.95,1.4) .. (10.93,3.29)   ;
	\draw    (181.4,106.19) -- (205.87,65.17) ;
	\draw [shift={(179.87,108.77)}, rotate = 300.81] [fill={rgb, 255:red, 0; green, 0; blue, 0 }  ][line width=0.08]  [draw opacity=0] (10.72,-5.15) -- (0,0) -- (10.72,5.15) -- (7.12,0) -- cycle    ;
	\draw    (151.69,107.51) -- (127.2,65.43) ;
	\draw [shift={(153.2,110.1)}, rotate = 239.8] [fill={rgb, 255:red, 0; green, 0; blue, 0 }  ][line width=0.08]  [draw opacity=0] (10.72,-5.15) -- (0,0) -- (10.72,5.15) -- (7.12,0) -- cycle    ;
	\draw   (138,133.77) -- (200,133.77) -- (200,164) -- (138,164) -- cycle ;
	\draw   (288.67,170.95) -- (336.11,170.95) -- (336.11,164.33) -- (367.73,177.57) -- (336.11,190.8) -- (336.11,184.18) -- (288.67,184.18) -- cycle ;
	\draw [color={rgb, 255:red, 0; green, 0; blue, 255 }  ,draw opacity=1 ]   (521.27,120.43) -- (562.87,159.7) ;
	\draw [shift={(519.09,118.37)}, rotate = 43.35] [fill={rgb, 255:red, 0; green, 0; blue, 255 }  ,fill opacity=1 ][line width=0.08]  [draw opacity=0] (10.72,-5.15) -- (0,0) -- (10.72,5.15) -- (7.12,0) -- cycle    ;
	\draw [color={rgb, 255:red, 0; green, 0; blue, 255 }  ,draw opacity=1 ]   (483.47,120.23) -- (435.53,158.37) ;
	\draw [shift={(485.82,118.37)}, rotate = 141.5] [fill={rgb, 255:red, 0; green, 0; blue, 255 }  ,fill opacity=1 ][line width=0.08]  [draw opacity=0] (10.72,-5.15) -- (0,0) -- (10.72,5.15) -- (7.12,0) -- cycle    ;
	\draw [color={rgb, 255:red, 0; green, 0; blue, 255 }  ,draw opacity=1 ]   (456.6,250.37) -- (432.33,188.23) ;
	\draw [shift={(431.6,186.37)}, rotate = 68.66] [color={rgb, 255:red, 0; green, 0; blue, 255 }  ,draw opacity=1 ][line width=0.75]    (10.93,-3.29) .. controls (6.95,-1.4) and (3.31,-0.3) .. (0,0) .. controls (3.31,0.3) and (6.95,1.4) .. (10.93,3.29)   ;
	\draw [color={rgb, 255:red, 0; green, 0; blue, 255 }  ,draw opacity=1 ]   (539.6,258.37) -- (442.16,180.62) ;
	\draw [shift={(440.6,179.37)}, rotate = 38.59] [color={rgb, 255:red, 0; green, 0; blue, 255 }  ,draw opacity=1 ][line width=0.75]    (10.93,-3.29) .. controls (6.95,-1.4) and (3.31,-0.3) .. (0,0) .. controls (3.31,0.3) and (6.95,1.4) .. (10.93,3.29)   ;
	\draw [color={rgb, 255:red, 0; green, 0; blue, 255 }  ,draw opacity=1 ]   (474.6,253.37) -- (557.07,183.66) ;
	\draw [shift={(558.6,182.37)}, rotate = 139.79] [color={rgb, 255:red, 0; green, 0; blue, 255 }  ,draw opacity=1 ][line width=0.75]    (10.93,-3.29) .. controls (6.95,-1.4) and (3.31,-0.3) .. (0,0) .. controls (3.31,0.3) and (6.95,1.4) .. (10.93,3.29)   ;
	\draw [color={rgb, 255:red, 0; green, 0; blue, 255 }  ,draw opacity=1 ]   (550.6,254.37) -- (571,189.28) ;
	\draw [shift={(571.6,187.37)}, rotate = 107.4] [color={rgb, 255:red, 0; green, 0; blue, 255 }  ,draw opacity=1 ][line width=0.75]    (10.93,-3.29) .. controls (6.95,-1.4) and (3.31,-0.3) .. (0,0) .. controls (3.31,0.3) and (6.95,1.4) .. (10.93,3.29)   ;
	\draw  [draw opacity=0][fill={rgb, 255:red, 0; green, 0; blue, 0 }  ,fill opacity=1 ] (495.6,107.8) .. controls (495.6,104.38) and (498.38,101.6) .. (501.8,101.6) .. controls (505.22,101.6) and (508,104.38) .. (508,107.8) .. controls (508,111.22) and (505.22,114) .. (501.8,114) .. controls (498.38,114) and (495.6,111.22) .. (495.6,107.8) -- cycle ;
	\draw  [draw opacity=0][fill={rgb, 255:red, 0; green, 0; blue, 0 }  ,fill opacity=1 ] (423.6,172.8) .. controls (423.6,169.38) and (426.38,166.6) .. (429.8,166.6) .. controls (433.22,166.6) and (436,169.38) .. (436,172.8) .. controls (436,176.22) and (433.22,179) .. (429.8,179) .. controls (426.38,179) and (423.6,176.22) .. (423.6,172.8) -- cycle ;
	\draw  [draw opacity=0][fill={rgb, 255:red, 0; green, 0; blue, 0 }  ,fill opacity=1 ] (562.6,172.8) .. controls (562.6,169.38) and (565.38,166.6) .. (568.8,166.6) .. controls (572.22,166.6) and (575,169.38) .. (575,172.8) .. controls (575,176.22) and (572.22,179) .. (568.8,179) .. controls (565.38,179) and (562.6,176.22) .. (562.6,172.8) -- cycle ;
	\draw  [draw opacity=0][fill={rgb, 255:red, 0; green, 0; blue, 0 }  ,fill opacity=1 ] (455.6,264.8) .. controls (455.6,261.38) and (458.38,258.6) .. (461.8,258.6) .. controls (465.22,258.6) and (468,261.38) .. (468,264.8) .. controls (468,268.22) and (465.22,271) .. (461.8,271) .. controls (458.38,271) and (455.6,268.22) .. (455.6,264.8) -- cycle ;
	\draw  [draw opacity=0][fill={rgb, 255:red, 0; green, 0; blue, 0 }  ,fill opacity=1 ] (539.6,264.8) .. controls (539.6,261.38) and (542.38,258.6) .. (545.8,258.6) .. controls (549.22,258.6) and (552,261.38) .. (552,264.8) .. controls (552,268.22) and (549.22,271) .. (545.8,271) .. controls (542.38,271) and (539.6,268.22) .. (539.6,264.8) -- cycle ;
	\draw [color={rgb, 255:red, 255; green, 0; blue, 0 }  ,draw opacity=1 ]   (455.6,264.8) .. controls (362.07,244.47) and (390.31,62.2) .. (494.03,107.11) ;
	\draw [shift={(495.6,107.8)}, rotate = 204.31] [fill={rgb, 255:red, 255; green, 0; blue, 0 }  ,fill opacity=1 ][line width=0.08]  [draw opacity=0] (10.72,-5.15) -- (0,0) -- (10.72,5.15) -- (7.12,0) -- cycle    ;
	\draw [color={rgb, 255:red, 255; green, 0; blue, 0 }  ,draw opacity=1 ]   (552,264.8) .. controls (648.12,241.49) and (615.93,62.17) .. (509.61,107.11) ;
	\draw [shift={(508,107.8)}, rotate = 336.21] [fill={rgb, 255:red, 255; green, 0; blue, 0 }  ,fill opacity=1 ][line width=0.08]  [draw opacity=0] (10.72,-5.15) -- (0,0) -- (10.72,5.15) -- (7.12,0) -- cycle    ;
	
	\draw (120.13,187.97) node   [align=left] {\begin{minipage}[lt]{27.83pt}\setlength\topsep{0pt}
			*? 
	\end{minipage}};
	\draw (231,184.23) node   [align=left] {\begin{minipage}[lt]{27.83pt}\setlength\topsep{0pt}
			*? 
	\end{minipage}};
	\draw (41.27,326.17) node   [align=left] {\begin{minipage}[lt]{29.83pt}\setlength\topsep{0pt}
			Head
	\end{minipage}};
	\draw (45.6,292.17) node   [align=left] {\begin{minipage}[lt]{15.78pt}\setlength\topsep{0pt}
			*1
	\end{minipage}};
	\draw (142.6,325.5) node   [align=left] {\begin{minipage}[lt]{29.83pt}\setlength\topsep{0pt}
			Foot
	\end{minipage}};
	\draw (137.27,289.83) node   [align=left] {\begin{minipage}[lt]{15.78pt}\setlength\topsep{0pt}
			*4
	\end{minipage}};
	\draw (204.27,325.17) node   [align=left] {\begin{minipage}[lt]{29.83pt}\setlength\topsep{0pt}
			Head
	\end{minipage}};
	\draw (207.6,291.17) node   [align=left] {\begin{minipage}[lt]{15.78pt}\setlength\topsep{0pt}
			*1
	\end{minipage}};
	\draw (292.6,324.5) node   [align=left] {\begin{minipage}[lt]{29.83pt}\setlength\topsep{0pt}
			Foot
	\end{minipage}};
	\draw (287.27,288.83) node   [align=left] {\begin{minipage}[lt]{15.78pt}\setlength\topsep{0pt}
			*2
	\end{minipage}};
	\draw (122.27,55.27) node   [align=left] {\begin{minipage}[lt]{29.83pt}\setlength\topsep{0pt}
			Head
	\end{minipage}};
	\draw (208.27,92.93) node   [align=left] {\begin{minipage}[lt]{15.78pt}\setlength\topsep{0pt}
			*94
	\end{minipage}};
	\draw (210.6,54.6) node   [align=left] {\begin{minipage}[lt]{29.83pt}\setlength\topsep{0pt}
			Foot
	\end{minipage}};
	\draw (121.27,91.6) node   [align=left] {\begin{minipage}[lt]{15.78pt}\setlength\topsep{0pt}
			*35
	\end{minipage}};
	\draw (182.13,193.97) node   [align=left] {\begin{minipage}[lt]{27.83pt}\setlength\topsep{0pt}
			+
	\end{minipage}};
	\draw (105.2,291.3) node   [align=left] {\begin{minipage}[lt]{27.83pt}\setlength\topsep{0pt}
			+
	\end{minipage}};
	\draw (261.13,289.97) node   [align=left] {\begin{minipage}[lt]{27.83pt}\setlength\topsep{0pt}
			+
	\end{minipage}};
	\draw (181.63,93.97) node   [align=left] {\begin{minipage}[lt]{27.83pt}\setlength\topsep{0pt}
			+
	\end{minipage}};
	\draw (328.53,209.6) node   [align=left] {\begin{minipage}[lt]{55.13pt}\setlength\topsep{0pt}
			Abstraction
	\end{minipage}};
	\draw (455,136.4) node   [align=left] {\begin{minipage}[lt]{27.83pt}\setlength\topsep{0pt}
			\textcolor[rgb]{0,0,1}{*a }
	\end{minipage}};
	\draw (563,135) node   [align=left] {\begin{minipage}[lt]{27.83pt}\setlength\topsep{0pt}
			\textcolor[rgb]{0,0,1}{*b }
	\end{minipage}};
	\draw (431.27,219.27) node   [align=left] {\begin{minipage}[lt]{15.78pt}\setlength\topsep{0pt}
			\textcolor[rgb]{0,0,1}{*1}
	\end{minipage}};
	\draw (584.93,208.6) node   [align=left] {\begin{minipage}[lt]{15.78pt}\setlength\topsep{0pt}
			\textcolor[rgb]{0,0,1}{*2}
	\end{minipage}};
	\draw (552.27,84.7) node   [align=left] {\begin{minipage}[lt]{15.78pt}\setlength\topsep{0pt}
			\textcolor[rgb]{1,0,0}{*94}
	\end{minipage}};
	\draw (450.27,84.37) node   [align=left] {\begin{minipage}[lt]{15.78pt}\setlength\topsep{0pt}
			\textcolor[rgb]{1,0,0}{*35}
	\end{minipage}};
	\draw (518.13,144.73) node   [align=left] {\begin{minipage}[lt]{27.83pt}\setlength\topsep{0pt}
			\textcolor[rgb]{0,0,1}{+}
	\end{minipage}};
	\draw (565.47,207.07) node   [align=left] {\begin{minipage}[lt]{27.83pt}\setlength\topsep{0pt}
			\textcolor[rgb]{0,0,1}{+}
	\end{minipage}};
	\draw (487.13,202.57) node   [align=left] {\begin{minipage}[lt]{21.03pt}\setlength\topsep{0pt}
			\textcolor[rgb]{0,0,1}{*4}
	\end{minipage}};
	\draw (460.17,209.93) node   [align=left] {\begin{minipage}[lt]{12.83pt}\setlength\topsep{0pt}
			\textcolor[rgb]{0,0,1}{+}
	\end{minipage}};
	\draw (523.27,204.27) node   [align=left] {\begin{minipage}[lt]{15.78pt}\setlength\topsep{0pt}
			\textcolor[rgb]{0,0,1}{*1}
	\end{minipage}};
	\draw (517.13,84.73) node  [color={rgb, 255:red, 255; green, 0; blue, 0 }  ,opacity=1 ] [align=left] {\begin{minipage}[lt]{27.83pt}\setlength\topsep{0pt}
			+
	\end{minipage}};
	\draw (464.3,281.6) node   [align=left] {\begin{minipage}[lt]{11.29pt}\setlength\topsep{0pt}
			X
	\end{minipage}};
	\draw (549.3,282.6) node   [align=left] {\begin{minipage}[lt]{11.29pt}\setlength\topsep{0pt}
			Y
	\end{minipage}};
	\draw (504.7,123.97) node   [align=left] {\begin{minipage}[lt]{11.29pt}\setlength\topsep{0pt}
			Z
	\end{minipage}};

\end{tikzpicture}

%% file: 10.tex
\tikzset{every picture/.style={line width=0.75pt}} 

\begin{tikzpicture}[x=0.75pt,y=0.75pt,yscale=-1,xscale=1]
	
	\draw  [draw opacity=0][fill={rgb, 255:red, 255; green, 128; blue, 0 }  ,fill opacity=0.1 ] (110,87.46) .. controls (110,74.93) and (120.15,64.78) .. (132.68,64.78) -- (319.66,64.78) .. controls (332.18,64.78) and (342.33,74.93) .. (342.33,87.46) -- (342.33,155.49) .. controls (342.33,168.01) and (332.18,178.17) .. (319.66,178.17) -- (132.68,178.17) .. controls (120.15,178.17) and (110,168.01) .. (110,155.49) -- cycle ;
	\draw  [draw opacity=0][fill={rgb, 255:red, 0; green, 0; blue, 0 }  ,fill opacity=1 ] (152.23,109.48) .. controls (152.23,106.69) and (154.5,104.42) .. (157.3,104.42) .. controls (160.1,104.42) and (162.37,106.69) .. (162.37,109.48) .. controls (162.37,112.28) and (160.1,114.55) .. (157.3,114.55) .. controls (154.5,114.55) and (152.23,112.28) .. (152.23,109.48) -- cycle ;
	\draw  [draw opacity=0][fill={rgb, 255:red, 0; green, 0; blue, 0 }  ,fill opacity=1 ] (177.5,124.43) .. controls (177.5,121.64) and (179.77,119.37) .. (182.57,119.37) .. controls (185.36,119.37) and (187.63,121.64) .. (187.63,124.43) .. controls (187.63,127.23) and (185.36,129.5) .. (182.57,129.5) .. controls (179.77,129.5) and (177.5,127.23) .. (177.5,124.43) -- cycle ;
	\draw  [draw opacity=0][fill={rgb, 255:red, 0; green, 0; blue, 0 }  ,fill opacity=1 ] (160,141.43) .. controls (160,138.64) and (162.27,136.37) .. (165.07,136.37) .. controls (167.86,136.37) and (170.13,138.64) .. (170.13,141.43) .. controls (170.13,144.23) and (167.86,146.5) .. (165.07,146.5) .. controls (162.27,146.5) and (160,144.23) .. (160,141.43) -- cycle ;
	\draw  [draw opacity=0][fill={rgb, 255:red, 0; green, 0; blue, 0 }  ,fill opacity=1 ] (134.5,129.43) .. controls (134.5,126.64) and (136.77,124.37) .. (139.57,124.37) .. controls (142.36,124.37) and (144.63,126.64) .. (144.63,129.43) .. controls (144.63,132.23) and (142.36,134.5) .. (139.57,134.5) .. controls (136.77,134.5) and (134.5,132.23) .. (134.5,129.43) -- cycle ;
	\draw [color={rgb, 255:red, 100; green, 100; blue, 100 }  ,draw opacity=1 ]   (161.8,111.05) -- (174.78,119.42) ;
	\draw [shift={(177.3,121.05)}, rotate = 212.83] [fill={rgb, 255:red, 100; green, 100; blue, 100 }  ,fill opacity=1 ][line width=0.08]  [draw opacity=0] (5.36,-2.57) -- (0,0) -- (5.36,2.57) -- (3.56,0) -- cycle    ;
	\draw [color={rgb, 255:red, 100; green, 100; blue, 100 }  ,draw opacity=1 ]   (157.94,117.48) -- (161.16,132.12) ;
	\draw [shift={(161.8,135.05)}, rotate = 257.62] [fill={rgb, 255:red, 100; green, 100; blue, 100 }  ,fill opacity=1 ][line width=0.08]  [draw opacity=0] (5.36,-2.57) -- (0,0) -- (5.36,2.57) -- (3.56,0) -- cycle    ;
	\draw [shift={(157.3,114.55)}, rotate = 77.62] [fill={rgb, 255:red, 100; green, 100; blue, 100 }  ,fill opacity=1 ][line width=0.08]  [draw opacity=0] (5.36,-2.57) -- (0,0) -- (5.36,2.57) -- (3.56,0) -- cycle    ;
	\draw [color={rgb, 255:red, 100; green, 100; blue, 100 }  ,draw opacity=1 ]   (176.51,131.49) -- (168.8,138.05) ;
	\draw [shift={(178.8,129.55)}, rotate = 139.64] [fill={rgb, 255:red, 100; green, 100; blue, 100 }  ,fill opacity=1 ][line width=0.08]  [draw opacity=0] (5.36,-2.57) -- (0,0) -- (5.36,2.57) -- (3.56,0) -- cycle    ;
	\draw [color={rgb, 255:red, 100; green, 100; blue, 100 }  ,draw opacity=1 ]   (151.8,113.05) -- (145.94,122.04) ;
	\draw [shift={(144.3,124.55)}, rotate = 303.11] [fill={rgb, 255:red, 100; green, 100; blue, 100 }  ,fill opacity=1 ][line width=0.08]  [draw opacity=0] (5.36,-2.57) -- (0,0) -- (5.36,2.57) -- (3.56,0) -- cycle    ;
	\draw  [draw opacity=0][fill={rgb, 255:red, 0; green, 0; blue, 0 }  ,fill opacity=1 ] (202.83,127.1) .. controls (202.83,124.3) and (205.1,122.03) .. (207.9,122.03) .. controls (210.7,122.03) and (212.97,124.3) .. (212.97,127.1) .. controls (212.97,129.9) and (210.7,132.17) .. (207.9,132.17) .. controls (205.1,132.17) and (202.83,129.9) .. (202.83,127.1) -- cycle ;
	\draw [color={rgb, 255:red, 100; green, 100; blue, 100 }  ,draw opacity=1 ]   (187.63,124.43) -- (199.88,126.58) ;
	\draw [shift={(202.83,127.1)}, rotate = 189.95] [fill={rgb, 255:red, 100; green, 100; blue, 100 }  ,fill opacity=1 ][line width=0.08]  [draw opacity=0] (5.36,-2.57) -- (0,0) -- (5.36,2.57) -- (3.56,0) -- cycle    ;
	\draw  [draw opacity=0][fill={rgb, 255:red, 0; green, 0; blue, 0 }  ,fill opacity=1 ] (205.5,151.1) .. controls (205.5,148.3) and (207.77,146.03) .. (210.57,146.03) .. controls (213.36,146.03) and (215.63,148.3) .. (215.63,151.1) .. controls (215.63,153.9) and (213.36,156.17) .. (210.57,156.17) .. controls (207.77,156.17) and (205.5,153.9) .. (205.5,151.1) -- cycle ;
	\draw [color={rgb, 255:red, 100; green, 100; blue, 100 }  ,draw opacity=1 ]   (208.73,130.3) -- (210.22,143.05) ;
	\draw [shift={(210.57,146.03)}, rotate = 263.35] [fill={rgb, 255:red, 100; green, 100; blue, 100 }  ,fill opacity=1 ][line width=0.08]  [draw opacity=0] (5.36,-2.57) -- (0,0) -- (5.36,2.57) -- (3.56,0) -- cycle    ;
	\draw  [draw opacity=0][fill={rgb, 255:red, 0; green, 0; blue, 0 }  ,fill opacity=1 ] (232.83,102.43) .. controls (232.83,99.64) and (235.1,97.37) .. (237.9,97.37) .. controls (240.7,97.37) and (242.97,99.64) .. (242.97,102.43) .. controls (242.97,105.23) and (240.7,107.5) .. (237.9,107.5) .. controls (235.1,107.5) and (232.83,105.23) .. (232.83,102.43) -- cycle ;
	\draw  [draw opacity=0][fill={rgb, 255:red, 0; green, 0; blue, 0 }  ,fill opacity=1 ] (262.17,119.1) .. controls (262.17,116.3) and (264.44,114.03) .. (267.23,114.03) .. controls (270.03,114.03) and (272.3,116.3) .. (272.3,119.1) .. controls (272.3,121.9) and (270.03,124.17) .. (267.23,124.17) .. controls (264.44,124.17) and (262.17,121.9) .. (262.17,119.1) -- cycle ;
	\draw  [draw opacity=0][fill={rgb, 255:red, 0; green, 0; blue, 0 }  ,fill opacity=1 ] (263.5,149.77) .. controls (263.5,146.97) and (265.77,144.7) .. (268.57,144.7) .. controls (271.36,144.7) and (273.63,146.97) .. (273.63,149.77) .. controls (273.63,152.56) and (271.36,154.83) .. (268.57,154.83) .. controls (265.77,154.83) and (263.5,152.56) .. (263.5,149.77) -- cycle ;
	\draw  [draw opacity=0][fill={rgb, 255:red, 0; green, 0; blue, 0 }  ,fill opacity=1 ] (306.83,108.43) .. controls (306.83,105.64) and (309.1,103.37) .. (311.9,103.37) .. controls (314.7,103.37) and (316.97,105.64) .. (316.97,108.43) .. controls (316.97,111.23) and (314.7,113.5) .. (311.9,113.5) .. controls (309.1,113.5) and (306.83,111.23) .. (306.83,108.43) -- cycle ;
	\draw [color={rgb, 255:red, 100; green, 100; blue, 100 }  ,draw opacity=1 ]   (188.97,121.1) -- (227.52,106.72) ;
	\draw [shift={(230.33,105.67)}, rotate = 159.54] [fill={rgb, 255:red, 100; green, 100; blue, 100 }  ,fill opacity=1 ][line width=0.08]  [draw opacity=0] (5.36,-2.57) -- (0,0) -- (5.36,2.57) -- (3.56,0) -- cycle    ;
	\draw [color={rgb, 255:red, 100; green, 100; blue, 100 }  ,draw opacity=1 ]   (245.95,102.71) -- (303.85,108.15) ;
	\draw [shift={(306.83,108.43)}, rotate = 185.37] [fill={rgb, 255:red, 100; green, 100; blue, 100 }  ,fill opacity=1 ][line width=0.08]  [draw opacity=0] (5.36,-2.57) -- (0,0) -- (5.36,2.57) -- (3.56,0) -- cycle    ;
	\draw [shift={(242.97,102.43)}, rotate = 5.37] [fill={rgb, 255:red, 100; green, 100; blue, 100 }  ,fill opacity=1 ][line width=0.08]  [draw opacity=0] (5.36,-2.57) -- (0,0) -- (5.36,2.57) -- (3.56,0) -- cycle    ;
	\draw [color={rgb, 255:red, 100; green, 100; blue, 100 }  ,draw opacity=1 ]   (272.3,119.1) -- (302.06,112.94) ;
	\draw [shift={(305,112.33)}, rotate = 168.31] [fill={rgb, 255:red, 100; green, 100; blue, 100 }  ,fill opacity=1 ][line width=0.08]  [draw opacity=0] (5.36,-2.57) -- (0,0) -- (5.36,2.57) -- (3.56,0) -- cycle    ;
	\draw [color={rgb, 255:red, 100; green, 100; blue, 100 }  ,draw opacity=1 ]   (241.9,107.5) -- (259.56,117.61) ;
	\draw [shift={(262.17,119.1)}, rotate = 209.79] [fill={rgb, 255:red, 100; green, 100; blue, 100 }  ,fill opacity=1 ][line width=0.08]  [draw opacity=0] (5.36,-2.57) -- (0,0) -- (5.36,2.57) -- (3.56,0) -- cycle    ;
	\draw [color={rgb, 255:red, 100; green, 100; blue, 100 }  ,draw opacity=1 ]   (267.43,127.16) -- (268.37,141.71) ;
	\draw [shift={(268.57,144.7)}, rotate = 266.28] [fill={rgb, 255:red, 100; green, 100; blue, 100 }  ,fill opacity=1 ][line width=0.08]  [draw opacity=0] (5.36,-2.57) -- (0,0) -- (5.36,2.57) -- (3.56,0) -- cycle    ;
	\draw [shift={(267.23,124.17)}, rotate = 86.28] [fill={rgb, 255:red, 100; green, 100; blue, 100 }  ,fill opacity=1 ][line width=0.08]  [draw opacity=0] (5.36,-2.57) -- (0,0) -- (5.36,2.57) -- (3.56,0) -- cycle    ;
	\draw [color={rgb, 255:red, 100; green, 100; blue, 100 }  ,draw opacity=1 ]   (215.63,147.1) -- (259.67,124.38) ;
	\draw [shift={(262.33,123)}, rotate = 152.7] [fill={rgb, 255:red, 100; green, 100; blue, 100 }  ,fill opacity=1 ][line width=0.08]  [draw opacity=0] (5.36,-2.57) -- (0,0) -- (5.36,2.57) -- (3.56,0) -- cycle    ;
	\draw [color={rgb, 255:red, 100; green, 100; blue, 100 }  ,draw opacity=1 ]   (214.09,121.38) -- (231.21,109.39) ;
	\draw [shift={(233.67,107.67)}, rotate = 144.99] [fill={rgb, 255:red, 100; green, 100; blue, 100 }  ,fill opacity=1 ][line width=0.08]  [draw opacity=0] (5.36,-2.57) -- (0,0) -- (5.36,2.57) -- (3.56,0) -- cycle    ;
	\draw [shift={(211.63,123.1)}, rotate = 324.99] [fill={rgb, 255:red, 100; green, 100; blue, 100 }  ,fill opacity=1 ][line width=0.08]  [draw opacity=0] (5.36,-2.57) -- (0,0) -- (5.36,2.57) -- (3.56,0) -- cycle    ;
	\draw  [draw opacity=0][fill={rgb, 255:red, 0; green, 0; blue, 0 }  ,fill opacity=1 ] (160.83,215.1) .. controls (160.83,212.3) and (163.1,210.03) .. (165.9,210.03) .. controls (168.7,210.03) and (170.97,212.3) .. (170.97,215.1) .. controls (170.97,217.9) and (168.7,220.17) .. (165.9,220.17) .. controls (163.1,220.17) and (160.83,217.9) .. (160.83,215.1) -- cycle ;
	\draw  [draw opacity=0][fill={rgb, 255:red, 0; green, 0; blue, 0 }  ,fill opacity=1 ] (176.83,233.1) .. controls (176.83,230.3) and (179.1,228.03) .. (181.9,228.03) .. controls (184.7,228.03) and (186.97,230.3) .. (186.97,233.1) .. controls (186.97,235.9) and (184.7,238.17) .. (181.9,238.17) .. controls (179.1,238.17) and (176.83,235.9) .. (176.83,233.1) -- cycle ;
	\draw  [draw opacity=0][fill={rgb, 255:red, 0; green, 0; blue, 0 }  ,fill opacity=1 ] (192.83,209.77) .. controls (192.83,206.97) and (195.1,204.7) .. (197.9,204.7) .. controls (200.7,204.7) and (202.97,206.97) .. (202.97,209.77) .. controls (202.97,212.56) and (200.7,214.83) .. (197.9,214.83) .. controls (195.1,214.83) and (192.83,212.56) .. (192.83,209.77) -- cycle ;
	\draw  [draw opacity=0][fill={rgb, 255:red, 0; green, 0; blue, 0 }  ,fill opacity=1 ] (174.17,195.77) .. controls (174.17,192.97) and (176.44,190.7) .. (179.23,190.7) .. controls (182.03,190.7) and (184.3,192.97) .. (184.3,195.77) .. controls (184.3,198.56) and (182.03,200.83) .. (179.23,200.83) .. controls (176.44,200.83) and (174.17,198.56) .. (174.17,195.77) -- cycle ;
	\draw [color={rgb, 255:red, 0; green, 255; blue, 0 }  ,draw opacity=1 ] [dash pattern={on 0.84pt off 2.51pt}]  (166.51,207.1) -- (182.57,129.5) ;
	\draw [shift={(165.9,210.03)}, rotate = 281.69] [fill={rgb, 255:red, 0; green, 255; blue, 0 }  ,fill opacity=1 ][line width=0.08]  [draw opacity=0] (5.36,-2.57) -- (0,0) -- (5.36,2.57) -- (3.56,0) -- cycle    ;
	\draw [color={rgb, 255:red, 0; green, 255; blue, 0 }  ,draw opacity=1 ] [dash pattern={on 0.84pt off 2.51pt}]  (179.4,187.7) -- (182.57,129.5) ;
	\draw [shift={(179.23,190.7)}, rotate = 273.12] [fill={rgb, 255:red, 0; green, 255; blue, 0 }  ,fill opacity=1 ][line width=0.08]  [draw opacity=0] (5.36,-2.57) -- (0,0) -- (5.36,2.57) -- (3.56,0) -- cycle    ;
	\draw [color={rgb, 255:red, 0; green, 255; blue, 0 }  ,draw opacity=1 ] [dash pattern={on 0.84pt off 2.51pt}]  (197.3,201.76) -- (182.57,129.5) ;
	\draw [shift={(197.9,204.7)}, rotate = 258.48] [fill={rgb, 255:red, 0; green, 255; blue, 0 }  ,fill opacity=1 ][line width=0.08]  [draw opacity=0] (5.36,-2.57) -- (0,0) -- (5.36,2.57) -- (3.56,0) -- cycle    ;
	\draw [color={rgb, 255:red, 0; green, 255; blue, 0 }  ,draw opacity=1 ][line width=0.75]  [dash pattern={on 0.84pt off 2.51pt}]  (181.95,225.03) -- (183.44,133.67) ;
	\draw [shift={(181.9,228.03)}, rotate = 270.94] [fill={rgb, 255:red, 0; green, 255; blue, 0 }  ,fill opacity=1 ][line width=0.08]  [draw opacity=0] (5.36,-2.57) -- (0,0) -- (5.36,2.57) -- (3.56,0) -- cycle    ;
	\draw [color={rgb, 255:red, 0; green, 255; blue, 255 }  ,draw opacity=0.3 ]   (166.38,149.43) -- (174.8,187.4) ;
	\draw [shift={(175.44,190.33)}, rotate = 257.51] [fill={rgb, 255:red, 0; green, 255; blue, 255 }  ,fill opacity=0.3 ][line width=0.08]  [draw opacity=0] (5.36,-2.57) -- (0,0) -- (5.36,2.57) -- (3.56,0) -- cycle    ;
	\draw [shift={(165.73,146.5)}, rotate = 77.51] [fill={rgb, 255:red, 0; green, 255; blue, 255 }  ,fill opacity=0.3 ][line width=0.08]  [draw opacity=0] (5.36,-2.57) -- (0,0) -- (5.36,2.57) -- (3.56,0) -- cycle    ;
	\draw [color={rgb, 255:red, 0; green, 255; blue, 255 }  ,draw opacity=0.3 ]   (204.3,205.77) -- (263.16,155.61) ;
	\draw [shift={(265.44,153.67)}, rotate = 139.57] [fill={rgb, 255:red, 0; green, 255; blue, 255 }  ,fill opacity=0.3 ][line width=0.08]  [draw opacity=0] (5.36,-2.57) -- (0,0) -- (5.36,2.57) -- (3.56,0) -- cycle    ;
	\draw [color={rgb, 255:red, 100; green, 100; blue, 100 }  ,draw opacity=1 ]   (184.33,197.72) -- (191.78,203.77) ;
	\draw [shift={(194.11,205.67)}, rotate = 219.09] [fill={rgb, 255:red, 100; green, 100; blue, 100 }  ,fill opacity=1 ][line width=0.08]  [draw opacity=0] (5.36,-2.57) -- (0,0) -- (5.36,2.57) -- (3.56,0) -- cycle    ;
	\draw [color={rgb, 255:red, 100; green, 100; blue, 100 }  ,draw opacity=1 ]   (173.26,212.95) -- (189.87,210.25) ;
	\draw [shift={(192.83,209.77)}, rotate = 170.76] [fill={rgb, 255:red, 100; green, 100; blue, 100 }  ,fill opacity=1 ][line width=0.08]  [draw opacity=0] (5.36,-2.57) -- (0,0) -- (5.36,2.57) -- (3.56,0) -- cycle    ;
	\draw [shift={(170.3,213.43)}, rotate = 350.76] [fill={rgb, 255:red, 100; green, 100; blue, 100 }  ,fill opacity=1 ][line width=0.08]  [draw opacity=0] (5.36,-2.57) -- (0,0) -- (5.36,2.57) -- (3.56,0) -- cycle    ;
	\draw [color={rgb, 255:red, 100; green, 100; blue, 100 }  ,draw opacity=1 ]   (168.3,219.77) -- (175.03,227.46) ;
	\draw [shift={(177,229.72)}, rotate = 228.85] [fill={rgb, 255:red, 100; green, 100; blue, 100 }  ,fill opacity=1 ][line width=0.08]  [draw opacity=0] (5.36,-2.57) -- (0,0) -- (5.36,2.57) -- (3.56,0) -- cycle    ;
	\draw [color={rgb, 255:red, 100; green, 100; blue, 100 }  ,draw opacity=1 ]   (170.69,208.31) -- (174.28,203.18) ;
	\draw [shift={(176,200.72)}, rotate = 125] [fill={rgb, 255:red, 100; green, 100; blue, 100 }  ,fill opacity=1 ][line width=0.08]  [draw opacity=0] (5.36,-2.57) -- (0,0) -- (5.36,2.57) -- (3.56,0) -- cycle    ;
	\draw [shift={(168.97,210.77)}, rotate = 305] [fill={rgb, 255:red, 100; green, 100; blue, 100 }  ,fill opacity=1 ][line width=0.08]  [draw opacity=0] (5.36,-2.57) -- (0,0) -- (5.36,2.57) -- (3.56,0) -- cycle    ;
	\draw  [draw opacity=0][fill={rgb, 255:red, 0; green, 0; blue, 0 }  ,fill opacity=1 ] (152.17,283.1) .. controls (152.17,280.3) and (154.44,278.03) .. (157.23,278.03) .. controls (160.03,278.03) and (162.3,280.3) .. (162.3,283.1) .. controls (162.3,285.9) and (160.03,288.17) .. (157.23,288.17) .. controls (154.44,288.17) and (152.17,285.9) .. (152.17,283.1) -- cycle ;
	\draw  [draw opacity=0][fill={rgb, 255:red, 0; green, 0; blue, 0 }  ,fill opacity=1 ] (168.17,269.77) .. controls (168.17,266.97) and (170.44,264.7) .. (173.23,264.7) .. controls (176.03,264.7) and (178.3,266.97) .. (178.3,269.77) .. controls (178.3,272.56) and (176.03,274.83) .. (173.23,274.83) .. controls (170.44,274.83) and (168.17,272.56) .. (168.17,269.77) -- cycle ;
	\draw  [draw opacity=0][fill={rgb, 255:red, 0; green, 0; blue, 0 }  ,fill opacity=1 ] (185.5,285.77) .. controls (185.5,282.97) and (187.77,280.7) .. (190.57,280.7) .. controls (193.36,280.7) and (195.63,282.97) .. (195.63,285.77) .. controls (195.63,288.56) and (193.36,290.83) .. (190.57,290.83) .. controls (187.77,290.83) and (185.5,288.56) .. (185.5,285.77) -- cycle ;
	\draw  [draw opacity=0][fill={rgb, 255:red, 0; green, 0; blue, 0 }  ,fill opacity=1 ] (199.5,269.1) .. controls (199.5,266.3) and (201.77,264.03) .. (204.57,264.03) .. controls (207.36,264.03) and (209.63,266.3) .. (209.63,269.1) .. controls (209.63,271.9) and (207.36,274.17) .. (204.57,274.17) .. controls (201.77,274.17) and (199.5,271.9) .. (199.5,269.1) -- cycle ;
	\draw [color={rgb, 255:red, 255; green, 0; blue, 0 }  ,draw opacity=1 ] [dash pattern={on 0.84pt off 2.51pt}]  (158.81,275.48) -- (181.9,238.17) ;
	\draw [shift={(157.23,278.03)}, rotate = 301.75] [fill={rgb, 255:red, 255; green, 0; blue, 0 }  ,fill opacity=1 ][line width=0.08]  [draw opacity=0] (5.36,-2.57) -- (0,0) -- (5.36,2.57) -- (3.56,0) -- cycle    ;
	\draw [color={rgb, 255:red, 255; green, 0; blue, 0 }  ,draw opacity=1 ] [dash pattern={on 0.84pt off 2.51pt}]  (176.99,260.07) -- (181.9,238.17) ;
	\draw [shift={(176.33,263)}, rotate = 282.63] [fill={rgb, 255:red, 255; green, 0; blue, 0 }  ,fill opacity=1 ][line width=0.08]  [draw opacity=0] (5.36,-2.57) -- (0,0) -- (5.36,2.57) -- (3.56,0) -- cycle    ;
	\draw [color={rgb, 255:red, 255; green, 0; blue, 0 }  ,draw opacity=1 ] [dash pattern={on 0.84pt off 2.51pt}]  (189.97,277.76) -- (181.9,238.17) ;
	\draw [shift={(190.57,280.7)}, rotate = 258.48] [fill={rgb, 255:red, 255; green, 0; blue, 0 }  ,fill opacity=1 ][line width=0.08]  [draw opacity=0] (5.36,-2.57) -- (0,0) -- (5.36,2.57) -- (3.56,0) -- cycle    ;
	\draw [color={rgb, 255:red, 255; green, 0; blue, 0 }  ,draw opacity=1 ][line width=0.75]  [dash pattern={on 0.84pt off 2.51pt}]  (199.47,260.42) -- (186.33,238.33) ;
	\draw [shift={(201,263)}, rotate = 239.26] [fill={rgb, 255:red, 255; green, 0; blue, 0 }  ,fill opacity=1 ][line width=0.08]  [draw opacity=0] (5.36,-2.57) -- (0,0) -- (5.36,2.57) -- (3.56,0) -- cycle    ;
	\draw [color={rgb, 255:red, 100; green, 100; blue, 100 }  ,draw opacity=1 ]   (181.3,269.67) -- (196.5,269.19) ;
	\draw [shift={(199.5,269.1)}, rotate = 178.2] [fill={rgb, 255:red, 100; green, 100; blue, 100 }  ,fill opacity=1 ][line width=0.08]  [draw opacity=0] (5.36,-2.57) -- (0,0) -- (5.36,2.57) -- (3.56,0) -- cycle    ;
	\draw [shift={(178.3,269.77)}, rotate = 358.2] [fill={rgb, 255:red, 100; green, 100; blue, 100 }  ,fill opacity=1 ][line width=0.08]  [draw opacity=0] (5.36,-2.57) -- (0,0) -- (5.36,2.57) -- (3.56,0) -- cycle    ;
	\draw [color={rgb, 255:red, 100; green, 100; blue, 100 }  ,draw opacity=1 ]   (160.33,280.33) -- (167.23,275.41) ;
	\draw [shift={(169.67,273.67)}, rotate = 144.46] [fill={rgb, 255:red, 100; green, 100; blue, 100 }  ,fill opacity=1 ][line width=0.08]  [draw opacity=0] (5.36,-2.57) -- (0,0) -- (5.36,2.57) -- (3.56,0) -- cycle    ;
	\draw [color={rgb, 255:red, 100; green, 100; blue, 100 }  ,draw opacity=1 ]   (165.28,283.44) -- (185.5,285.77) ;
	\draw [shift={(162.3,283.1)}, rotate = 6.56] [fill={rgb, 255:red, 100; green, 100; blue, 100 }  ,fill opacity=1 ][line width=0.08]  [draw opacity=0] (5.36,-2.57) -- (0,0) -- (5.36,2.57) -- (3.56,0) -- cycle    ;
	\draw [color={rgb, 255:red, 0; green, 255; blue, 255 }  ,draw opacity=0.3 ]   (206.42,261.28) -- (263.67,128.11) ;
	\draw [shift={(205.23,264.03)}, rotate = 293.26] [fill={rgb, 255:red, 0; green, 255; blue, 255 }  ,fill opacity=0.3 ][line width=0.08]  [draw opacity=0] (5.36,-2.57) -- (0,0) -- (5.36,2.57) -- (3.56,0) -- cycle    ;
	\draw  [draw opacity=0][fill={rgb, 255:red, 0; green, 0; blue, 0 }  ,fill opacity=1 ] (196.83,311.77) .. controls (196.83,308.97) and (199.1,306.7) .. (201.9,306.7) .. controls (204.7,306.7) and (206.97,308.97) .. (206.97,311.77) .. controls (206.97,314.56) and (204.7,316.83) .. (201.9,316.83) .. controls (199.1,316.83) and (196.83,314.56) .. (196.83,311.77) -- cycle ;
	\draw  [draw opacity=0][fill={rgb, 255:red, 0; green, 0; blue, 0 }  ,fill opacity=1 ] (218.83,319.77) .. controls (218.83,316.97) and (221.1,314.7) .. (223.9,314.7) .. controls (226.7,314.7) and (228.97,316.97) .. (228.97,319.77) .. controls (228.97,322.56) and (226.7,324.83) .. (223.9,324.83) .. controls (221.1,324.83) and (218.83,322.56) .. (218.83,319.77) -- cycle ;
	\draw  [draw opacity=0][fill={rgb, 255:red, 0; green, 0; blue, 0 }  ,fill opacity=1 ] (228.83,299.1) .. controls (228.83,296.3) and (231.1,294.03) .. (233.9,294.03) .. controls (236.7,294.03) and (238.97,296.3) .. (238.97,299.1) .. controls (238.97,301.9) and (236.7,304.17) .. (233.9,304.17) .. controls (231.1,304.17) and (228.83,301.9) .. (228.83,299.1) -- cycle ;
	\draw [color={rgb, 255:red, 0; green, 0; blue, 255 }  ,draw opacity=1 ][line width=0.75]  [dash pattern={on 0.84pt off 2.51pt}]  (226.97,292.57) -- (208.33,272.33) ;
	\draw [shift={(229,294.78)}, rotate = 227.36] [fill={rgb, 255:red, 0; green, 0; blue, 255 }  ,fill opacity=1 ][line width=0.08]  [draw opacity=0] (5.36,-2.57) -- (0,0) -- (5.36,2.57) -- (3.56,0) -- cycle    ;
	\draw [color={rgb, 255:red, 0; green, 0; blue, 255 }  ,draw opacity=1 ] [dash pattern={on 0.84pt off 2.51pt}]  (202.15,303.71) -- (204.57,274.17) ;
	\draw [shift={(201.9,306.7)}, rotate = 274.69] [fill={rgb, 255:red, 0; green, 0; blue, 255 }  ,fill opacity=1 ][line width=0.08]  [draw opacity=0] (5.36,-2.57) -- (0,0) -- (5.36,2.57) -- (3.56,0) -- cycle    ;
	\draw [color={rgb, 255:red, 0; green, 0; blue, 255 }  ,draw opacity=1 ] [dash pattern={on 0.84pt off 2.51pt}]  (222.73,311.94) -- (207.67,276.33) ;
	\draw [shift={(223.9,314.7)}, rotate = 247.07] [fill={rgb, 255:red, 0; green, 0; blue, 255 }  ,fill opacity=1 ][line width=0.08]  [draw opacity=0] (5.36,-2.57) -- (0,0) -- (5.36,2.57) -- (3.56,0) -- cycle    ;
	\draw [color={rgb, 255:red, 100; green, 100; blue, 100 }  ,draw opacity=1 ]   (206.3,308.43) -- (226.06,300.25) ;
	\draw [shift={(228.83,299.1)}, rotate = 157.5] [fill={rgb, 255:red, 100; green, 100; blue, 100 }  ,fill opacity=1 ][line width=0.08]  [draw opacity=0] (5.36,-2.57) -- (0,0) -- (5.36,2.57) -- (3.56,0) -- cycle    ;
	\draw [color={rgb, 255:red, 100; green, 100; blue, 100 }  ,draw opacity=1 ]   (209.11,316.15) -- (216.02,318.72) ;
	\draw [shift={(218.83,319.77)}, rotate = 200.42] [fill={rgb, 255:red, 100; green, 100; blue, 100 }  ,fill opacity=1 ][line width=0.08]  [draw opacity=0] (5.36,-2.57) -- (0,0) -- (5.36,2.57) -- (3.56,0) -- cycle    ;
	\draw [shift={(206.3,315.1)}, rotate = 20.42] [fill={rgb, 255:red, 100; green, 100; blue, 100 }  ,fill opacity=1 ][line width=0.08]  [draw opacity=0] (5.36,-2.57) -- (0,0) -- (5.36,2.57) -- (3.56,0) -- cycle    ;
	\draw [color={rgb, 255:red, 100; green, 100; blue, 100 }  ,draw opacity=1 ]   (229.88,310.87) -- (233.9,304.17) ;
	\draw [shift={(228.33,313.44)}, rotate = 300.96] [fill={rgb, 255:red, 100; green, 100; blue, 100 }  ,fill opacity=1 ][line width=0.08]  [draw opacity=0] (5.36,-2.57) -- (0,0) -- (5.36,2.57) -- (3.56,0) -- cycle    ;
	\draw [color={rgb, 255:red, 0; green, 255; blue, 255 }  ,draw opacity=0.3 ]   (234.57,294.03) -- (211.73,159.12) ;
	\draw [shift={(211.23,156.17)}, rotate = 80.39] [fill={rgb, 255:red, 0; green, 255; blue, 255 }  ,fill opacity=0.3 ][line width=0.08]  [draw opacity=0] (5.36,-2.57) -- (0,0) -- (5.36,2.57) -- (3.56,0) -- cycle    ;
	\draw [color={rgb, 255:red, 0; green, 255; blue, 255 }  ,draw opacity=0.3 ]   (201.9,306.7) -- (165.08,290.64) ;
	\draw [shift={(162.33,289.44)}, rotate = 23.56] [fill={rgb, 255:red, 0; green, 255; blue, 255 }  ,fill opacity=0.3 ][line width=0.08]  [draw opacity=0] (5.36,-2.57) -- (0,0) -- (5.36,2.57) -- (3.56,0) -- cycle    ;
	\draw [color={rgb, 255:red, 0; green, 255; blue, 255 }  ,draw opacity=0.3 ]   (189.67,274.78) -- (202.37,212.71) ;
	\draw [shift={(202.97,209.77)}, rotate = 101.56] [fill={rgb, 255:red, 0; green, 255; blue, 255 }  ,fill opacity=0.3 ][line width=0.08]  [draw opacity=0] (5.36,-2.57) -- (0,0) -- (5.36,2.57) -- (3.56,0) -- cycle    ;
	\draw [color={rgb, 255:red, 0; green, 255; blue, 255 }  ,draw opacity=0.3 ]   (154.33,275.44) -- (140.51,135.79) ;
	\draw [shift={(140.22,132.81)}, rotate = 84.35] [fill={rgb, 255:red, 0; green, 255; blue, 255 }  ,fill opacity=0.3 ][line width=0.08]  [draw opacity=0] (5.36,-2.57) -- (0,0) -- (5.36,2.57) -- (3.56,0) -- cycle    ;
	\draw  [draw opacity=0][fill={rgb, 255:red, 0; green, 0; blue, 0 }  ,fill opacity=1 ] (318.17,196.43) .. controls (318.17,193.64) and (320.44,191.37) .. (323.23,191.37) .. controls (326.03,191.37) and (328.3,193.64) .. (328.3,196.43) .. controls (328.3,199.23) and (326.03,201.5) .. (323.23,201.5) .. controls (320.44,201.5) and (318.17,199.23) .. (318.17,196.43) -- cycle ;
	\draw [color={rgb, 255:red, 0; green, 255; blue, 0 }  ,draw opacity=1 ] [dash pattern={on 0.84pt off 2.51pt}]  (323.23,217.78) -- (323.23,201.5) ;
	\draw [shift={(323.23,220.78)}, rotate = 270] [fill={rgb, 255:red, 0; green, 255; blue, 0 }  ,fill opacity=1 ][line width=0.08]  [draw opacity=0] (5.36,-2.57) -- (0,0) -- (5.36,2.57) -- (3.56,0) -- cycle    ;
	\draw  [draw opacity=0][fill={rgb, 255:red, 0; green, 0; blue, 0 }  ,fill opacity=1 ] (318.17,225.84) .. controls (318.17,223.05) and (320.44,220.78) .. (323.23,220.78) .. controls (326.03,220.78) and (328.3,223.05) .. (328.3,225.84) .. controls (328.3,228.64) and (326.03,230.91) .. (323.23,230.91) .. controls (320.44,230.91) and (318.17,228.64) .. (318.17,225.84) -- cycle ;
	\draw [color={rgb, 255:red, 255; green, 0; blue, 0 }  ,draw opacity=1 ] [dash pattern={on 0.84pt off 2.51pt}]  (323.23,247.19) -- (323.23,230.91) ;
	\draw [shift={(323.23,250.19)}, rotate = 270] [fill={rgb, 255:red, 255; green, 0; blue, 0 }  ,fill opacity=1 ][line width=0.08]  [draw opacity=0] (5.36,-2.57) -- (0,0) -- (5.36,2.57) -- (3.56,0) -- cycle    ;
	\draw  [draw opacity=0][fill={rgb, 255:red, 0; green, 0; blue, 0 }  ,fill opacity=1 ] (318.17,255.26) .. controls (318.17,252.46) and (320.44,250.19) .. (323.23,250.19) .. controls (326.03,250.19) and (328.3,252.46) .. (328.3,255.26) .. controls (328.3,258.05) and (326.03,260.32) .. (323.23,260.32) .. controls (320.44,260.32) and (318.17,258.05) .. (318.17,255.26) -- cycle ;
	\draw [color={rgb, 255:red, 0; green, 0; blue, 255 }  ,draw opacity=1 ] [dash pattern={on 0.84pt off 2.51pt}]  (323.23,276.6) -- (323.23,260.32) ;
	\draw [shift={(323.23,279.6)}, rotate = 270] [fill={rgb, 255:red, 0; green, 0; blue, 255 }  ,fill opacity=1 ][line width=0.08]  [draw opacity=0] (5.36,-2.57) -- (0,0) -- (5.36,2.57) -- (3.56,0) -- cycle    ;
	\draw  [draw opacity=0][fill={rgb, 255:red, 0; green, 0; blue, 0 }  ,fill opacity=1 ] (318.17,284.67) .. controls (318.17,281.87) and (320.44,279.6) .. (323.23,279.6) .. controls (326.03,279.6) and (328.3,281.87) .. (328.3,284.67) .. controls (328.3,287.46) and (326.03,289.73) .. (323.23,289.73) .. controls (320.44,289.73) and (318.17,287.46) .. (318.17,284.67) -- cycle ;
	
	\draw (316.25,161.5) node   [align=left] {\begin{minipage}[lt]{35.7pt}\setlength\topsep{0pt}
			Scope
	\end{minipage}};
	\draw (237.33,81) node   [align=left] {\begin{minipage}[lt]{163.65pt}\setlength\topsep{0pt}
			breadth=10: \#objects in scope
	\end{minipage}};
	\draw (281.5,222.89) node   [align=left] {\begin{minipage}[lt]{49.19pt}\setlength\topsep{0pt}
			depth=4: 
	\end{minipage}};

\end{tikzpicture}

%% file: 11.tex
\tikzset{every picture/.style={line width=0.75pt}} 

\begin{tikzpicture}[x=0.75pt,y=0.75pt,yscale=-1,xscale=1]
	
	\draw  [draw opacity=0][fill={rgb, 255:red, 255; green, 215; blue, 208 }  ,fill opacity=0.5 ] (72.44,92.51) .. controls (72.44,80.45) and (82.22,70.67) .. (94.29,70.67) -- (206.27,70.67) .. controls (218.33,70.67) and (228.11,80.45) .. (228.11,92.51) -- (228.11,158.04) .. controls (228.11,170.11) and (218.33,179.89) .. (206.27,179.89) -- (94.29,179.89) .. controls (82.22,179.89) and (72.44,170.11) .. (72.44,158.04) -- cycle ;
	\draw  [draw opacity=0][fill={rgb, 255:red, 255; green, 128; blue, 0 }  ,fill opacity=0.1 ] (300.44,92.51) .. controls (300.44,80.45) and (310.22,70.67) .. (322.29,70.67) -- (410.66,70.67) .. controls (422.72,70.67) and (432.5,80.45) .. (432.5,92.51) -- (432.5,158.04) .. controls (432.5,170.11) and (422.72,179.89) .. (410.66,179.89) -- (322.29,179.89) .. controls (310.22,179.89) and (300.44,170.11) .. (300.44,158.04) -- cycle ;
	\draw  [draw opacity=0][fill={rgb, 255:red, 0; green, 0; blue, 0 }  ,fill opacity=1 ] (139,95.5) .. controls (139,92.46) and (141.46,90) .. (144.5,90) .. controls (147.54,90) and (150,92.46) .. (150,95.5) .. controls (150,98.54) and (147.54,101) .. (144.5,101) .. controls (141.46,101) and (139,98.54) .. (139,95.5) -- cycle ;
	\draw  [draw opacity=0][fill={rgb, 255:red, 0; green, 0; blue, 0 }  ,fill opacity=1 ] (122,128.5) .. controls (122,125.46) and (124.46,123) .. (127.5,123) .. controls (130.54,123) and (133,125.46) .. (133,128.5) .. controls (133,131.54) and (130.54,134) .. (127.5,134) .. controls (124.46,134) and (122,131.54) .. (122,128.5) -- cycle ;
	\draw  [draw opacity=0][fill={rgb, 255:red, 0; green, 0; blue, 0 }  ,fill opacity=1 ] (158,128.5) .. controls (158,125.46) and (160.46,123) .. (163.5,123) .. controls (166.54,123) and (169,125.46) .. (169,128.5) .. controls (169,131.54) and (166.54,134) .. (163.5,134) .. controls (160.46,134) and (158,131.54) .. (158,128.5) -- cycle ;
	\draw [color={rgb, 255:red, 100; green, 100; blue, 100 }  ,draw opacity=1 ] [dash pattern={on 0.84pt off 2.51pt}]  (131.39,120.34) -- (141,101.9) ;
	\draw [shift={(130,123)}, rotate = 297.53] [fill={rgb, 255:red, 100; green, 100; blue, 100 }  ,fill opacity=1 ][line width=0.08]  [draw opacity=0] (5.36,-2.57) -- (0,0) -- (5.36,2.57) -- (3.56,0) -- cycle    ;
	\draw [color={rgb, 255:red, 100; green, 100; blue, 100 }  ,draw opacity=1 ] [dash pattern={on 0.84pt off 2.51pt}]  (147.08,100.68) -- (157.96,119.29) ;
	\draw [shift={(159.48,121.88)}, rotate = 239.68] [fill={rgb, 255:red, 100; green, 100; blue, 100 }  ,fill opacity=1 ][line width=0.08]  [draw opacity=0] (5.36,-2.57) -- (0,0) -- (5.36,2.57) -- (3.56,0) -- cycle    ;
	\draw  [draw opacity=0][fill={rgb, 255:red, 0; green, 0; blue, 0 }  ,fill opacity=1 ] (140.8,154.1) .. controls (140.8,151.06) and (143.26,148.6) .. (146.3,148.6) .. controls (149.34,148.6) and (151.8,151.06) .. (151.8,154.1) .. controls (151.8,157.14) and (149.34,159.6) .. (146.3,159.6) .. controls (143.26,159.6) and (140.8,157.14) .. (140.8,154.1) -- cycle ;
	\draw [color={rgb, 255:red, 100; green, 100; blue, 100 }  ,draw opacity=1 ]   (130.4,133.8) -- (140.93,147.05) ;
	\draw [shift={(142.8,149.4)}, rotate = 231.52] [fill={rgb, 255:red, 100; green, 100; blue, 100 }  ,fill opacity=1 ][line width=0.08]  [draw opacity=0] (5.36,-2.57) -- (0,0) -- (5.36,2.57) -- (3.56,0) -- cycle    ;
	\draw [color={rgb, 255:red, 100; green, 100; blue, 100 }  ,draw opacity=1 ]   (133,128.5) -- (155,128.5) ;
	\draw [shift={(158,128.5)}, rotate = 180] [fill={rgb, 255:red, 100; green, 100; blue, 100 }  ,fill opacity=1 ][line width=0.08]  [draw opacity=0] (5.36,-2.57) -- (0,0) -- (5.36,2.57) -- (3.56,0) -- cycle    ;
	\draw  [draw opacity=0][fill={rgb, 255:red, 0; green, 0; blue, 0 }  ,fill opacity=1 ] (328,107.83) .. controls (328,104.8) and (330.46,102.33) .. (333.5,102.33) .. controls (336.54,102.33) and (339,104.8) .. (339,107.83) .. controls (339,110.87) and (336.54,113.33) .. (333.5,113.33) .. controls (330.46,113.33) and (328,110.87) .. (328,107.83) -- cycle ;
	\draw  [draw opacity=0][fill={rgb, 255:red, 0; green, 0; blue, 0 }  ,fill opacity=1 ] (375,102.5) .. controls (375,99.46) and (377.46,97) .. (380.5,97) .. controls (383.54,97) and (386,99.46) .. (386,102.5) .. controls (386,105.54) and (383.54,108) .. (380.5,108) .. controls (377.46,108) and (375,105.54) .. (375,102.5) -- cycle ;
	\draw  [draw opacity=0][fill={rgb, 255:red, 0; green, 0; blue, 0 }  ,fill opacity=1 ] (331,155.17) .. controls (331,152.13) and (333.46,149.67) .. (336.5,149.67) .. controls (339.54,149.67) and (342,152.13) .. (342,155.17) .. controls (342,158.2) and (339.54,160.67) .. (336.5,160.67) .. controls (333.46,160.67) and (331,158.2) .. (331,155.17) -- cycle ;
	\draw  [draw opacity=0][fill={rgb, 255:red, 0; green, 0; blue, 0 }  ,fill opacity=1 ] (401,121.83) .. controls (401,118.8) and (403.46,116.33) .. (406.5,116.33) .. controls (409.54,116.33) and (412,118.8) .. (412,121.83) .. controls (412,124.87) and (409.54,127.33) .. (406.5,127.33) .. controls (403.46,127.33) and (401,124.87) .. (401,121.83) -- cycle ;
	\draw  [draw opacity=0][fill={rgb, 255:red, 0; green, 0; blue, 0 }  ,fill opacity=1 ] (356.33,137.83) .. controls (356.33,134.8) and (358.8,132.33) .. (361.83,132.33) .. controls (364.87,132.33) and (367.33,134.8) .. (367.33,137.83) .. controls (367.33,140.87) and (364.87,143.33) .. (361.83,143.33) .. controls (358.8,143.33) and (356.33,140.87) .. (356.33,137.83) -- cycle ;
	\draw  [draw opacity=0][fill={rgb, 255:red, 0; green, 0; blue, 0 }  ,fill opacity=1 ] (384.33,159.17) .. controls (384.33,156.13) and (386.8,153.67) .. (389.83,153.67) .. controls (392.87,153.67) and (395.33,156.13) .. (395.33,159.17) .. controls (395.33,162.2) and (392.87,164.67) .. (389.83,164.67) .. controls (386.8,164.67) and (384.33,162.2) .. (384.33,159.17) -- cycle ;
	\draw [color={rgb, 255:red, 100; green, 100; blue, 100 }  ,draw opacity=1 ]   (385,106.83) -- (398.43,116.28) ;
	\draw [shift={(400.89,118)}, rotate = 215.1] [fill={rgb, 255:red, 100; green, 100; blue, 100 }  ,fill opacity=1 ][line width=0.08]  [draw opacity=0] (5.36,-2.57) -- (0,0) -- (5.36,2.57) -- (3.56,0) -- cycle    ;
	\draw [color={rgb, 255:red, 100; green, 100; blue, 100 }  ,draw opacity=1 ]   (375.49,110.83) -- (366.86,128.63) ;
	\draw [shift={(365.56,131.33)}, rotate = 295.86] [fill={rgb, 255:red, 100; green, 100; blue, 100 }  ,fill opacity=1 ][line width=0.08]  [draw opacity=0] (5.36,-2.57) -- (0,0) -- (5.36,2.57) -- (3.56,0) -- cycle    ;
	\draw [shift={(376.8,108.13)}, rotate = 115.86] [fill={rgb, 255:red, 100; green, 100; blue, 100 }  ,fill opacity=1 ][line width=0.08]  [draw opacity=0] (5.36,-2.57) -- (0,0) -- (5.36,2.57) -- (3.56,0) -- cycle    ;
	\draw [color={rgb, 255:red, 100; green, 100; blue, 100 }  ,draw opacity=1 ]   (402.43,129.3) -- (393.57,150.17) ;
	\draw [shift={(392.4,152.93)}, rotate = 292.99] [fill={rgb, 255:red, 100; green, 100; blue, 100 }  ,fill opacity=1 ][line width=0.08]  [draw opacity=0] (5.36,-2.57) -- (0,0) -- (5.36,2.57) -- (3.56,0) -- cycle    ;
	\draw [shift={(403.6,126.53)}, rotate = 112.99] [fill={rgb, 255:red, 100; green, 100; blue, 100 }  ,fill opacity=1 ][line width=0.08]  [draw opacity=0] (5.36,-2.57) -- (0,0) -- (5.36,2.57) -- (3.56,0) -- cycle    ;
	\draw [color={rgb, 255:red, 100; green, 100; blue, 100 }  ,draw opacity=1 ]   (356.4,140.53) -- (344.37,149.89) ;
	\draw [shift={(342,151.73)}, rotate = 322.13] [fill={rgb, 255:red, 100; green, 100; blue, 100 }  ,fill opacity=1 ][line width=0.08]  [draw opacity=0] (5.36,-2.57) -- (0,0) -- (5.36,2.57) -- (3.56,0) -- cycle    ;
	\draw [color={rgb, 255:red, 100; green, 100; blue, 100 }  ,draw opacity=1 ]   (379.81,157.54) -- (342,155.17) ;
	\draw [shift={(382.8,157.73)}, rotate = 183.6] [fill={rgb, 255:red, 100; green, 100; blue, 100 }  ,fill opacity=1 ][line width=0.08]  [draw opacity=0] (5.36,-2.57) -- (0,0) -- (5.36,2.57) -- (3.56,0) -- cycle    ;
	\draw [color={rgb, 255:red, 100; green, 100; blue, 100 }  ,draw opacity=1 ]   (398.21,122.93) -- (366.8,135.33) ;
	\draw [shift={(401,121.83)}, rotate = 158.46] [fill={rgb, 255:red, 100; green, 100; blue, 100 }  ,fill opacity=1 ][line width=0.08]  [draw opacity=0] (5.36,-2.57) -- (0,0) -- (5.36,2.57) -- (3.56,0) -- cycle    ;
	\draw  [color={rgb, 255:red, 0; green, 0; blue, 0 }  ,draw opacity=0.5 ] (234.5,109.33) -- (271.1,109.33) -- (271.1,103) -- (295.5,115.67) -- (271.1,128.33) -- (271.1,122) -- (234.5,122) -- cycle ;
	\draw  [color={rgb, 255:red, 0; green, 0; blue, 0 }  ,draw opacity=0.5 ] (333.92,186.96) -- (334.15,213.55) -- (340.48,213.5) -- (327.97,231.33) -- (315.15,213.72) -- (321.48,213.66) -- (321.25,187.07) -- cycle ;
	\draw  [color={rgb, 255:red, 0; green, 0; blue, 0 }  ,draw opacity=0.5 ] (335.88,294.29) -- (336.14,323.28) -- (342.47,323.23) -- (329.97,342.67) -- (317.14,323.45) -- (323.47,323.39) -- (323.22,294.4) -- cycle ;
	\draw  [draw opacity=0][fill={rgb, 255:red, 0; green, 255; blue, 0 }  ,fill opacity=0.1 ] (156.83,235.56) -- (199,277.72) -- (156.83,319.89) -- (114.67,277.72) -- cycle ;
	\draw [color={rgb, 255:red, 255; green, 87; blue, 87 }  ,draw opacity=1 ][line width=2.25]    (157.67,233.56) -- (157.67,196.89) ;
	\draw [shift={(157.67,191.89)}, rotate = 90] [fill={rgb, 255:red, 255; green, 87; blue, 87 }  ,fill opacity=1 ][line width=0.08]  [draw opacity=0] (16.07,-7.72) -- (0,0) -- (16.07,7.72) -- (10.67,0) -- cycle    ;
	\draw [color={rgb, 255:red, 255; green, 87; blue, 87 }  ,draw opacity=1 ][line width=2.25]    (184.5,250) -- (296.76,212.58) ;
	\draw [shift={(301.5,211)}, rotate = 161.57] [fill={rgb, 255:red, 255; green, 87; blue, 87 }  ,fill opacity=1 ][line width=0.08]  [draw opacity=0] (16.07,-7.72) -- (0,0) -- (16.07,7.72) -- (10.67,0) -- cycle    ;
	\draw [color={rgb, 255:red, 255; green, 87; blue, 87 }  ,draw opacity=1 ][line width=2.25]    (189.5,298) -- (299.51,305.65) ;
	\draw [shift={(304.5,306)}, rotate = 183.98] [fill={rgb, 255:red, 255; green, 87; blue, 87 }  ,fill opacity=1 ][line width=0.08]  [draw opacity=0] (16.07,-7.72) -- (0,0) -- (16.07,7.72) -- (10.67,0) -- cycle    ;
	\draw [color={rgb, 255:red, 100; green, 100; blue, 100 }  ,draw opacity=1 ] [dash pattern={on 0.84pt off 2.51pt}]  (144.5,101) -- (146.19,145.6) ;
	\draw [shift={(146.3,148.6)}, rotate = 267.83] [fill={rgb, 255:red, 100; green, 100; blue, 100 }  ,fill opacity=1 ][line width=0.08]  [draw opacity=0] (5.36,-2.57) -- (0,0) -- (5.36,2.57) -- (3.56,0) -- cycle    ;
	
	\draw (155.67,53.42) node   [align=left] {\begin{minipage}[lt]{75.71pt}\setlength\topsep{0pt}
			World category
	\end{minipage}};
	\draw (371.22,52.25) node   [align=left] {\begin{minipage}[lt]{37.11pt}\setlength\topsep{0pt}
			Scope
	\end{minipage}};
	\draw (150.5,81.25) node   [align=left] {\begin{minipage}[lt]{15.64pt}\setlength\topsep{0pt}
			$I^\vee$
	\end{minipage}};
	\draw (106.5,126.2) node  [font=\footnotesize] [align=left] {\begin{minipage}[lt]{26.79pt}\setlength\topsep{0pt}
			robot
	\end{minipage}};
	\draw (149.3,169.8) node  [font=\footnotesize] [align=left] {\begin{minipage}[lt]{45.56pt}\setlength\topsep{0pt}
			harmless
	\end{minipage}};
	\draw (203.05,133.08) node  [font=\footnotesize] [align=left] {\begin{minipage}[lt]{44.13pt}\setlength\topsep{0pt}
			human happiness
	\end{minipage}};
	\draw (339.5,93.58) node   [align=left] {\begin{minipage}[lt]{15.64pt}\setlength\topsep{0pt}
			$I^\vee$
	\end{minipage}};
	\draw (387.61,260.33) node   [align=left] {\begin{minipage}[lt]{141.3pt}\setlength\topsep{0pt}
			Objective: as a harmless robot for human happiness, what should I do in this case?
	\end{minipage}};
	\draw (350.63,357.14) node   [align=left] {\begin{minipage}[lt]{83.01pt}\setlength\topsep{0pt}
			Planned action
	\end{minipage}};
	\draw (156.83,277.72) node   [align=left] {\begin{minipage}[lt]{35.5pt}\setlength\topsep{0pt}
			Verifier
	\end{minipage}};
	\draw (86.25,217.39) node   [align=left] {\begin{minipage}[lt]{76.5pt}\setlength\topsep{0pt}
			{\footnotesize Are you a harmless robot for human happiness?}
	\end{minipage}};
	\draw (235.75,207.05) node   [align=left] {\begin{minipage}[lt]{48.62pt}\setlength\topsep{0pt}
			{\footnotesize Is objective reasonable?}
	\end{minipage}};
	\draw (244.58,327.05) node   [align=left] {\begin{minipage}[lt]{50.21pt}\setlength\topsep{0pt}
			{\footnotesize Is action }\\{\footnotesize reasonable?}
	\end{minipage}};

\end{tikzpicture}

%% file: 9.tex
\tikzset{every picture/.style={line width=0.75pt}} 
{\hypersetup{hidelinks}
\begin{tikzpicture}[x=0.75pt,y=0.75pt,yscale=-1,xscale=1]
\pgfdeclarelayer{background}
  \pgfsetlayers{background,main}

\draw  [draw opacity=0][fill={rgb, 255:red, 0; green, 0; blue, 0 }  ,fill opacity=1 ] (109.67,221.92) .. controls (109.67,218.65) and (112.32,216) .. (115.58,216) .. controls (118.85,216) and (121.5,218.65) .. (121.5,221.92) .. controls (121.5,225.18) and (118.85,227.83) .. (115.58,227.83) .. controls (112.32,227.83) and (109.67,225.18) .. (109.67,221.92) -- cycle ;
\draw  [draw opacity=0][fill={rgb, 255:red, 0; green, 0; blue, 0 }  ,fill opacity=1 ] (154.33,195.08) .. controls (154.33,191.82) and (156.98,189.17) .. (160.25,189.17) .. controls (163.52,189.17) and (166.17,191.82) .. (166.17,195.08) .. controls (166.17,198.35) and (163.52,201) .. (160.25,201) .. controls (156.98,201) and (154.33,198.35) .. (154.33,195.08) -- cycle ;
\draw  [draw opacity=0][fill={rgb, 255:red, 0; green, 0; blue, 0 }  ,fill opacity=1 ] (101.67,168.42) .. controls (101.67,165.15) and (104.32,162.5) .. (107.58,162.5) .. controls (110.85,162.5) and (113.5,165.15) .. (113.5,168.42) .. controls (113.5,171.68) and (110.85,174.33) .. (107.58,174.33) .. controls (104.32,174.33) and (101.67,171.68) .. (101.67,168.42) -- cycle ;
\draw  [draw opacity=0][fill={rgb, 255:red, 0; green, 0; blue, 0 }  ,fill opacity=1 ] (158.17,154.42) .. controls (158.17,151.15) and (160.82,148.5) .. (164.08,148.5) .. controls (167.35,148.5) and (170,151.15) .. (170,154.42) .. controls (170,157.68) and (167.35,160.33) .. (164.08,160.33) .. controls (160.82,160.33) and (158.17,157.68) .. (158.17,154.42) -- cycle ;
\draw [color={rgb, 255:red, 100; green, 100; blue, 100 }  ,draw opacity=1 ]   (111.81,165.29) -- (155.25,155.1) ;
\draw [shift={(158.17,154.42)}, rotate = 166.8] [fill={rgb, 255:red, 100; green, 100; blue, 100 }  ,fill opacity=1 ][line width=0.08]  [draw opacity=0] (10.72,-5.15) -- (0,0) -- (10.72,5.15) -- (7.12,0) -- cycle    ;
\draw [color={rgb, 255:red, 255; green, 0; blue, 255 }  ,draw opacity=1 ]   (123.95,220.19) -- (154.48,198.62) ;
\draw [shift={(121.5,221.92)}, rotate = 324.76] [fill={rgb, 255:red, 255; green, 0; blue, 255 }  ,fill opacity=1 ][line width=0.08]  [draw opacity=0] (10.72,-5.15) -- (0,0) -- (10.72,5.15) -- (7.12,0) -- cycle    ;
\draw [color={rgb, 255:red, 100; green, 100; blue, 100 }  ,draw opacity=1 ]   (160.25,189.17) -- (163.69,163.31) ;
\draw [shift={(164.08,160.33)}, rotate = 97.57] [fill={rgb, 255:red, 100; green, 100; blue, 100 }  ,fill opacity=1 ][line width=0.08]  [draw opacity=0] (10.72,-5.15) -- (0,0) -- (10.72,5.15) -- (7.12,0) -- cycle    ;
\draw [color={rgb, 255:red, 255; green, 0; blue, 0 }  ,draw opacity=1 ]   (112.11,211.64) -- (107.95,177.31) ;
\draw [shift={(107.58,174.33)}, rotate = 83.08] [fill={rgb, 255:red, 255; green, 0; blue, 0 }  ,fill opacity=1 ][line width=0.08]  [draw opacity=0] (10.72,-5.15) -- (0,0) -- (10.72,5.15) -- (7.12,0) -- cycle    ;
\draw [shift={(112.48,214.62)}, rotate = 263.08] [fill={rgb, 255:red, 255; green, 0; blue, 0 }  ,fill opacity=1 ][line width=0.08]  [draw opacity=0] (10.72,-5.15) -- (0,0) -- (10.72,5.15) -- (7.12,0) -- cycle    ;
\draw  [draw opacity=0][fill={rgb, 255:red, 0; green, 0; blue, 0 }  ,fill opacity=1 ] (124.33,63.08) .. controls (124.33,59.82) and (126.98,57.17) .. (130.25,57.17) .. controls (133.52,57.17) and (136.17,59.82) .. (136.17,63.08) .. controls (136.17,66.35) and (133.52,69) .. (130.25,69) .. controls (126.98,69) and (124.33,66.35) .. (124.33,63.08) -- cycle ;
\draw [color={rgb, 255:red, 0; green, 255; blue, 0 }  ,draw opacity=1 ] [dash pattern={on 0.84pt off 2.51pt}]  (108.13,159.55) -- (124.86,69.5) ;
\draw [shift={(107.58,162.5)}, rotate = 280.52] [fill={rgb, 255:red, 0; green, 255; blue, 0 }  ,fill opacity=1 ][line width=0.08]  [draw opacity=0] (10.72,-5.15) -- (0,0) -- (10.72,5.15) -- (7.12,0) -- cycle    ;
\draw [color={rgb, 255:red, 0; green, 255; blue, 0 }  ,draw opacity=1 ] [dash pattern={on 0.84pt off 2.51pt}]  (115.83,213.01) -- (127.5,69.61) ;
\draw [shift={(115.58,216)}, rotate = 274.65] [fill={rgb, 255:red, 0; green, 255; blue, 0 }  ,fill opacity=1 ][line width=0.08]  [draw opacity=0] (10.72,-5.15) -- (0,0) -- (10.72,5.15) -- (7.12,0) -- cycle    ;
\draw [color={rgb, 255:red, 0; green, 255; blue, 0 }  ,draw opacity=1 ] [dash pattern={on 0.84pt off 2.51pt}]  (156.34,184.68) -- (130.25,69) ;
\draw [shift={(157,187.61)}, rotate = 257.29] [fill={rgb, 255:red, 0; green, 255; blue, 0 }  ,fill opacity=1 ][line width=0.08]  [draw opacity=0] (10.72,-5.15) -- (0,0) -- (10.72,5.15) -- (7.12,0) -- cycle    ;
\draw [color={rgb, 255:red, 0; green, 255; blue, 0 }  ,draw opacity=1 ] [dash pattern={on 0.84pt off 2.51pt}]  (160.01,142.78) -- (134,68.61) ;
\draw [shift={(161,145.61)}, rotate = 250.68] [fill={rgb, 255:red, 0; green, 255; blue, 0 }  ,fill opacity=1 ][line width=0.08]  [draw opacity=0] (10.72,-5.15) -- (0,0) -- (10.72,5.15) -- (7.12,0) -- cycle    ;
\draw    (220,19) -- (220,275.21) ;
\draw  [draw opacity=0][fill={rgb, 255:red, 0; green, 0; blue, 0 }  ,fill opacity=1 ] (277.56,221.71) .. controls (277.56,218.45) and (280.21,215.8) .. (283.48,215.8) .. controls (286.74,215.8) and (289.39,218.45) .. (289.39,221.71) .. controls (289.39,224.98) and (286.74,227.63) .. (283.48,227.63) .. controls (280.21,227.63) and (277.56,224.98) .. (277.56,221.71) -- cycle ;
\draw  [draw opacity=0][fill={rgb, 255:red, 0; green, 0; blue, 0 }  ,fill opacity=1 ] (333.33,221.26) .. controls (333.33,217.99) and (335.98,215.35) .. (339.25,215.35) .. controls (342.52,215.35) and (345.17,217.99) .. (345.17,221.26) .. controls (345.17,224.53) and (342.52,227.18) .. (339.25,227.18) .. controls (335.98,227.18) and (333.33,224.53) .. (333.33,221.26) -- cycle ;
\draw  [draw opacity=0][fill={rgb, 255:red, 0; green, 0; blue, 0 }  ,fill opacity=1 ] (272.67,169.6) .. controls (272.67,166.33) and (275.32,163.68) .. (278.58,163.68) .. controls (281.85,163.68) and (284.5,166.33) .. (284.5,169.6) .. controls (284.5,172.86) and (281.85,175.51) .. (278.58,175.51) .. controls (275.32,175.51) and (272.67,172.86) .. (272.67,169.6) -- cycle ;
\draw  [draw opacity=0][fill={rgb, 255:red, 0; green, 0; blue, 0 }  ,fill opacity=1 ] (306.99,149.03) .. controls (306.99,145.76) and (309.64,143.11) .. (312.9,143.11) .. controls (316.17,143.11) and (318.82,145.76) .. (318.82,149.03) .. controls (318.82,152.3) and (316.17,154.95) .. (312.9,154.95) .. controls (309.64,154.95) and (306.99,152.3) .. (306.99,149.03) -- cycle ;
\draw [color={rgb, 255:red, 100; green, 100; blue, 100 }  ,draw opacity=1 ]   (284.5,169.6) -- (304.77,151.05) ;
\draw [shift={(306.99,149.03)}, rotate = 137.56] [fill={rgb, 255:red, 100; green, 100; blue, 100 }  ,fill opacity=1 ][line width=0.08]  [draw opacity=0] (10.72,-5.15) -- (0,0) -- (10.72,5.15) -- (7.12,0) -- cycle    ;
\draw [color={rgb, 255:red, 100; green, 100; blue, 100 }  ,draw opacity=1 ]   (284.67,213.04) -- (308.57,157.79) ;
\draw [shift={(283.48,215.8)}, rotate = 293.39] [fill={rgb, 255:red, 100; green, 100; blue, 100 }  ,fill opacity=1 ][line width=0.08]  [draw opacity=0] (10.72,-5.15) -- (0,0) -- (10.72,5.15) -- (7.12,0) -- cycle    ;
\draw [color={rgb, 255:red, 100; green, 100; blue, 100 }  ,draw opacity=1 ]   (342.57,180.79) -- (320.61,156.99) ;
\draw [shift={(318.57,154.79)}, rotate = 47.29] [fill={rgb, 255:red, 100; green, 100; blue, 100 }  ,fill opacity=1 ][line width=0.08]  [draw opacity=0] (10.72,-5.15) -- (0,0) -- (10.72,5.15) -- (7.12,0) -- cycle    ;
\draw [color={rgb, 255:red, 100; green, 100; blue, 100 }  ,draw opacity=1 ]   (283.11,212.82) -- (278.58,175.51) ;
\draw [shift={(283.48,215.8)}, rotate = 263.08] [fill={rgb, 255:red, 100; green, 100; blue, 100 }  ,fill opacity=1 ][line width=0.08]  [draw opacity=0] (10.72,-5.15) -- (0,0) -- (10.72,5.15) -- (7.12,0) -- cycle    ;
\draw  [draw opacity=0][fill={rgb, 255:red, 0; green, 0; blue, 0 }  ,fill opacity=1 ] (295.33,64.26) .. controls (295.33,60.99) and (297.98,58.35) .. (301.25,58.35) .. controls (304.52,58.35) and (307.17,60.99) .. (307.17,64.26) .. controls (307.17,67.53) and (304.52,70.18) .. (301.25,70.18) .. controls (297.98,70.18) and (295.33,67.53) .. (295.33,64.26) -- cycle ;
\draw [color={rgb, 255:red, 0; green, 0; blue, 0 }  ,draw opacity=1 ] [dash pattern={on 0.84pt off 2.51pt}]  (279.13,160.73) -- (295.86,70.68) ;
\draw [shift={(278.58,163.68)}, rotate = 280.52] [fill={rgb, 255:red, 0; green, 0; blue, 0 }  ,fill opacity=1 ][line width=0.08]  [draw opacity=0] (10.72,-5.15) -- (0,0) -- (10.72,5.15) -- (7.12,0) -- cycle    ;
\draw [color={rgb, 255:red, 100; green, 100; blue, 100 }  ,draw opacity=1 ]   (117.85,174.58) -- (151.57,190.79) ;
\draw [shift={(115.14,173.29)}, rotate = 25.66] [fill={rgb, 255:red, 100; green, 100; blue, 100 }  ,fill opacity=1 ][line width=0.08]  [draw opacity=0] (10.72,-5.15) -- (0,0) -- (10.72,5.15) -- (7.12,0) -- cycle    ;
\draw  [draw opacity=0][fill={rgb, 255:red, 0; green, 0; blue, 0 }  ,fill opacity=1 ] (342.38,186.68) .. controls (342.38,183.41) and (345.03,180.76) .. (348.3,180.76) .. controls (351.57,180.76) and (354.21,183.41) .. (354.21,186.68) .. controls (354.21,189.95) and (351.57,192.6) .. (348.3,192.6) .. controls (345.03,192.6) and (342.38,189.95) .. (342.38,186.68) -- cycle ;
\draw [color={rgb, 255:red, 100; green, 100; blue, 100 }  ,draw opacity=1 ]   (339.79,188.19) -- (290.57,216.79) ;
\draw [shift={(342.38,186.68)}, rotate = 149.84] [fill={rgb, 255:red, 100; green, 100; blue, 100 }  ,fill opacity=1 ][line width=0.08]  [draw opacity=0] (10.72,-5.15) -- (0,0) -- (10.72,5.15) -- (7.12,0) -- cycle    ;
\draw [color={rgb, 255:red, 100; green, 100; blue, 100 }  ,draw opacity=1 ]   (313.9,157.78) -- (333.57,213.79) ;
\draw [shift={(312.9,154.95)}, rotate = 70.65] [fill={rgb, 255:red, 100; green, 100; blue, 100 }  ,fill opacity=1 ][line width=0.08]  [draw opacity=0] (10.72,-5.15) -- (0,0) -- (10.72,5.15) -- (7.12,0) -- cycle    ;
\draw [color={rgb, 255:red, 100; green, 100; blue, 100 }  ,draw opacity=1 ]   (292.39,221.68) -- (333.33,221.26) ;
\draw [shift={(289.39,221.71)}, rotate = 359.41] [fill={rgb, 255:red, 100; green, 100; blue, 100 }  ,fill opacity=1 ][line width=0.08]  [draw opacity=0] (10.72,-5.15) -- (0,0) -- (10.72,5.15) -- (7.12,0) -- cycle    ;
\draw [color={rgb, 255:red, 0; green, 0; blue, 0 }  ,draw opacity=1 ] [dash pattern={on 0.84pt off 2.51pt}]  (283.81,212.82) -- (299.33,73.67) ;
\draw [shift={(283.48,215.8)}, rotate = 276.37] [fill={rgb, 255:red, 0; green, 0; blue, 0 }  ,fill opacity=1 ][line width=0.08]  [draw opacity=0] (10.72,-5.15) -- (0,0) -- (10.72,5.15) -- (7.12,0) -- cycle    ;
\draw [color={rgb, 255:red, 0; green, 0; blue, 0 }  ,draw opacity=1 ] [dash pattern={on 0.84pt off 2.51pt}]  (312.43,140.15) -- (301.25,70.18) ;
\draw [shift={(312.9,143.11)}, rotate = 260.92] [fill={rgb, 255:red, 0; green, 0; blue, 0 }  ,fill opacity=1 ][line width=0.08]  [draw opacity=0] (10.72,-5.15) -- (0,0) -- (10.72,5.15) -- (7.12,0) -- cycle    ;
\draw [color={rgb, 255:red, 0; green, 0; blue, 0 }  ,draw opacity=1 ] [dash pattern={on 0.84pt off 2.51pt}]  (347.23,177.96) -- (306.57,71.64) ;
\draw [shift={(348.3,180.76)}, rotate = 249.07] [fill={rgb, 255:red, 0; green, 0; blue, 0 }  ,fill opacity=1 ][line width=0.08]  [draw opacity=0] (10.72,-5.15) -- (0,0) -- (10.72,5.15) -- (7.12,0) -- cycle    ;
\draw [color={rgb, 255:red, 0; green, 0; blue, 0 }  ,draw opacity=1 ] [dash pattern={on 0.84pt off 2.51pt}]  (338.52,212.44) -- (303.33,73) ;
\draw [shift={(339.25,215.35)}, rotate = 255.84] [fill={rgb, 255:red, 0; green, 0; blue, 0 }  ,fill opacity=1 ][line width=0.08]  [draw opacity=0] (10.72,-5.15) -- (0,0) -- (10.72,5.15) -- (7.12,0) -- cycle    ;
\draw [color={rgb, 255:red, 0; green, 0; blue, 255 }  ,draw opacity=1 ]   (269.7,169.16) -- (170,154.42) ;
\draw [shift={(272.67,169.6)}, rotate = 188.41] [fill={rgb, 255:red, 0; green, 0; blue, 255 }  ,fill opacity=1 ][line width=0.08]  [draw opacity=0] (10.72,-5.15) -- (0,0) -- (10.72,5.15) -- (7.12,0) -- cycle    ;
\draw [color={rgb, 255:red, 0; green, 0; blue, 255 }  ,draw opacity=1 ]   (274.64,221.02) -- (166.17,195.08) ;
\draw [shift={(277.56,221.71)}, rotate = 193.45] [fill={rgb, 255:red, 0; green, 0; blue, 255 }  ,fill opacity=1 ][line width=0.08]  [draw opacity=0] (10.72,-5.15) -- (0,0) -- (10.72,5.15) -- (7.12,0) -- cycle    ;
\draw [color={rgb, 255:red, 100; green, 100; blue, 100 }  ,draw opacity=1 ]   (289.86,226.93) .. controls (307.78,235.53) and (325.92,234.21) .. (336.76,228.62) ;
\draw [shift={(339.25,227.18)}, rotate = 147] [fill={rgb, 255:red, 100; green, 100; blue, 100 }  ,fill opacity=1 ][line width=0.08]  [draw opacity=0] (10.72,-5.15) -- (0,0) -- (10.72,5.15) -- (7.12,0) -- cycle    ;
\draw [color={rgb, 255:red, 255; green, 0; blue, 255 }  ,draw opacity=1 ]   (121.5,221.92) .. controls (139.04,230.33) and (157.25,210.93) .. (155.66,201.29) ;
\draw [shift={(154.48,198.62)}, rotate = 50.19] [fill={rgb, 255:red, 255; green, 0; blue, 255 }  ,fill opacity=1 ][line width=0.08]  [draw opacity=0] (10.72,-5.15) -- (0,0) -- (10.72,5.15) -- (7.12,0) -- cycle    ;
\draw [color={rgb, 255:red, 255; green, 0; blue, 255 }  ,draw opacity=1 ]   (122.27,218.8) .. controls (127.66,198.96) and (138.92,194.66) .. (151.77,197.85) ;
\draw [shift={(154.48,198.62)}, rotate = 197.79] [fill={rgb, 255:red, 255; green, 0; blue, 255 }  ,fill opacity=1 ][line width=0.08]  [draw opacity=0] (10.72,-5.15) -- (0,0) -- (10.72,5.15) -- (7.12,0) -- cycle    ;
\draw [shift={(121.5,221.92)}, rotate = 282.59] [fill={rgb, 255:red, 255; green, 0; blue, 255 }  ,fill opacity=1 ][line width=0.08]  [draw opacity=0] (10.72,-5.15) -- (0,0) -- (10.72,5.15) -- (7.12,0) -- cycle    ;
\draw [color={rgb, 255:red, 100; green, 100; blue, 100 }  ,draw opacity=1 ]   (292.33,64.24) -- (136.17,63.08) ;
\draw [shift={(295.33,64.26)}, rotate = 180.42] [fill={rgb, 255:red, 100; green, 100; blue, 100 }  ,fill opacity=1 ][line width=0.08]  [draw opacity=0] (10.72,-5.15) -- (0,0) -- (10.72,5.15) -- (7.12,0) -- cycle    ;
\draw [color={rgb, 255:red, 255; green, 128; blue, 0 }  ,draw opacity=1 ] [dash pattern={on 4.5pt off 4.5pt}]  (174.86,142.93) .. controls (113.48,45.91) and (201.07,73.36) .. (293.07,70.77) ;
\draw [shift={(295.86,70.68)}, rotate = 178] [fill={rgb, 255:red, 255; green, 128; blue, 0 }  ,fill opacity=1 ][line width=0.08]  [draw opacity=0] (10.72,-5.15) -- (0,0) -- (10.72,5.15) -- (7.12,0) -- cycle    ;

\draw (135.86,262.82) node  [xscale=0.8,yscale=0.8] [align=left] {\begin{minipage}[lt]{60.76pt}\setlength\topsep{0pt}
	Category $A^\vee$
	\end{minipage}};
\draw (135.57,39) node  [xscale=0.8,yscale=0.8] [align=left] {\begin{minipage}[lt]{75.58pt}\setlength\topsep{0pt}
	Projective Limit
	\end{minipage}};
\draw (306.57,40.18) node  [xscale=0.8,yscale=0.8] [align=left] {\begin{minipage}[lt]{75.58pt}\setlength\topsep{0pt}
	Projective Limit
	\end{minipage}};
\begin{pgfonlayer}{background}

\draw (225,184.11) node  [color={rgb, 255:red, 0; green, 0; blue, 255 }  ,opacity=1] [align=left] {
	\fontsize{6}{6}\selectfont \textcolor[rgb]{0,0,1}{CLIP~\citep{radford2021learning}}
};
\draw (155,235.11) node  [color={rgb, 255:red, 0; green, 0; blue, 255 }  ,opacity=1] [align=left] {
	\fontsize{6}{6}\selectfont \textcolor[rgb]{1,0,1}{GPT~\citep{brown2020language}}
};

\draw (69.21,194.11) node  [color={rgb, 255:red, 0; green, 0; blue, 255 }  ,opacity=1] [align=left] {
	\fontsize{6}{6}\selectfont \textcolor[rgb]{1,0,0}{SimCLR~\citep{chen2020simple}}
};

\draw (222.71,88.11) node  [color={rgb, 255:red, 0; green, 0; blue, 255 }  ,opacity=1] [align=left] {
	\fontsize{6}{6}\selectfont \textcolor[rgb]{1,0.5,0}{FLIP~\citep{li2022scaling}}
};
\draw (222.71,98.11) node  [color={rgb, 255:red, 0; green, 0; blue, 255 }  ,opacity=1] [align=left] {
	\fontsize{6}{6}\selectfont \textcolor[rgb]{1,0.5,0}{EVA~\citep{fang2022eva}}
};

\draw (79.21,92.93) node  [color={rgb, 255:red, 0; green, 0; blue, 255 }  ,opacity=1] [align=left] {
\fontsize{6}{6}\selectfont \textcolor[rgb]{0,1,0}{MAE~\citep{he2022masked}}
};
\draw (79.21,105.93) node  [color={rgb, 255:red, 0; green, 0; blue, 255 }  ,opacity=1] [align=left] {
	\fontsize{6}{6}\selectfont \textcolor[rgb]{0,1,0}{Bert~\citep{devlin2018bert}}
};
\end{pgfonlayer}

\draw (313.86,263.82) node  [xscale=0.8,yscale=0.8] [align=left] {\begin{minipage}[lt]{60.76pt}\setlength\topsep{0pt}
	Category $B^\vee$
	\end{minipage}};

\end{tikzpicture}
}

%% file: paper.bbl
\begin{thebibliography}{}

\bibitem[Ad{\'a}mek et~al., 1990]{adamek1990abstract}
Ad{\'a}mek, J., Herrlich, H., and Strecker, G. (1990).
\newblock {\em Abstract and concrete categories}.
\newblock Wiley-Interscience.

\bibitem[Blum and Blum, 2022]{blum2022theory}
Blum, L. and Blum, M. (2022).
\newblock A theory of consciousness from a theoretical computer science
  perspective: Insights from the conscious turing machine.
\newblock {\em Proceedings of the National Academy of Sciences},
  119(21):e2115934119.

\bibitem[Botvinick and Cohen, 1998]{botvinick1998rubber}
Botvinick, M. and Cohen, J. (1998).
\newblock Rubber hands ‘feel’touch that eyes see.
\newblock {\em Nature}, 391(6669):756--756.

\bibitem[Brown et~al., 2020]{brown2020language}
Brown, T., Mann, B., Ryder, N., Subbiah, M., Kaplan, J.~D., Dhariwal, P.,
  Neelakantan, A., Shyam, P., Sastry, G., Askell, A., et~al. (2020).
\newblock Language models are few-shot learners.
\newblock {\em Advances in neural information processing systems},
  33:1877--1901.

\bibitem[Buetler et~al., 2022]{buetler2022tricking}
Buetler, K.~A., Penalver-Andres, J., {\"O}zen, {\"O}., Ferriroli, L., M{\"u}ri,
  R.~M., Cazzoli, D., and Marchal-Crespo, L. (2022).
\newblock “tricking the brain” using immersive virtual reality: Modifying
  the self-perception over embodied avatar influences motor cortical
  excitability and action initiation.
\newblock {\em Frontiers in human neuroscience}, 15:814.

\bibitem[Chen et~al., 2020]{chen2020simple}
Chen, T., Kornblith, S., Norouzi, M., and Hinton, G. (2020).
\newblock A simple framework for contrastive learning of visual
  representations.
\newblock In {\em International conference on machine learning}, pages
  1597--1607. PMLR.

\bibitem[Devlin et~al., 2018]{devlin2018bert}
Devlin, J., Chang, M.-W., Lee, K., and Toutanova, K. (2018).
\newblock Bert: Pre-training of deep bidirectional transformers for language
  understanding.
\newblock {\em arXiv preprint arXiv:1810.04805}.

\bibitem[Ehrsson et~al., 2004]{ehrsson2004s}
Ehrsson, H.~H., Spence, C., and Passingham, R.~E. (2004).
\newblock That's my hand! activity in premotor cortex reflects feeling of
  ownership of a limb.
\newblock {\em Science}, 305(5685):875--877.

\bibitem[Fang et~al., 2022]{fang2022eva}
Fang, Y., Wang, W., Xie, B., Sun, Q., Wu, L., Wang, X., Huang, T., Wang, X.,
  and Cao, Y. (2022).
\newblock Eva: Exploring the limits of masked visual representation learning at
  scale.
\newblock {\em arXiv preprint arXiv:2211.07636}.

\bibitem[Graziano, 2022]{graziano2022conceptual}
Graziano, M.~S. (2022).
\newblock A conceptual framework for consciousness.
\newblock {\em Proceedings of the National Academy of Sciences},
  119(18):e2116933119.

\bibitem[Guterstam et~al., 2015]{guterstam2015posterior}
Guterstam, A., Bj{\"o}rnsdotter, M., Gentile, G., and Ehrsson, H.~H. (2015).
\newblock Posterior cingulate cortex integrates the senses of self-location and
  body ownership.
\newblock {\em Current Biology}, 25(11):1416--1425.

\bibitem[He et~al., 2022]{he2022masked}
He, K., Chen, X., Xie, S., Li, Y., Doll{\'a}r, P., and Girshick, R. (2022).
\newblock Masked autoencoders are scalable vision learners.
\newblock In {\em Proceedings of the IEEE/CVF Conference on Computer Vision and
  Pattern Recognition}, pages 16000--16009.

\bibitem[Kilteni et~al., 2012]{kilteni2012sense}
Kilteni, K., Groten, R., and Slater, M. (2012).
\newblock The sense of embodiment in virtual reality.
\newblock {\em Presence: Teleoperators and Virtual Environments},
  21(4):373--387.

\bibitem[LeCun, 2022]{lecun2022path}
LeCun, Y. (2022).
\newblock A path towards autonomous machine intelligence version 0.9. 2,
  2022-06-27.
\newblock {\em Open Review}, 62.

\bibitem[Lee et~al., 2021]{lee2021predicting}
Lee, J.~D., Lei, Q., Saunshi, N., and Zhuo, J. (2021).
\newblock Predicting what you already know helps: Provable self-supervised
  learning.
\newblock {\em Advances in Neural Information Processing Systems}, 34:309--323.

\bibitem[Li, 2017]{li2017deep}
Li, Y. (2017).
\newblock Deep reinforcement learning: An overview.
\newblock {\em arXiv preprint arXiv:1701.07274}.

\bibitem[Li et~al., 2022]{li2022scaling}
Li, Y., Fan, H., Hu, R., Feichtenhofer, C., and He, K. (2022).
\newblock Scaling language-image pre-training via masking.
\newblock {\em arXiv preprint arXiv:2212.00794}.

\bibitem[Lundberg and Lee, 2017]{lundberg2017unified}
Lundberg, S.~M. and Lee, S.-I. (2017).
\newblock A unified approach to interpreting model predictions.
\newblock {\em Advances in neural information processing systems}, 30.

\bibitem[Mac~Lane, 2013]{mac2013categories}
Mac~Lane, S. (2013).
\newblock {\em Categories for the working mathematician}, volume~5.
\newblock Springer Science \& Business Media.

\bibitem[Masaki~Kashiwara, 2006]{kashiwara2006categories}
Masaki~Kashiwara, P.~S. (2006).
\newblock {\em Categories and Sheaves}.
\newblock Springer.

\bibitem[Pavone et~al., 2016]{pavone2016embodying}
Pavone, E.~F., Tieri, G., Rizza, G., Tidoni, E., Grisoni, L., and Aglioti,
  S.~M. (2016).
\newblock Embodying others in immersive virtual reality: electro-cortical
  signatures of monitoring the errors in the actions of an avatar seen from a
  first-person perspective.
\newblock {\em Journal of Neuroscience}, 36(2):268--279.

\bibitem[Radford et~al., 2021]{radford2021learning}
Radford, A., Kim, J.~W., Hallacy, C., Ramesh, A., Goh, G., Agarwal, S., Sastry,
  G., Askell, A., Mishkin, P., Clark, J., et~al. (2021).
\newblock Learning transferable visual models from natural language
  supervision.
\newblock In {\em International Conference on Machine Learning}, pages
  8748--8763. PMLR.

\bibitem[Radford et~al., 2019]{radford2019language}
Radford, A., Wu, J., Child, R., Luan, D., Amodei, D., Sutskever, I., et~al.
  (2019).
\newblock Language models are unsupervised multitask learners.
\newblock {\em OpenAI blog}, 1(8):9.

\bibitem[Ramesh et~al., 2022]{ramesh2022hierarchical}
Ramesh, A., Dhariwal, P., Nichol, A., Chu, C., and Chen, M. (2022).
\newblock Hierarchical text-conditional image generation with clip latents.
\newblock {\em arXiv preprint arXiv:2204.06125}.

\bibitem[Ramesh et~al., 2021]{ramesh2021zero}
Ramesh, A., Pavlov, M., Goh, G., Gray, S., Voss, C., Radford, A., Chen, M., and
  Sutskever, I. (2021).
\newblock Zero-shot text-to-image generation.
\newblock In {\em International Conference on Machine Learning}, pages
  8821--8831. PMLR.

\bibitem[Riehl, 2017]{riehl2017category}
Riehl, E. (2017).
\newblock {\em Category theory in context}.
\newblock Courier Dover Publications.

\bibitem[Rombach et~al., 2022]{rombach2022high}
Rombach, R., Blattmann, A., Lorenz, D., Esser, P., and Ommer, B. (2022).
\newblock High-resolution image synthesis with latent diffusion models.
\newblock In {\em Proceedings of the IEEE/CVF Conference on Computer Vision and
  Pattern Recognition}, pages 10684--10695.

\bibitem[Sohl-Dickstein et~al., 2015]{sohl2015deep}
Sohl-Dickstein, J., Weiss, E., Maheswaranathan, N., and Ganguli, S. (2015).
\newblock Deep unsupervised learning using nonequilibrium thermodynamics.
\newblock In {\em International Conference on Machine Learning}, pages
  2256--2265. PMLR.

\bibitem[Sundararajan et~al., 2017]{sundararajan2017axiomatic}
Sundararajan, M., Taly, A., and Yan, Q. (2017).
\newblock Axiomatic attribution for deep networks.
\newblock In {\em International conference on machine learning}, pages
  3319--3328. PMLR.

\bibitem[Sutton and Barto, 2018]{sutton2018reinforcement}
Sutton, R.~S. and Barto, A.~G. (2018).
\newblock {\em Reinforcement learning: An introduction}.
\newblock MIT press.

\bibitem[Tan et~al., 2023]{tan2023contrastive}
Tan, Z., Zhang, Y., Yang, J., and Yuan, Y. (2023).
\newblock Contrastive learning is spectral clustering on similarity graph.
\newblock {\em arXiv preprint arXiv:2303.15103}.

\bibitem[Tsakiris and Haggard, 2005]{tsakiris2005rubber}
Tsakiris, M. and Haggard, P. (2005).
\newblock The rubber hand illusion revisited: visuotactile integration and
  self-attribution.
\newblock {\em Journal of experimental psychology: Human perception and
  performance}, 31(1):80.

\bibitem[Tsuchiya and Saigo, 2021]{tsuchiya2021relational}
Tsuchiya, N. and Saigo, H. (2021).
\newblock A relational approach to consciousness: categories of level and
  contents of consciousness.
\newblock {\em Neuroscience of Consciousness}, 2021(2):niab034.

\bibitem[Yuan, 2023a]{yuan2022power}
Yuan, Y. (2023a).
\newblock On the power of foundation models.
\newblock In {\em International Conference on Machine Learning}. PMLR.

\bibitem[Yuan, 2023b]{yuan2023concept}
Yuan, Y. (2023b).
\newblock Succinct representations for concepts.
\newblock {\em arXiv preprint arXiv:2303.00446}.

\end{thebibliography}
